\documentclass[acmtog]{acmart}
\acmSubmissionID{664}

\usepackage{booktabs} 
\usepackage{xcolor}
\usepackage{listings}
\usepackage{colortbl}
\usepackage{cancel}

\definecolor{codegreen}{rgb}{0,0.6,0}
\definecolor{codegray}{rgb}{0.5,0.5,0.5}
\definecolor{codepurple}{rgb}{0.58,0,0.82}
\definecolor{backcolour}{rgb}{0.95,0.95,0.92}

\lstdefinestyle{mystyle}{
    backgroundcolor=\color{backcolour},   
    commentstyle=\color{codegreen},
    keywordstyle=\color{magenta},
    numberstyle=\tiny\color{codegray},
    stringstyle=\color{codepurple},
    basicstyle=\ttfamily\footnotesize,
    breakatwhitespace=false,         
    breaklines=true,                 
    captionpos=b,                    
    keepspaces=true,                 
    numbers=left,                    
    numbersep=5pt,                  
    showspaces=false,                
    showstringspaces=false,
    showtabs=false,                  
    tabsize=2
}

\colorlet{yellowish}{green!10!orange}
\colorlet{greenish}{green!70!orange}
\colorlet{blueish}{blue!70}
\colorlet{purpleish}{purple!70}
\colorlet{reddish}{red!70}

\usepackage[ruled]{algorithm2e} 
\usepackage[noabbrev, capitalise]{cleveref}
\usepackage{caption}
\usepackage{subcaption}
\usepackage{dirtytalk}
\usepackage{selectp}
\usepackage{booktabs}
\usepackage{multirow}
\usepackage{fontawesome5}
\usepackage{mathtools}
\usepackage[normalem]{ulem}
\usepackage{lipsum}

\SetAlFnt{\small}
\SetAlCapFnt{\small}
\SetAlCapNameFnt{\small}
\SetAlCapHSkip{0pt}

\newcommand\numberthis{\addtocounter{equation}{1}\tag{\theequation}}

\newcommand{\alg}{MaskedMimic}

\acmJournal{TOG}

\setcopyright{rightsretained}

\acmYear{2024} \acmVolume{43} \acmNumber{6} \acmArticle{209} \acmMonth{12}\acmDOI{10.1145/3687951}

\begin{document}
\title{\alg: Unified Physics-Based Character Control Through Masked Motion Inpainting}

\author{Chen Tessler}
\affiliation{%
 \institution{NVIDIA}
 \country{Israel}}
\email{ctessler@nvidia.com}
\author{Yunrong Guo}
\affiliation{%
\institution{NVIDIA}
\country{Canada}
}
\email{kellyg@nvidia.com}
\author{Ofir Nabati}
\affiliation{%
\institution{NVIDIA}
\country{Israel}
}
\email{ofirnabati@gmail.com}
\author{Gal Chechik}
\affiliation{%
 \institution{NVIDIA}
 \country{Israel}
 }
 \affiliation{%
 \institution{Bar-Ilan University}
 \country{Israel}
 }
 \email{gchechik@nvidia.com}
\author{Xue Bin Peng}
\affiliation{%
 \institution{NVIDIA}
 \country{Canada}
}
\affiliation{%
 \institution{Simon Fraser University}
 \country{Canada}
}
\email{japeng@nvidia.com}

\renewcommand\shortauthors{Tessler, C. et al}

\begin{abstract}
    Crafting a single, versatile physics-based controller that can breathe life into interactive characters across a wide spectrum of scenarios represents an exciting frontier in character animation.
    An ideal controller should support diverse control modalities, such as sparse target keyframes, text instructions, and scene information. While previous works have proposed physically simulated, scene-aware control models, these systems have predominantly focused on developing controllers that each specializes in a narrow set of tasks and control modalities. 
    This work presents \alg, a novel approach that formulates physics-based character control as a general motion inpainting problem. Our key insight is to train a single unified model to synthesize motions from partial (masked) motion descriptions, such as masked keyframes, objects, text descriptions, or any combination thereof. This is achieved by leveraging motion tracking data and designing a scalable training method that can effectively utilize diverse motion descriptions to produce coherent animations.
    Through this process, our approach learns a physics-based controller that provides an intuitive control interface without requiring tedious reward engineering for all behaviors of interest. The resulting controller supports a wide range of control modalities and enables seamless transitions between disparate tasks.
    By unifying character control through motion inpainting, \alg~ creates versatile virtual characters. These characters can dynamically adapt to complex scenes and compose diverse motions on demand, enabling more interactive and immersive experiences.
\end{abstract}

%
%

\ccsdesc[500]{Computing methodologies~Physical simulation}
\ccsdesc[500]{Computing methodologies~Procedural animation}

%
%

\keywords{reinforcement learning, animated character control, motion tracking, motion capture data}

\begin{teaserfigure}
  \includegraphics[trim={3cm 0cm 3cm 0.02cm},clip,width=\textwidth]{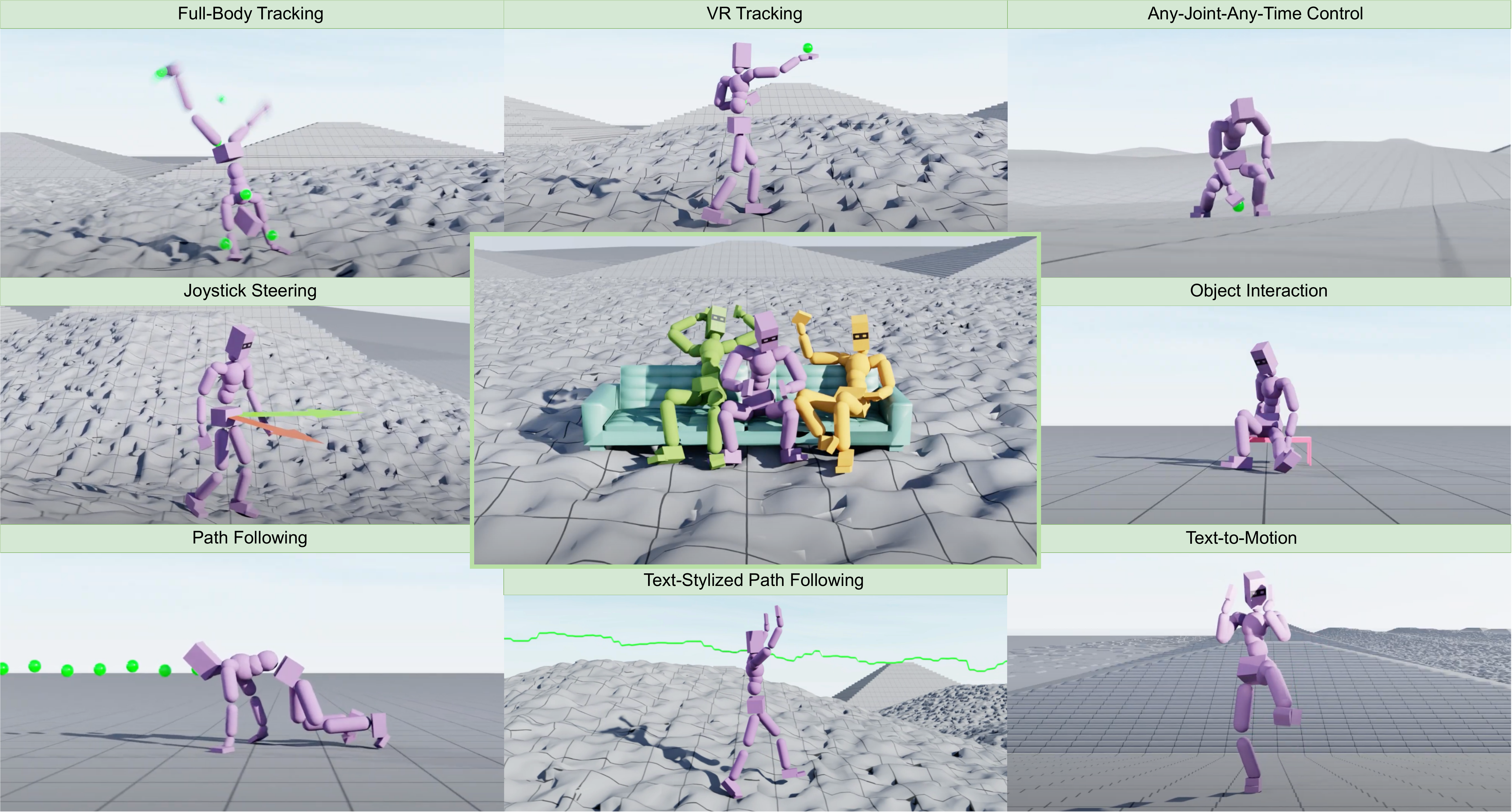}\\
  \caption{We present \alg, a versatile control model that enables physically simulated characters to generate diverse behaviors from flexible user-specified constraints. \alg~ can be used for a wide range of applications, including generating full-body motion from partially observed joint target positions, joystick steering, object interactions, path-following, text commands, and combinations thereof, such as text-stylized path following.}
  \label{fig:teaser}
\end{teaserfigure}

\maketitle

\begin{figure*}[t!]
    \centering
    \includegraphics[width=0.95\textwidth]{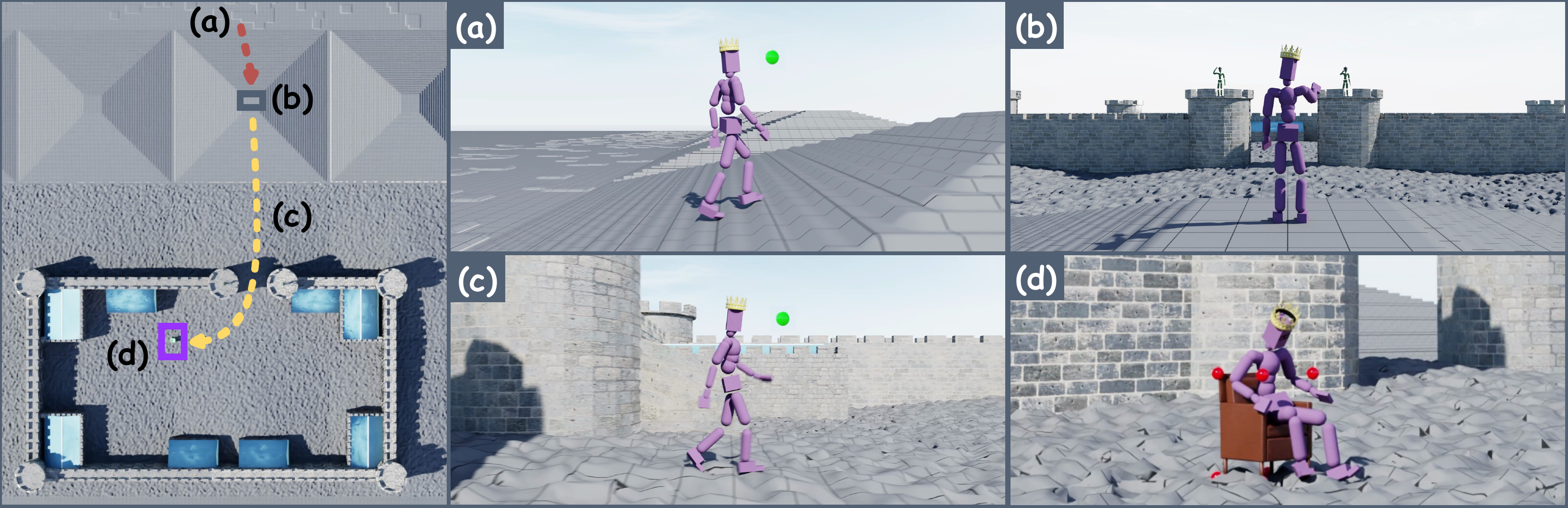}
    \caption{\textbf{Partial motion plans.} \alg~ synthesizes full-body physics-based character animations. It achieves this by inpainting conditioned on multi-modal partial objectives. (a) The character climbs up a hill by tracking target head coordinates. (b) Text-to-motion synthesis enables the character to perform a waving motion. (c) The character navigates across irregular terrain by combining head-tracking with text-based style conditioning. (d) Interacting with a goal object, in this case sitting on an armchair, is achieved by conditioning on the object.}
    \label{fig: castle}
\end{figure*}

\section{Introduction}

The development of virtual characters capable of following dynamic user instructions and interacting with diverse scenes has been a significant challenge in computer graphics. This challenge spans a wide range of applications, including gaming, digital humans, virtual reality, and many more. For instance, a character might be instructed to "Climb the hill to the castle, wave to the guard, go inside, navigate to the throne room, and sit on the throne". This scenario requires the integration of multiple complex behaviors: locomotion across uneven terrain, text-guided animation, and object interaction. 
Prior works in physics-based simulation has addressed these challenges by developing specialized controllers for specific tasks such as locomotion, object interaction, and VR tracking. These methods typically involve training controllers for each task \cite{rempe2023trace,hassan2023synthesizing,winkler2022questsim} or encoding atomic motions into reusable latent spaces, which are then combined by a high-level controller to perform new tasks \cite{peng2022ase,tessler2023calm,yao2022controlvae,luo2023universal}. These systems tend to lack versatility, as each new task requires the lengthy process of training new task-specific controllers. Additionally, these models often rely on meticulous handcrafted reward functions, which can be difficult to design and tune, often leading to unsolicited behaviors.

The goal of this work is to develop a versatile unified motion control model that can be conveniently reused across a wide variety of tasks, eliminating the need for task-specific training and complex reward engineering. This approach not only simplifies the training process, but also enhances the model's ability to generalize across different tasks. We propose a framework that trains a versatile control model by leveraging the rich multi-modal information within existing motion capture datasets, such as kinematic trajectories, text descriptions, and scene information.

Our proposed framework, \alg\footnote{Detailed video demonstrations are provided in the project page\\ \href{https://research.nvidia.com/labs/par/maskedmimic/}{https://research.nvidia.com/labs/par/maskedmimic/}}, trains a single unified controller capable of executing a wide range of tasks. The \alg~ model is trained on randomly \textit{masked} motion sequences. Conditioned on a masked motion sequence, \alg~ predicts actions that reproduce the original (unmasked) full-motion sequence. Once trained, this inpainting approach provides an intuitive interface for directing the behavior of the simulated character. Through a technique we call \textit{goal-engineering}, akin to prompt-engineering from natural language processing, users can provide a variety of different constraints that guide the controller to perform a desired task.
This approach offers several advantages over prior methods. Training on masked motion sequences enables the model to generalize to novel combinations of objectives. Intuitive partial constraints replace complex, error-prone reward functions, simplifying the design process (\cref{fig: castle}). Moreover, training \alg~ as a single unified model allows for positive transfer, where knowledge gained from one task enhances performance on others. For example, \alg~ outperforms prior task-specific methods in generating full-body motion from VR inputs, while also generalizing to moving across irregular terrain and novel objects. The central contributions of our work include:
\begin{enumerate}
    \item \alg, a unified physics-based character control framework. It produces full-body motions by inpainting from partial motion descriptions. These partial descriptions can include target keyframes, target joint positions/rotations, text instructions, object interactions, or any combination thereof.
    
    \item A suite of \emph{goal-engineering} techniques enabling \alg \ to reproduce tasks from prior systems, including full-body tracking \cite{wang2020unicon,luo2023perpetual}, VR tracking \cite{winkler2022questsim}, scene interaction \cite{hassan2023synthesizing,pan2023synthesizing}, terrain traversal \cite{rempe2023trace,wang2024pacer+}, text-control \cite{juravsky2022padl}, and more, within a single model.
\end{enumerate}

\section{Related Work}

\textbf{Physics-Based Character Animation:} Early approaches in physics-based animation focused on manually-designing task-specific controllers. These controllers can produce compelling results. However, they typically require a lengthy engineering process for each task of interest, and do not scale well to the diverse repertoire needed for general-purpose control \cite{FeatureBased2010,DataBiped2010,2010-TOG-sampControl,2013-TOG-MuscleBasedBipeds}. More recent work has shown how to learn these controllers to perform complex, scene aware behaviors, such as locomote \cite{rempe2023trace} or sit on objects \cite{hassan2023synthesizing}. However, these approaches typically require a manual selection of motions fitting the expected behaviors combined with delicate reward design. Compared to kinematic animation, physics enables scene-aware motions, such as object interactions \cite{hassan2023synthesizing,pan2023synthesizing,xiao2024unified} and locomotion across irregular terrain \cite{rempe2023trace,wang2024pacer+}.

Our work leverages physics-based animation to learn robust behaviors that generalize to unseen terrains and objects.

\textbf{Human Object Interaction:} Generating realistic human-object interactions (HOI) requires accurate modeling of the physical dynamics between humans and objects, particularly regarding contacts.
While kinematic-based HOI methods have made progress in areas like 3D-aware scene traversal \cite{wang2021scene,wang2022towards} and interactions\cite{zhao2023synthesizing,xu2023interdiff}, they often produce unrealistic artifacts such as penetration and floating objects.

Recent advancements in physics-based HOI methods have addressed these issues by incorporating physics simulations into the motion generation process. Notable examples include PhysHOI \cite{wang2023physhoi}, InterPhys \cite{hassan2023synthesizing}, and UniHSI \cite{xiao2024unified}, which can produce more natural and physically-plausible scene interactions. These methods leverage physics engines to ensure that the generated motions adhere to physical laws, resulting in more natural interactions.
Our work builds upon these physics-based HOI efforts by introducing a unified controller that is also capable of performing object interaction behaviors. By framing motion generation as an inpainting task from partial goals, our method, \alg, is able to interact with novel scene compositions, such as placing furniture on irregular terrain, extending the applicability of HOI systems to more diverse and complex scenarios.

\textbf{Text to motion:} Text provides a high-level interface for controlling virtual characters. The availability of large text-labeled motion datasets, such as BABEL \cite{BABEL:CVPR:2021} and HumanML3D \cite{Guo_2022_CVPR}, have enabled the development of text-conditioned motion models. Initial results focused on kinematic animation, with methods such as ACTOR \cite{petrovich21actor} and MDM \cite{tevet2023human} showing promising text control capabilities. However, careful engineering of the text prompts are often necessary to elicit the desired behaviors from a model, and the resulting motions nonetheless exhibit artifacts such as floating and sliding.

PACER++ \cite{wang2024pacer+} aimed to mitigate non-physical motion artifacts by combining kinematic diffusion models with physics-simulation. A text-conditioned kinematic model is used to produce the upper-body motion, then a physics-based controller is used to follow a given path while matching the kinematically-generated upper-body motion. A parallel line of work, PADL \cite{juravsky2022padl} and SuperPADL \cite{superpadl2024}, trained physics-based controllers that can be directly conditioned on text commands.
In this work, we develop a single unified physics-based controller that can be directly conditioned on both text and kinematic constraints, without requiring a separate text-to-motion model. This enables intuitive text-based stylization of the simulated motions.

\textbf{Latent Generative Models} have emerged as a more scalable and generalizable approach to address the inefficiency of task-specific controllers. These models utilize large motion datasets to map latent codes to different behaviors. Notable prior work includes ASE, CALM, and CASE \cite{peng2022ase,tessler2023calm,dou2023c}, which used an adversarial objective, and ControlVAE, PhysicsVAE, PULSE, and NCP \cite{yao2022controlvae,won2022physics,luo2023universal,zhu2023neural}, which leverage motion tracking.

By modeling a large skill corpus, these methods remove the need for task-specific data curation. However, their learned latent representations are often abstract, lacking intuitive grounding for user control. Consequently, to solve new tasks, additional hierarchical controllers are typically trained to produce desired motions through latent control, limiting their usability \cite{peng2022ase,yao2022controlvae,luo2023universal}.

Our work, \alg, presents a unified interface by formulating character control as a motion inpainting problem over partial multi-modal constraints extracted directly from the data itself. Leveraging the VAE approach with a motion tracking objective, a single trained model supports locomotion across irregular terrain \cite{rempe2023trace}, generating full-body motion from VR controller signals \cite{winkler2022questsim,lee2023questenvsim}, inbetweening \cite{gopinath2022motion}, natural object interactions \cite{hassan2023synthesizing}, full-body tracking \cite{wang2020unicon,luo2023perpetual}, and more.

\textbf{Motion Inpainting:} Generating full-body motions from partial joint constraints is commonly called motion inpainting. Inpainting is a fundamental problem in character animation with notable applications like motion inbetweening and VR tracking. In the task of inpainting, a characters full-body motion must be inferred from a sparse set of available sensors or keyframes.

Prior work has explored inpainting for kinematic systems using autoregressive models \cite{huang2018deep,yang2021lobstr,zheng2023realistic}, variational inference \cite{dittadi2021full}, and diffusion\cite{tevet2022human,du2023avatars,xie2023omnicontrol}. However, while sparse tracking models have been proposed for physically animated systems, they are specialized for fixed sparsity patterns. For example, \citet{lee2023questenvsim} proposed a system for handling joint-sparsity, with fixed pre-defined joints, such as those obtained from VR systems.

In contrast, our unified physics-based system \alg~ supports flexible sparsity patterns. With \alg, any combination of modalities (joints, keyframes, objects, text) can be observed or unobserved. This enables various applications, from unconditional generation to multi-modal constraints like inbetweening or VR avatar control across irregular terrain, and object interactions. 

Previous kinematic inpainting methods struggle with such scenarios as they lack the physical grounding to reason about dynamics, contact, and multi-body interactions. In contrast, \alg's physics-based formulation allows seamless transitions across modes while ensuring plausible motions that obey physical laws. The user can intuitively specify high-level multi-modal constraints, and \alg~ automatically synthesizes the corresponding physically plausible motions. This presents an expressive multi-modal control interface.

\section{Preliminaries}
Our framework consists of two stages. In the first stage, we train a motion-tracking controller on a large dataset of motion clips using reinforcement learning. Then, we distill that  controller into a versatile multi-modal controller using behavior cloning. We now review the fundamental concepts and notations behind our framework.

\subsection{Reinforcement Learning}
The first stage of our approach leverages the framework of \emph{goal-conditioned reinforcement learning} (GCRL) to train a versatile motion controller that can be directed to perform a large variety of tasks. In this framework, an RL agent interacts with an environment according to a policy $\pi$. At each step $t$, the agent observes a state $s_t$ and a future goal $g_t$. The agent then samples an action $a_t$ from the policy $a_t \sim \pi (a_t | s_t, g_t)$. After applying the action, the environment transitions to a new state $s_{t+1}$ according to the environment dynamics $p(s_{t+1} | s_t, a_t)$, and the agent receives a reward $r_t = r(s_t, a_t, s_{t+1}, g_t)$. The agent's objective is to learn a policy that maximizes the discounted cumulative reward:
\begin{equation}
    J = \mathbb{E}_{p(\tau|\pi)} \left[ \sum_{t=0}^T \gamma^t r_t \right] \,,
\end{equation}
where $p(\tau|\pi) = p(s_0) \Pi_{t=0}^{T-1} p(s_{t+1}|s_t,a_t) \pi(a_t | s_t, g_t)$ is the likelihood of a trajectory $\tau = (s_0, a_0, r_0, \ldots, s_{T-1}, a_{T-1}, r_{T-1}, s_T)$. The discount factor $\gamma \in [0, 1)$ determines the effective horizon of the policy.

\subsection{Behavioral Cloning}

The second stage of our approach leverages behavioral cloning (BC) to distill a teacher policy $\pi^*$, trained through RL, into a more versatile student policy $\pi$, which can be directed through multi-modal inputs. The policy distillation process is performed using the DAgger method \citep{ross2011reduction}. In this online-distillation process, trajectories are collected by executing the student policy and then relabeled with actions from the teacher policy:
\begin{equation}
    \mathop{\mathrm{arg \ max}}_{\pi} \ \mathbb{E}_{(s, g) \sim p(s, g | \pi)} \mathbb{E}_{a \sim \pi^*(a | s, g)} \left[ \mathrm{log} \pi(a | s, g) \right].
\end{equation}
$p(s, g | \pi)$ denotes the distribution of states and goals observed under the student policy. This form of active behavioral cloning mitigates drift inherent in supervised distillation methods \citep{ross2011reduction}.

\section{System Overview}

\begin{figure}
    \centering
    \includegraphics[width=0.99\linewidth]{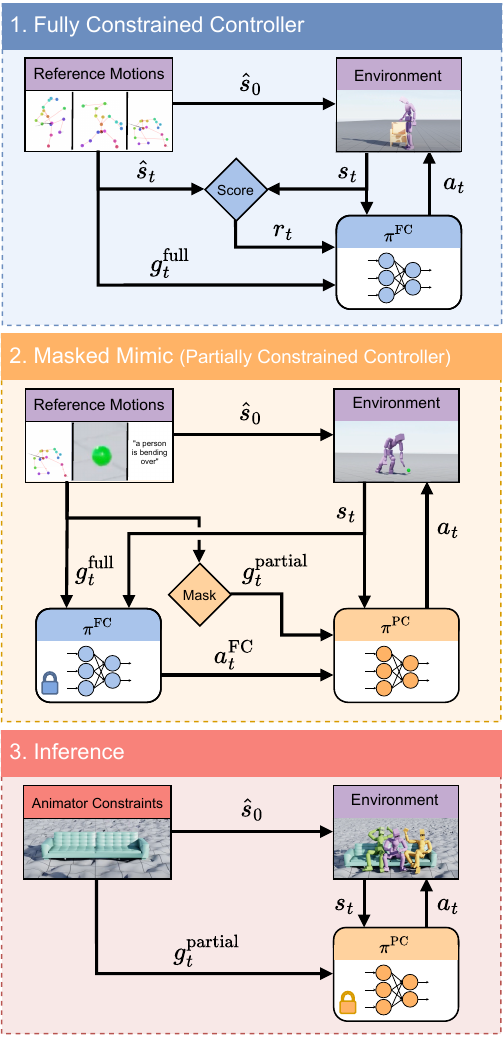}
    \caption{\textbf{The \alg~ framework:} The first phase produces a \textit{fully-constrained controller} $\pi^\text{FC}$. This full-body tracker is trained using reinforcement learning to imitate kinematic motion recordings across a wide range of complex scene-aware contexts. The second phase produces \alg. Treating $\pi^\text{FC}$ as a teacher, through supervised limitation learning its knowledge is distilled into a \textit{partially-constrained controller} $\pi^\text{PC}$. As $\pi^\text{PC}$ observes masked inputs, this process enables it to perform physics-based inpainting. Finally, at inference, without any further training, $\pi^\text{PC}$ is used to generate novel motions, in previously unseen scenes, from partial goals provided by the user.}
    \label{fig:overview}
\end{figure}

This work introduces a versatile controller for physics-based character animation. We aim to develop a scalable system that can learn a rich repertoire of behaviors from large and diverse multi-modal datasets. Our framework, illustrated in \cref{fig:overview}, supports multiple control modalities, providing users with a flexible and intuitive interface for directing the behavior of simulated characters. Our framework consists of two stages. First, we train a fully-constrained motion tracking controller on a large mocap dataset. This controller's inputs consist of the full-body target trajectories of a desired motion. The fully-constrained controller is trained to imitate a wide variety of motions, including those involving irregular terrains and object interactions. Next, this fully-constrained controller is distilled into a more versatile partially-constrained controller. This partially constrained controller can be directed via diverse control inputs. The versatility of the partially-constrained controller arises from a masked training scheme. During training, the controller is tasked with reconstructing a target full-body motion given randomly masked inputs. This process enables the partially-constrained model to generate full-body motion from arbitrary partial constraints.

\paragraph{Stage 1: Fully-Constrained Controller}

The goal of physics-based motion tracking is to generate controls (such as motor actuations), which enable a simulated character to produce a motion $\{q_t\}$ that closely resembles a kinematic target motion $\{\hat{q}_t\}$ \cite{MPCSilva2008,BipedLee2010,peng2018deepmimic,wang2020unicon}. We represent a motion as a sequence of poses $q_t$, where each pose $q_t = (p_t, \theta_t)$ is encoded with a redundant representation consisting of the the 3D cartesian positions of a character's $J$ joints $p_t = (p^0_t, p^1_t, ..., p^J_t)$ and their rotations $\theta_t = (\theta^0_t, \theta^1_t, ..., \theta^J_t)$.
To successfully track a reference motion, controllers are typically provided with information that describes the motion it should imitate. For example, motion tracking controllers are commonly conditioned on the target future poses $\hat{q}_t$ \cite{wang2020unicon,luo2023perpetual}. We will refer to target poses as fully-constrained goals $g_t^\text{full}$, since the future poses provide complete information about the target motion the character should imitate.

\paragraph{Stage 2: Partially-Constrained Controller}
In this work, we propose a control model that extends beyond only conditioning on the full target poses to more versatile partially observable goals. For example, a typical problem in VR is to generate full-body motion from only head and hands sensors.
Similarly, in some cases a controller may only observe an object (e.g., a chair) and will then be required to generate realistic full-body motions that interact with the target object \cite{hassan2023synthesizing,pan2023synthesizing}.
Throughout the paper, we will refer to partially observable goals as $g_t^\text{partial}$. These partial goals specify only some elements of a desired motion. To train a versatile controller that can be directed using partial goals, we propose a simple training scheme that trains the controller on randomly masked observations of target motions. These masked observations are constructed using a random masking function $\mathcal{M} : g_t^\text{partial} = \mathcal{M}(g_t^\text{full})$.

\section{Fully-Constrained Controller}
In the first stage of our framework, we train a fully-constrained motion tracking controller $\pi^\text{FC}$ using reinforcement learning. This controller can imitate a large library of reference motions across irregular environments and interact with objects when appropriate. Since the motion dataset only consist of kinematic motion clips, the primary purpose of $\pi^\text{FC}$ is to estimate the actions (motor actuations) required to control the simulated character. $\pi^\text{FC}$ then provides the foundations that greatly simplifies the training process of a more versatile controller in the subsequent stage.

\subsection{Model Representation}

Our fully-constrained controller is trained end-to-end to imitate target motions by conditioning on the full-body motion sequence and observations of the surrounding environment, such as the terrain and object heightmaps. The training objective is formulated as a motion-tracking reward and optimized using reinforcement learning \cite{mnih2016asynchronous,peng2018deepmimic}. In this section, we detail the design of various components of the model.

\paragraph{Character Observations:} At each step, $\pi^\text{FC}$ observes the current humanoid state $s_t$, consisting of the 3D body pose and velocity, canonicalized with respect to the character's local coordinate frame:
\begin{equation}\label{eqn: target pose rep}
    s_t = (\theta_t \ominus \theta_t^\text{root}, (p_t - p_t^\text{root}) \ominus \theta_t^\text{root}, v_t \ominus \theta_t^\text{root}) \,,
\end{equation}
where $\ominus$ denotes the quaternion difference between two rotations. In addition to the current state of the character, the policy also observes the next $K$ target poses from the reference motion $g_t^\text{FC} = [ \hat{f}_{t+1}, \ldots, \hat{f}_{t+K} ]$. The features for each joint $\hat{f}^j_t$ are canonicalized both relative to the current root, and relative to the current respective joint:
\begin{equation}
    \hat{f}^j = (\hat{\theta}^j \ominus \theta^j_{t}, \hat{\theta}^j \ominus \theta^\text{root}_{t}, (\hat{p}^j - p^j_{t}) \ominus \theta^\text{root}_{t}, (\hat{p}^j - p^\text{root}_{t}) \ominus \theta^\text{root}_{t}) \,.
\end{equation}
The features for each target pose $\hat{q}_{t+k}$ are also augmented with the time $\tau_{t+k}$ from the current timestep to the target pose, resulting in the following representation: $\hat{f}_{t+k} = \{\hat{f}_{t+k}^1, \ldots, \hat{f}_{t+k}^J, \tau_{t+k} \}$.

\paragraph{Scene Observations} To imitate motions on irregular terrain, we canonicalize the character's pose with respect to the height of the terrain under the character's root (i.e. pelvis). During training, the controller is provided with a heightmap of the surrounding environment, with the heighmap oriented along the root's facing direction \cite{rempe2023trace,pan2023synthesizing}. The heightmap has a fixed resolution, and records the height of the nearby terrain geometry and object surfaces.

\paragraph{Actions:} Similar to prior work \cite{peng2018deepmimic,tessler2023calm}, we opt for proportional derivative (PD) control. We do not utilize residual forces \cite{yuan2020residual,luo2021dynamics,zhang2023learning} or residual control \cite{luo2022embodied}. The policy's action distribution $\pi_\text{FC} (a_t | s_t, g_t^\text{full})$ is represented using a multi-dimensional Gaussian with a fixed diagonal covariance matrix $\sigma^\pi = \exp(-2.9)$.

\subsection{Model Architecture} 

Motion tracking is a sequence modeling problem. The objective is to predict the next actions based on the current character state, surrounding terrain, and a sequence of future target poses. Inspired by the success of transformers in natural language processing, we tokenize each of the inputs and design $\pi^\text{FC}$ as a transformer-based controller. This choice of architecture allows the controller to attend to relevant information across the input sequence and capture the dependencies between the various input tokens.

To further enhance the learning process, we employ a critic network alongside the transformer-based controller. The critic is implemented as a fully connected network that estimates the value function. This provides a learning signal to guide the controller towards optimal actions.

Once trained, this fully-conditioned controller provides the foundation for our unified character controller. In the following section, we introduce our physics-based motion inpainting approach. This allows users to specify partial or sparse motion constraints, such as keyframes or high-level goals, and synthesize complete motion sequences that satisfy these constraints while remaining consistent with the scene context.

\subsection{Reward Function} 
The reward $r_t$ encourages the character to track a reference motion by minimizing the difference between the state of the simulated character and the target motion: 
\begin{equation}
r_t = w^\text{gp} r_t^\text{gp}  + w^\text{gr} r_t^\text{gr} + w^\text{rh} r_t^\text{rh} + w^\text{jv} r_t^\text{jv} + w^\text{jav} r_t^\text{jav} + w^\text{eg} r_t^\text{eg} ,
\end{equation}
where $r_t^{\{ \cdot \}}$ denote various reward components and and $w^{\{ \cdot \}}$ are their respective weights. The terms in the reward function encourages the character to imitate the reference motion's global joint positions (gp), global joint rotations (gr), root height (rh), joint velocities (jv), joint angular velocities (jav), as well as an energy penalty (eg) to encourage smoother and less jittery motions \cite{lee2023questenvsim}. A more detailed description of the reward function is provided in the supplementary material.

\subsection{Training Playground}

\begin{figure}
    \centering
    \includegraphics[width=\linewidth]{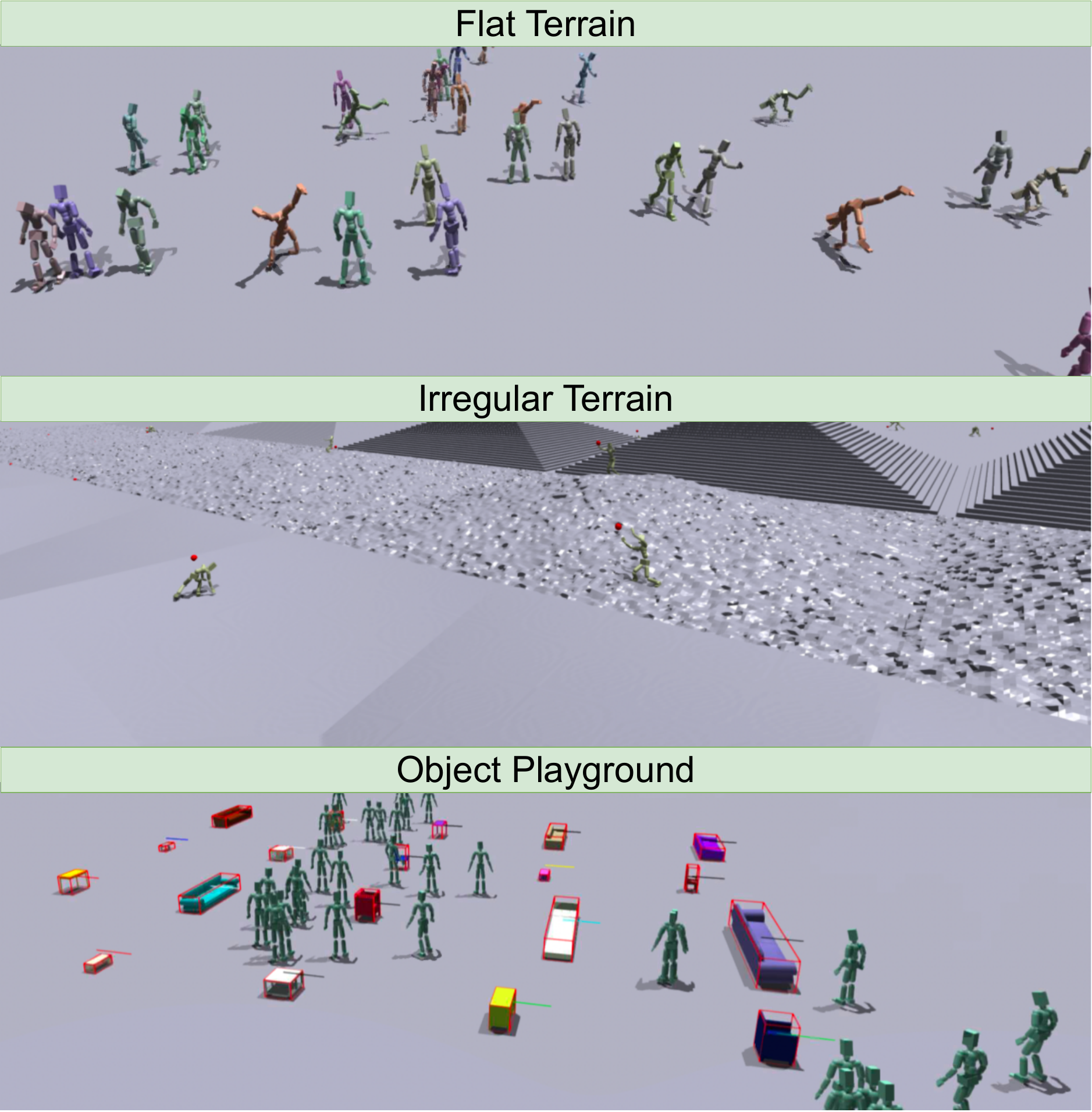}
    \caption{\textbf{Training scene} (screenshot)\textbf{:} The top region consists of standard flat terrain, enabling the controller to reproduce the original motions in a setting that best represents how they were recorded. The central region contains irregular terrain with stairs, slopes, and rough surfaces, allowing the controller to learn robust motion skills on varied ground geometries. The bottom region is reserved exclusively for object interactions, ensuring that the agent can practice interacting with objects in a clean and reproducible setup without interference from irregular terrain features.}
    \label{fig: playground}
\end{figure}

To train a controller that can operate in more complex irregular scenes, we construct a training environment, shown in \cref{fig: playground}, composed of three distinct regions: (1) flat terrain, (2) irregular terrain, and (3) object playground.

\paragraph{Flat Terrain} First, the flat terrain region is a simple environment where the model can focus primarily on imitating the reference motions, as most of the training data was recorded on flat ground. This region is a baseline for evaluating the model's ability to imitate motions in a simple, unobstructed setting.

\paragraph{Irregular Terrain} The irregular terrain region contains a wide variety of irregular terrain features, including stairs, rough gravel-like terrain, and slopes (both smooth and rough). When the agent is imitating a motion that does not involve object interactions, it can be spawned at any random location within flat and irregular terrain regions. This setup exposes the model to diverse terrain conditions, allowing it to learn robust locomotion skills that can accommodate different types of terrains.

\paragraph{Object Playground} Finally, the object playground region is reserved for object interaction motions. This region consists of various objects placed on flat ground, such as chairs, tables, and couches. Characters are only initialized in this region when they are imitating motions that involve object interactions.

\subsection{Early Termination and Prioritized Motion Sampling}
To improve the success rate on rare and more complex motions, we perform early termination \cite{peng2018deepmimic,luo2023universal}. Motions performed on flat terrain, are terminated once any joint position deviates by more than 0.25 meters. On irregular terrains, an episode is terminated when a joint error exceeds 0.5 meters, providing the controller more flexibility to adapt the original reference motion to a new environment. Furthermore, we prioritize training on motions with a higher failure rate \cite{luo2022universal,zhu2023neural}. As some motions are not expected to succeed in all scenarios (e.g., front-flip or cartwheel up a flight of stairs), the prioritized sampling only considers failures that occurred on flat terrain. The probability of prioritizing a motion $m_i$ is proportional to the probability of failing on that motion, clipped to a minimal weight of $3e^{-3}$.
This adaptive sampling strategy is vital to ensure that the agent collects a sufficient amount of data to reproduce more dynamic and challenging behaviors.

\section{Versatile Partially-Constrained Controller}\label{sec: sparsemimic}

Once the fully-constrained motion tracking model has been trained, it is then used to train a versatile partially-constrained model, denoted by $\pi^\text{PC}$. The training and inference process are illustrated in \cref{fig:overview}.
Given partial constraints, such as target positions for joints, text commands, or object locations, \alg~ generates diverse full-body motions that satisfy those constraints. $\pi^\text{PC}$ is trained to model the distribution of actions $\pi^\text{FC}(a_t | g_t^\text{full}, s_t)$
predicted by the fully-constrained controller $\pi^\text{FC}$, while only observing partial constraints $g_t^\text{partial}$. The partial constraints then provide users a versatile and convenient interface for directing $\pi^\text{PC}$ to perform new tasks, without requiring task-specific training.

\subsection{Partial Goals}

The objective of $\pi^\text{PC}$ is to produce motions that conform to constraints specified by partial goals, akin to the task of motion inpainting. In this work, we consider the following types of goals:
\begin{enumerate}
    \item \textit{Any-joint-any-time:} The model should support conditioning on target positions and rotations for any joint in arbitrary future timesteps.
    \item \textit{Text-to-motion:} The model should support high-level text commands, enabling more intuitive and expressive direction of the character's movements. 
    \item \textit{Objects:} When available, the model should support object-based goals, such as interacting with furniture.
\end{enumerate}
To produce a desired behavior, our model will support simultaneous conditioning on one or more of the aforementioned goals. For example, path following with raised arms can be achieved by conditioning the controller on a target root trajectory and a text command ``walking while raising your hands". This flexibility allows for a wide range of complex and expressive motions to be generated from concise partial specifications.

To train $\pi^\text{PC}$, flexible goals are extracted procedurally from mocap data by applying random masking. During training, $\pi^\text{PC}$ is trained to imitate the original full (unmasked) target motion by predicting the actions of the fully-constrained controller, which observes the ground-truth full target motion.

\begin{figure*}
    \centering
    \begin{subfigure}[b]{\textwidth}
        \centering
        \includegraphics[width=\linewidth]{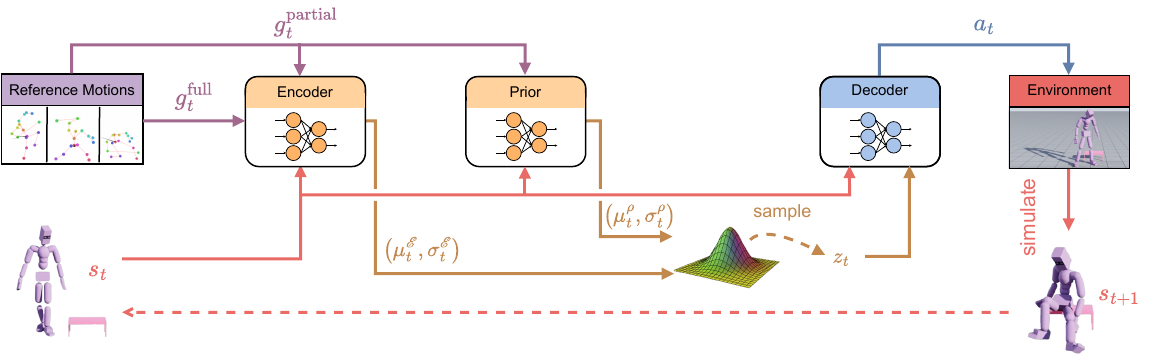}
        \caption{\textbf{System overview:} \alg~ is modeled as a VAE with a learned prior. The prior observes the partial goals, whereas the encoder, used only during training, observes both the full target pose and the partial objectives. During training, the encoder acts as a residual to the prior. It learns to provide an offset, in the latent space, towards the precise requested motion. At inference, the encoder is no longer used and the solutions are sampled directly from the prior.}
        \label{fig: vae}
     \end{subfigure}
     \begin{subfigure}[b]{\textwidth}
         \centering
         \includegraphics[width=\textwidth]{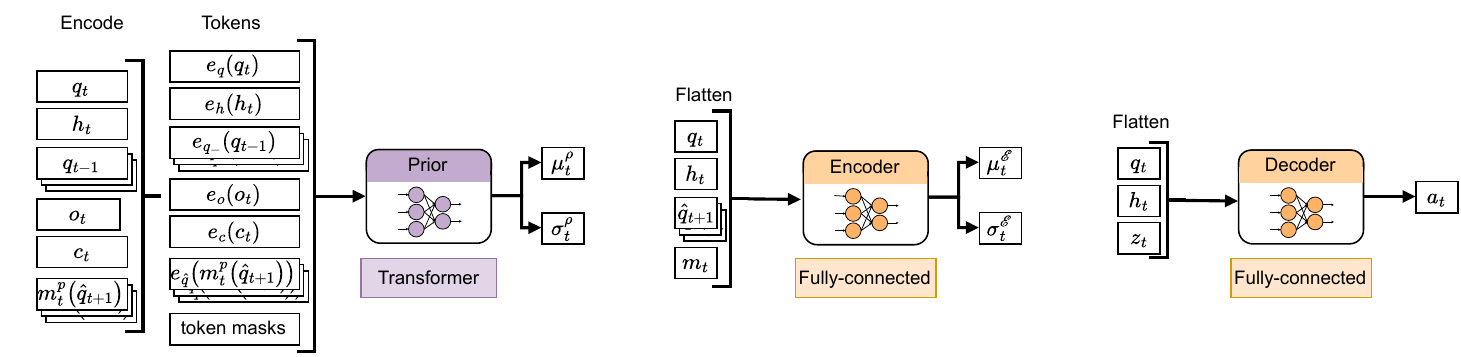}
         \caption{\textbf{Detailed view:} During training, features are extracted and masked from ground-truth motion sequences. The prior, a \textit{transformer} network, observes the current pose $q_t$, surrounding heightmap $h_t$, past poses $\{q_{t-\tau}\}$, object representation $o_t$, text command $c_t$, and target future poses $\{\hat{q}_{t+\tau}\}$. Each input modality is tokenized (encoded) using a modality-specific encoder $e_i( \cdot )$. Future poses are masked before encoding, ensuring the encoder only observes the conditionable joints. These tokens and the token masks are then provided to the prior (transformer). The token masks prevent the transformer from attending to unspecified inputs, such as a keyframe without any target joints, or a sequence without text or object conditioning. The encoder and decoder are modeled as fully-connected networks, and observe a flattened concatenation of the input features.}
         \label{fig: vae maskedmimic}
     \end{subfigure}
     \caption{\textbf{\alg~ VAE Architecture.}}
     \label{fig: maskedmimic system view}
\end{figure*}

\subsection{Modeling Diversity with Conditional VAEs}

Partial goals are an underspecified problem, as there may be multiple plausible motions that can satisfy a given set of partial goals. For example, when conditioned on reaching a target location within 1 second, there are a large variety of motions that can achieve this goal. To address this ambiguity, we model $\pi^\text{PC}$ as a conditional variational autoencoder (C-VAE). This generative model enables the $\pi^\text{PC}$ to model the distribution of different behaviors that satisfy a particular set of constraints, rather than simply producing a single deterministic behavior. By sampling from this learned distribution, the model can generate a variety of realistic and physically-plausible motions that adhere to the specified partial goals, while still allowing for natural variations and adaptability to different contexts.

\alg~ consists of 3 components: a learnable prior $\rho$, an encoder $\mathcal{E}$, and a decoder $\mathcal{D}$.
The encoder $\mathcal{E} (z_t | s_t, g_t^\text{full})$ outputs a latent distribution given the fully-observable future target poses from the desired reference motion. The decoder $\mathcal{D} (a_t | s_t, z_t)$ is then conditioned on a latent sampled from the encoder's distribution, and produces an action for the simulated character. The final component is the learned prior $\rho (z_t | s_t, g_t^\text{partial})$. The prior is trained to match the encoder's distribution given only partially observed constraints. The learnable prior is a crucial component of \alg's design as it allows the model to generate natural motions from simple user-defined partial constraints at runtime, without requiring users to specify full target trajectories for the character to follow. The encoder is used solely for training, and is not utilized at runtime.

The prior is modeled as a Gaussian distribution over latents $z_t$, with mean $\mu^\rho$ and diagonal standard deviation matrix $\sigma^\rho$,
\begin{equation}
\rho \left(z_t | s_t, g_t^\text{partial}\right) = \mathcal{N}\left(\mu^\rho\left(s_t, g_t^\text{partial}\right), \sigma^\rho\left(s_t, g_t^\text{partial}\right)\right).
\end{equation}
The encoder is modeled as a residual to the prior \cite{yao2022controlvae},
\begin{equation}
\mathcal{E} \left(z_t \middle| s_t, g_t^\text{full}\right) = \mathcal{N}\left(\mu^\rho\left(s_t, g_t^\text{partial}\right) + \mu^\mathcal{E}\left(s_t, g_t^\text{full}\right), \sigma^\mathcal{E}\left(s_t, g_t^\text{full}\right)\right).
\end{equation}
This design ensures that the embedding from the encoder, having access to full observations of the target motion, stays close to the prior that only receives partial observations. During training the latent variables $z_t$ are sampled from the encoder. All component are trained using an objective that maximizes the log-likelihood of actions predicted by $\pi^\text{FC}$ and minimizes the KL divergence between the encoder and prior:
\begin{align*}
    &\mathbb{E}_{(s,g^\text{partial}) \sim p\left(s,g^\text{partial} \middle| \pi^\text{PC}\right)} \mathbb{E}_{a \sim \pi^\text{FC} \left(a \middle| s, g^\text{full}\right)} \mathbb{E}_{z \sim \mathcal{E}\left(z \middle| s, g^\text{full} \right)} \left[ \log \mathcal{D} (a | s, z) \right. \\
    & \qquad \left. - \alpha  D_\text{KL} \left( \mathcal{E} \left(\cdot \middle| s, g^\text{full} \right) \middle|\middle| \rho \left(\cdot \middle| s, g^\text{partial} \right) \right) \right]\,, \numberthis
\end{align*}
where $g^\text{partial}$ is constructed by applying a random masking function $\mathcal{M}$ to the original fully-observed goals: $g^\text{partial} = \mathcal{M}(g^\text{full}$). In the formulation above, $\pi^\text{PC}$ interacts with the environment, while $\pi^\text{FC}$ labels the target actions for every timestep \cite[DAgger]{ross2011reduction}.
During inference, the encoder is discarded, and latents are sampled only from the prior $\rho$.

\subsection{Training}

We incorporate a number of strategies to improve the stability and effectiveness of the resulting \alg \ model. These strategies include: structured masking, KL-scheduling, episodic latent noise, and observation history. Furthermore, during the distillation process, deterministic actions are sampled from both $\pi^\text{FC}$ and $\pi^\text{PC}$ to reduce stochasticity during data collection. Early termination is also applied during distillation to prevent $\pi^\text{PC}$ from entering states that were not observed during the training of $\pi^\text{FC}$. Since $\pi^\text{FC}$ also trains with early termination, it may not provide appropriate actions in regions it has not experienced during training.

\paragraph{Masking.} Our masking process randomly removes individual target joints, the textual description, and the scene information (when applicable) from the input goals to the model. To better ensure temporally coherent behaviors, we leverage a masking scheme that is structured through time. A randomly sampled mask in one timestep has a chance of being repeated for multiple subsequent timesteps, as opposed to randomly re-sampling the mask at each step.

We observe that randomly re-sampling the mask on each step reduces the ambiguity the model encounters during training. Therefore, the resulting model generalizes worse. This is because different joints are likely to be visible across different frames, the cross-frame information provides a less ambiguous description of the requested motion. By using a temporally consistent sampling scheme, we ensure that certain joints are observed for multiple consecutive frames, while other joints remain consistently hidden. 

To ensure the model supports high-level goals, such as text-commands and interaction with a target object, all future poses can be masked out. This structured sampling mechanism guarantees that $\pi^\text{PC}$ encounters, and learns to handle, a range of different masking patterns during training. This results in increased robustness to possible user inputs. We provide pseudo-code of our mask sampling strategy in the supplementary material.

\paragraph{KL-scheduling.} Similar to $\beta$-VAE \cite{higgins2016beta}, we initialize the KL-coeff with a low value of $0.0001$, and linearly increase its value to $0.01$ over the course of training. Starting with a low KL coefficient enables the encoder-decoder to more closely imitate $\pi^\text{FC}$. Increasing the coefficient then encourages the model to impose more structure into the learned latent space, to be more amenable to sampling from the prior at runtime.

\paragraph{Episodic latent noise.} During training, latents are sampled via the reparametrization trick. To further encourage more temporally consistent behaviors, we keep the ``noise" parameter $\epsilon \sim N(0,1)$ fixed throughout the entire episode. Therefore, in each episode $\tau$ the latent variables are sampled according to $z_t^\tau = \epsilon^\tau \sigma_t^\tau + \mu_t^\tau$, and the noise $\epsilon^\tau$ is constant throughout an episode.

\paragraph{Observation history.} When conditioning on text commands, we find that providing $\pi^\text{PC}$ with past poses is crucial for generating long coherent motions that conform to the intent of a given text command. Therefore, following \citet{superpadl2024}, we provide the prior with 5 observations subsampled from the observations in the past 40 timesteps.

\subsection{Observation Representations}

We construct a representation for each type of input modality that $\pi^\text{PC}$ can receive as input. The objective is to provide a sufficiently rich representation, that is also computationally efficient and facilitates generalization to new tasks.

\paragraph{Keyframes} A future keyframe with partially observable joints is first canonicalized to the current pose (\cref{eqn: target pose rep}). The unobserved joints are then zeroed out, and the mask is appended alongside the time to reach the target frame $\tau$ $[\hat{q}_{t+\tau} * \text{mask}_{t+\tau}, \text{mask}_{t+\tau}, \tau]$. Observations of poses from previous timesteps are represented in a similar fashion, but all the joints are observed and no masking is applied.

\paragraph{Objects} We represent objects using the positions of the 8 corners of a bounding-box, canonicalized to the character's local coordinate frame. To identify different types of objects, we also provide an index representing the object type (e.g., chair, sofa, stool).

\paragraph{Text} Each text command is encoded using XCLIP embeddings \cite{ni2022expanding}, which are trained on video-language pairs to better capture temporal relationships. By leveraging the spatio-temporal information in videos during training, the XCLIP embeddings can encode the temporal aspects of language crucial for describing motions, making them well-suited for representing text commands to be translated into character animations.

\subsection{Architecture}
To provide a unified architecture capable of processing multi-modal inputs, we model the prior $\mathcal{R}$ using a transformer-encoder. This enables variable length input tokens depending on the observable goals at each timestep. Each input modality (target pose $\hat{q}_{t+\tau}$, object bounding box $o_t$, terrain heightmap $h_t$, current pose $s_t$, text $w_t$, and historical pose $q_{t-\tau}$) has a unique encoder that is shared across all inputs of the same modality. When an input is masked out, we utilize the transformer masking mechanism to exclude the respective tokens. The output of the transformer is provided to two fully-connected layers to output the mean and log-standard deviation for the prior distribution.

Since the encoder always observes the full target frames as input, it is represented as a fully connected model, as its inputs are always a fixed size. The encoder observes the full future poses $\hat{q}_{t+\tau}$ in addition to the masking applied to the keyframes, indicating which joints are visible to the prior. In addition, it observes the current pose $s_t$ and the terrain heightmap $h_t$. Like the prior, two fully-connected output heads output the residual mean and the logstd for the encoder. Similarly, the decoder is also modeled as a fully-connected network. It observes the current state $s_t$, the sampled latent $z_t$, and the terrain heightmap $h_t$. The decoder then outputs a deterministic action $a_t$.

A high level illustration is provided in \cref{fig: vae maskedmimic}.

\section{Experimental Setup}\label{sec: experiments}

All physics simulation are performed using Isaac Gym \cite{makoviychuk2021isaac}, each with 16,384 parallel environments, split across 4 A100 GPUs. Models are trained for approximately 2 weeks, amounting to approximately 30 (10) billion steps for $\pi^\text{FC}$ ($\pi^\text{PC}$).
The controllers operate at 30 Hz, and the simulation runs at 120 Hz. 
Detailed hyperparameter settings are available in the supplementary material. When training $\pi^\text{PC}$, we train with joint conditioning for a key subset of body parts: Left Ankle, Right Ankle, Pelvis, Head, Left Hand, and Right Hand.

\subsection{Datasets}

To train a unified controller capable of being directed using different control modalities, our models are trained using an aggregation of multiple datasets that collectively provide a range of different modalities.

\paragraph{Keyframe Conditioning} The core of our data is the AMASS dataset \cite{AMASS:ICCV:2019}. AMASS contains mocap recordings for a wide range of diverse human behaviors, without scene information or text. From this dataset, we extract the joint positions, rotations, and their relative timings. This enables any-joint-any-time conditioning, where a controller can be conditioned on target positions or rotations at various future timesteps. To improve generalization to new and unseen motions, we mirror the motions (flip left-to-right) as a form of data augmentation. However, it has been observed in prior work that some motions in the AMASS dataset contain severe artifacts \cite{luo2023perpetual,luo2023universal,superpadl2024}, including non-physical artifacts such as intersecting body parts, floating, or scene interactions without objects (such as walking up a staircase that does not exist). We follow the same filtering process as PHC \cite{luo2023perpetual} to obtain a filtered dataset.

\paragraph{Text Conditioning} To enable text control, we utilize the HumanML3D dataset \cite{Guo_2022_CVPR}. HumanML3D provides a breakdown of the AMASS motion sequences into atomic behaviors, each labeled with text descriptions. This allows \alg~ to be conditioned on text commands, providing a more intuitive and expressive way to direct the character's movements. We use the mirrored-text for mirrored motions, for example, ``a person turns left" is converted to ``a person turns right".

\paragraph{Scene Interaction} To synthesize natural interactions between characters and objects, we utilize the SAMP dataset \citep{hassan2021stochastic}. The SAMP dataset contains motion clips for interacting with different categories of objects, in addition to meshes of the corresponding objects for each motion clip. To train models that are able to interact with a wider array of objects, objects are randomly sampled within the class of objects associated with each motion clip \citet{hassan2023synthesizing}. For example, multiple different armchair models can be used to train the same sitting behavior, allowing our model to be conditioned on different target objects for interaction at runtime.

\subsection{Evaluation}

To evaluate the effectiveness of our framework, we construct a benchmark consisting of common tasks introduced by prior systems. For each tasks, we report a success rate metric and an error rate metric. Both are aimed to complement one another. During evaluation, the \alg~ model is conditioned on the mean of the prior's latent distribution. Sampling latents randomly produces more diverse behaviors, however, it can also produce less-likely solutions that are more likely to fail. This results in a slight degradation of performance with respect to the raw tracking metrics.

In all the evaluations, we analyze the performance of a unified model. This model is not fine-tuned for any specific task, but instead, it is controlled through user-specified goals or goals extracted from kinematic recordings (for the motion-tracking tasks). Similarly, the qualitative results showcase motions generated from new user-generated goals or tracking new motions that were not observed during training. Qualitative results are best viewed in the supplementary video.

\paragraph{Full-body tracking} We begin by evaluating both the fully constrained controller $\pi^\text{FC}$ and partially-constrained controller \alg~ $\pi^\text{PC}$ on the task of full-body motion tracking. Given a target motion, the controllers are required to closely track the sequence of future target poses from the target motion. For this task, all features of the target future poses are fully observed by the controllers. This test establishes the baseline capability for motion generation, both in terms of success rates and tracking quality, and allows comparison to prior systems for motion tracking.

\paragraph{Joint sparsity} To evaluate the model's effectiveness for generating plausible motions from partial constraints, we first consider the task of VR tracking. In this task, the controllers no longer observe the full target poses. Instead, they are provided with the target head position and rotation, in addition to the hand positions \cite{winkler2022questsim}. Note, \alg~ models are not explicitly trained for VR tracking. We compare to the results reported in \citet{luo2023universal}, consisting of 3 baselines: PULSE \cite{luo2023universal}, ASE \cite{peng2022ase}, and CALM \cite{tessler2023calm}. In addition, we evaluate the ability of tracking motions given varying joint targets. This is made possible by \alg's support for any-joint conditioning during inference.

\paragraph{Irregular terrains} To evaluate the robustness of our models to variations in the environment, we evaluate the performance of $\pi^\text{FC}$ and $\pi^\text{PC}$ on both full-body imitation and VR-tracking when spawned on randomly generated irregular terrains. The terrain consists of rough (gravel like) ground, stairs, and slopes (rough and smooth). Similar to the previous experiments, goals are still extracted from human motion data from AMASS. The controller is therefore evaluated on its ability to closely imitate a large variety of motions while accommodating irregular terrains.

\subsection{Tasks} 
By training \alg~ on randomly masked input goals, the model learns a versatile interface that can be easily used to direct the controller to perform new tasks. Performing new tasks often requires generalization to new and unseen scenarios. 
In this section, we evaluate \alg's ability to generalize and handle user-defined goals, which the model was not explicitly trained on. To direct the model to perform new tasks, we construct simple finite-state-machines that transition between goals provided to the controller. This form of \textit{goal-engineering} (akin to prompt-engineering for language models) enables \alg~ to perform a range of new tasks, without additional task-specific training.

\begin{figure*}[t]
     \begin{subfigure}[b]{0.495\textwidth}
         \centering
         \includegraphics[trim={4cm 3cm 4cm 2cm},clip,width=0.198\textwidth]{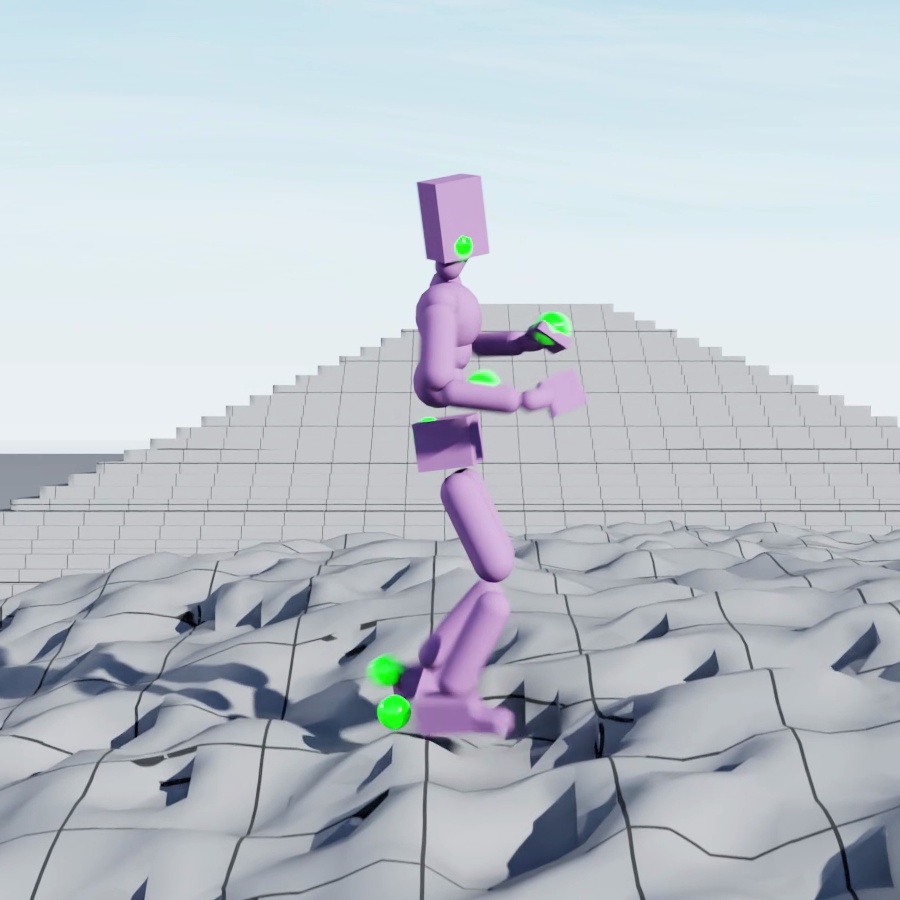}\hfill
         \includegraphics[trim={4cm 3cm 4cm 2cm},clip,width=0.198\textwidth]{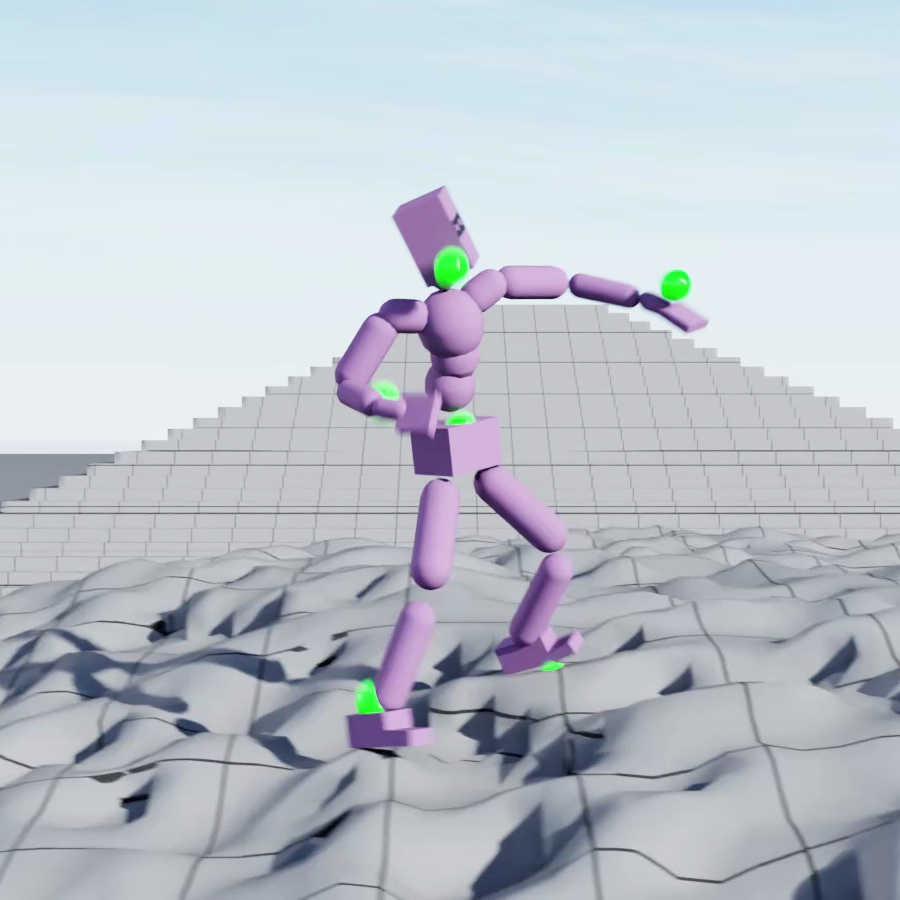}\hfill
         \includegraphics[trim={4cm 3cm 4cm 2cm},clip,width=0.198\textwidth]{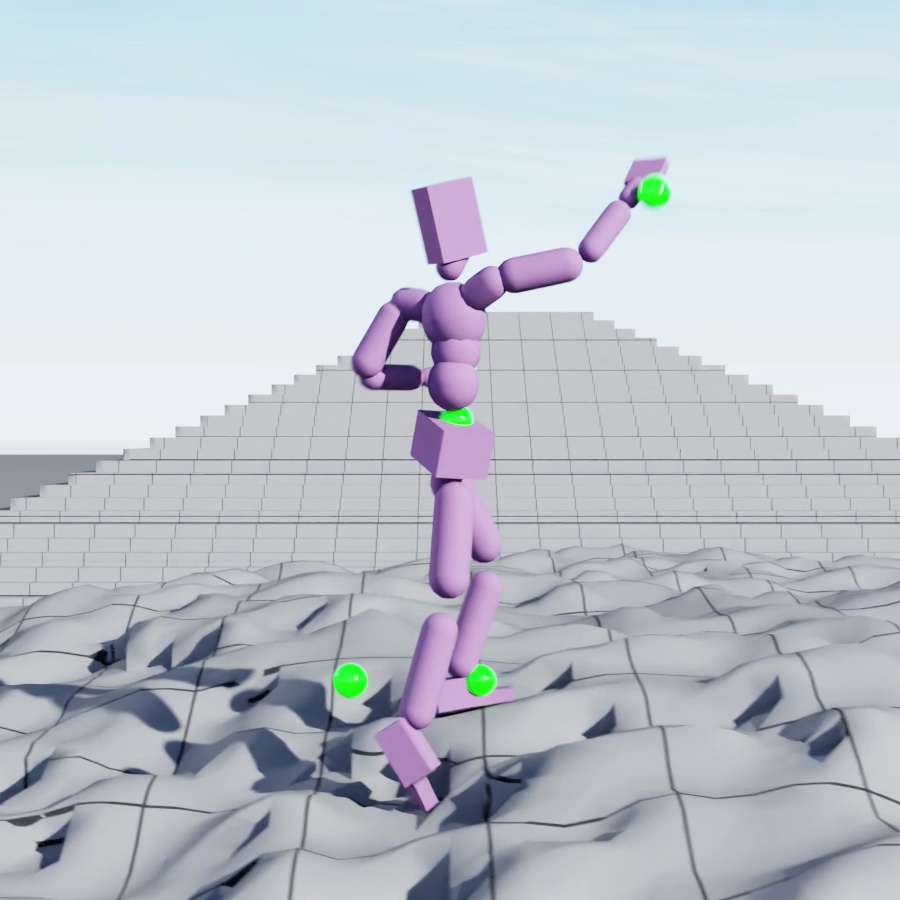}\hfill
         \includegraphics[trim={4cm 3cm 4cm 2cm},clip,width=0.198\textwidth]{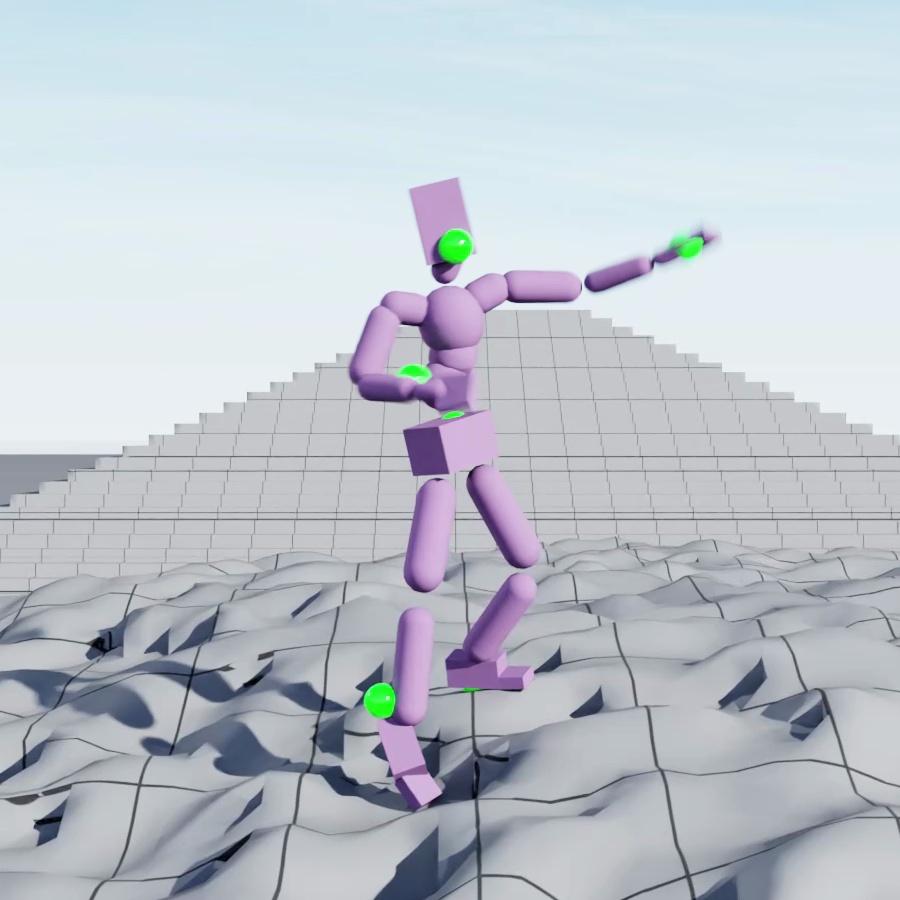}\hfill
         \includegraphics[trim={4cm 3cm 4cm 2cm},clip,width=0.198\textwidth]{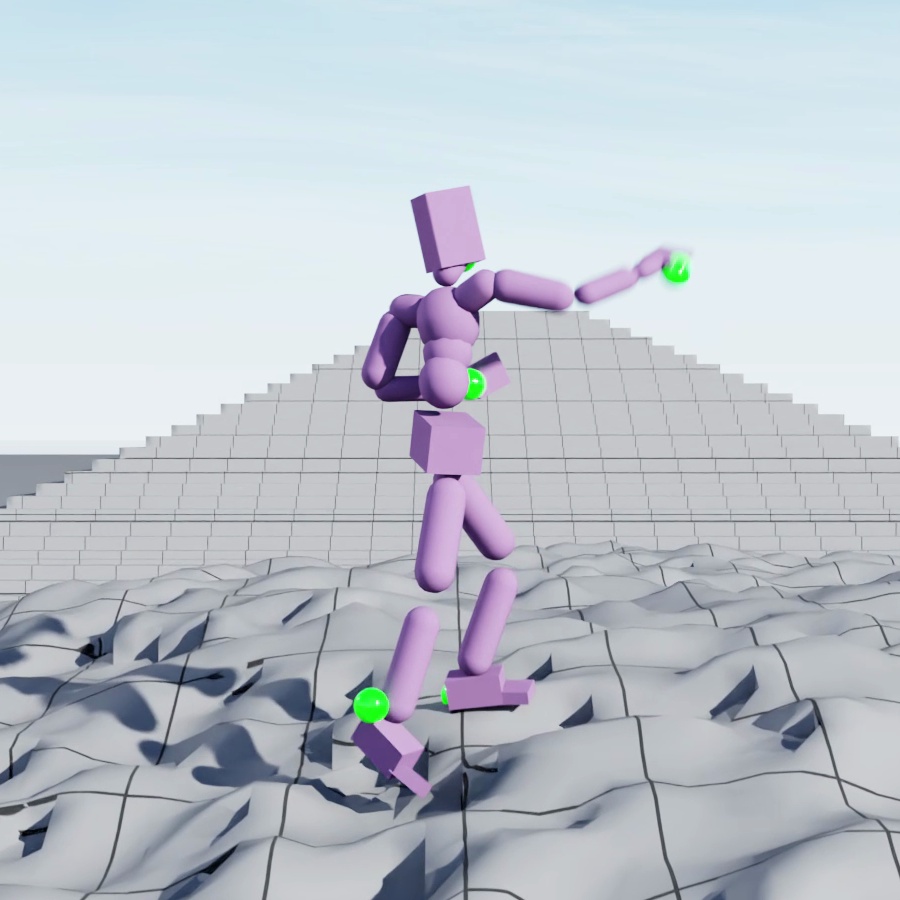}
         \caption{\textbf{Full-body tracking:} punching}
         \label{fig: fullbody punch}
     \end{subfigure}
     \begin{subfigure}[b]{0.495\textwidth}
         \centering
         \includegraphics[trim={4cm 3cm 4cm 2cm},clip,width=0.198\textwidth]{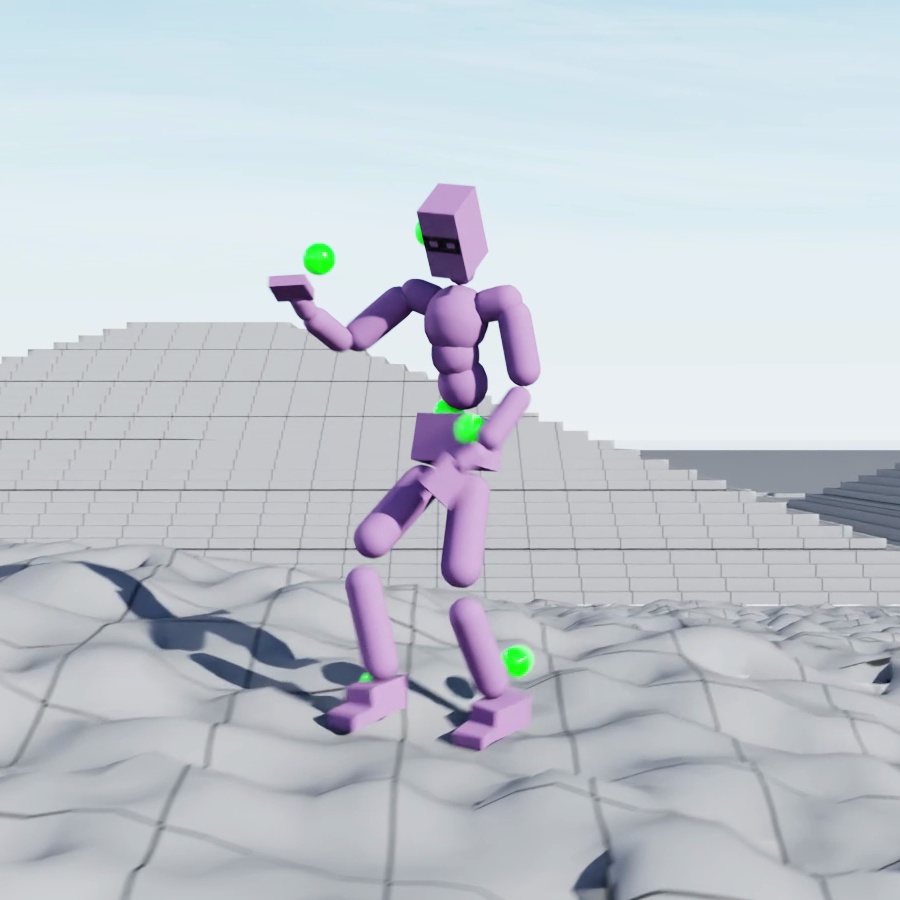}\hfill
         \includegraphics[trim={4cm 3cm 4cm 2cm},clip,width=0.198\textwidth]{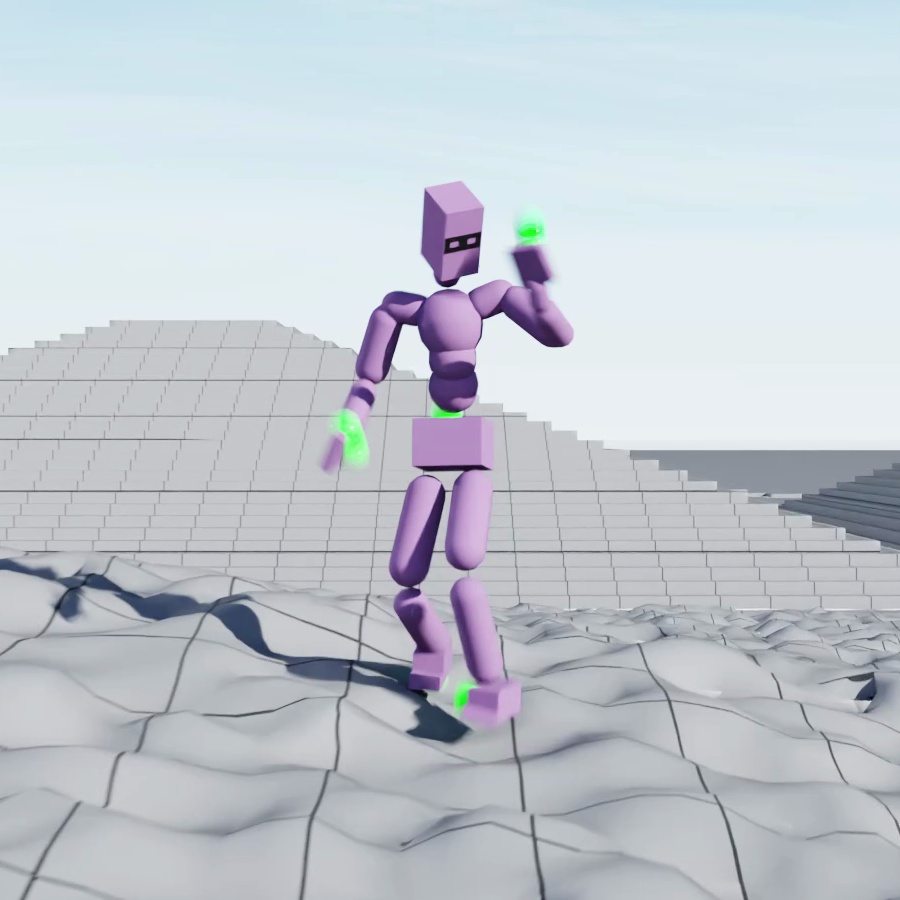}\hfill
         \includegraphics[trim={4cm 3cm 4cm 2cm},clip,width=0.198\textwidth]{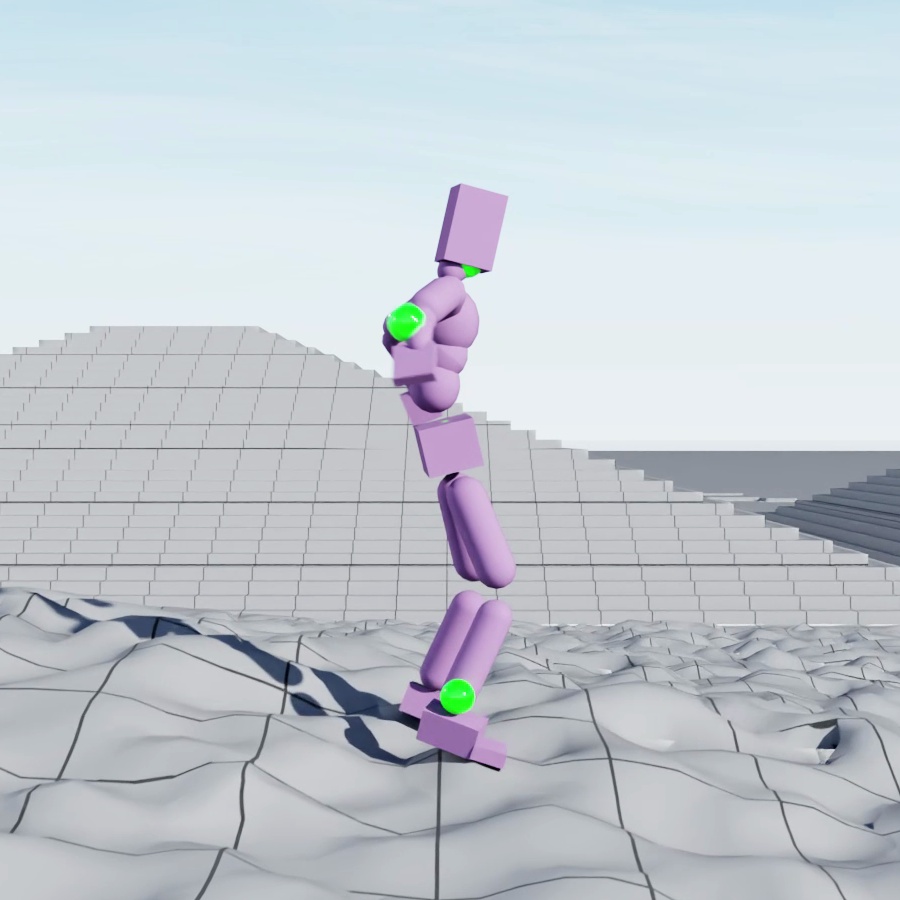}\hfill
         \includegraphics[trim={4cm 3cm 4cm 2cm},clip,width=0.198\textwidth]{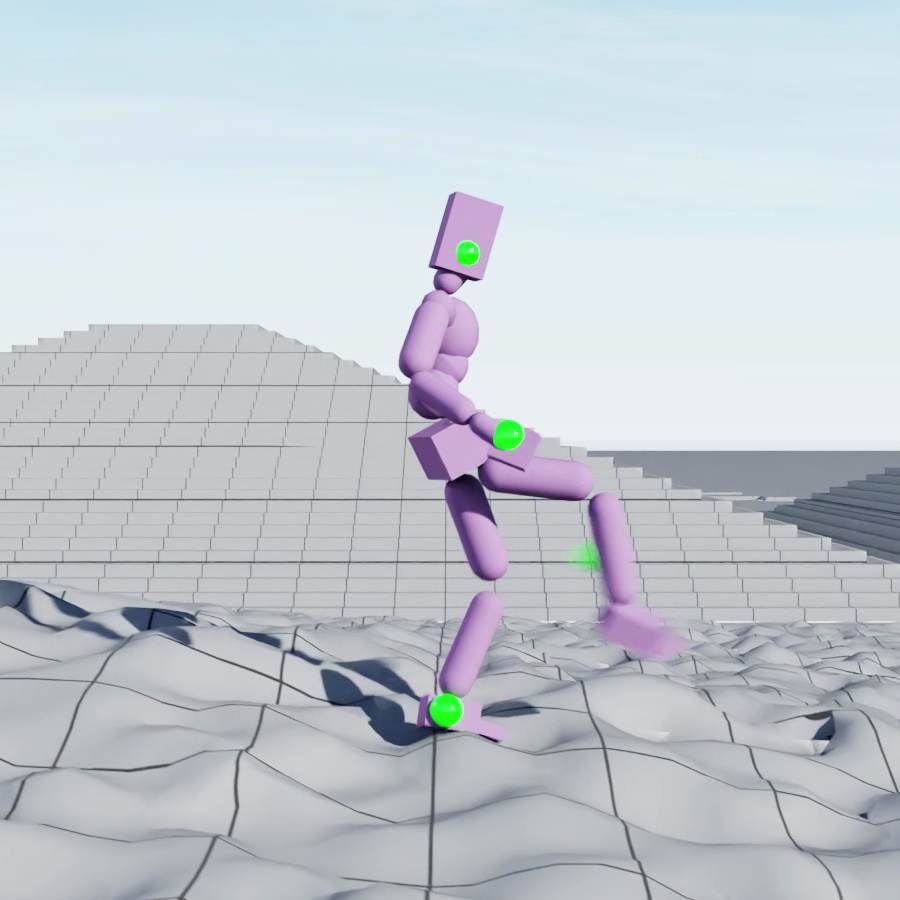}\hfill
         \includegraphics[trim={4cm 3cm 4cm 2cm},clip,width=0.198\textwidth]{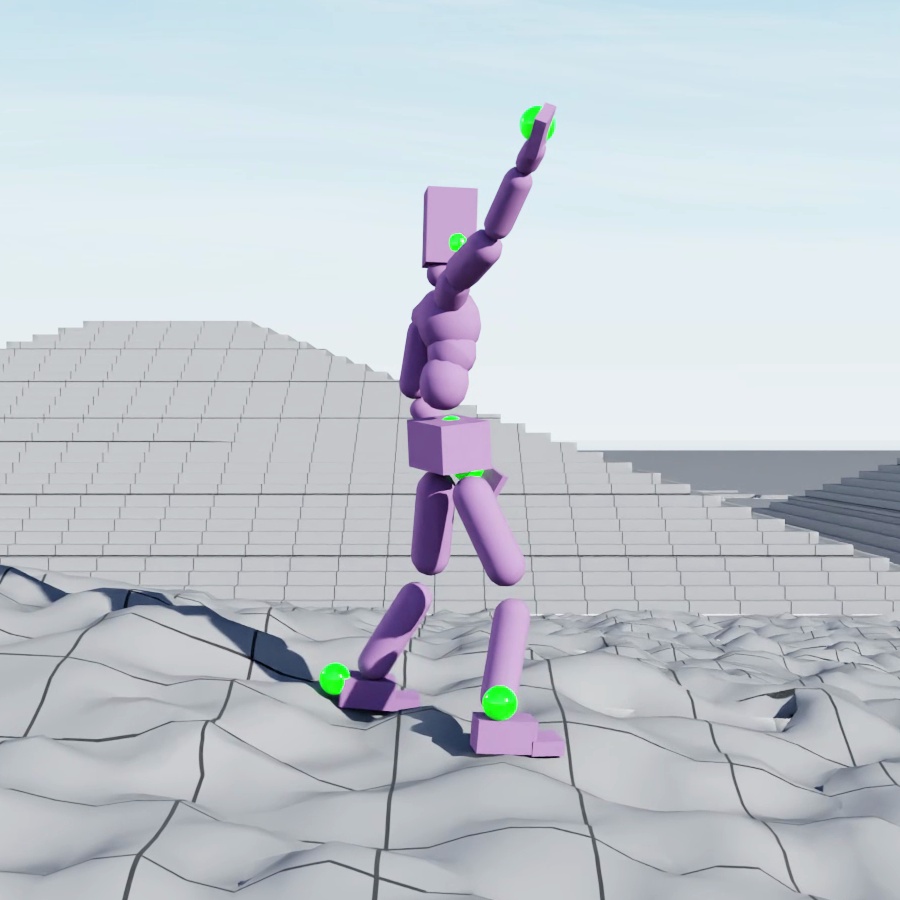}
         \caption{\textbf{Full-body tracking:} dancing}
         \label{fig: fullbody dance}
     \end{subfigure}\\
     
    \begin{subfigure}[b]{0.33\textwidth}
         \centering
         \scalebox{-1}[1]{\includegraphics[trim={12.5cm 8cm 0cm 8cm},clip,width=0.99\textwidth]{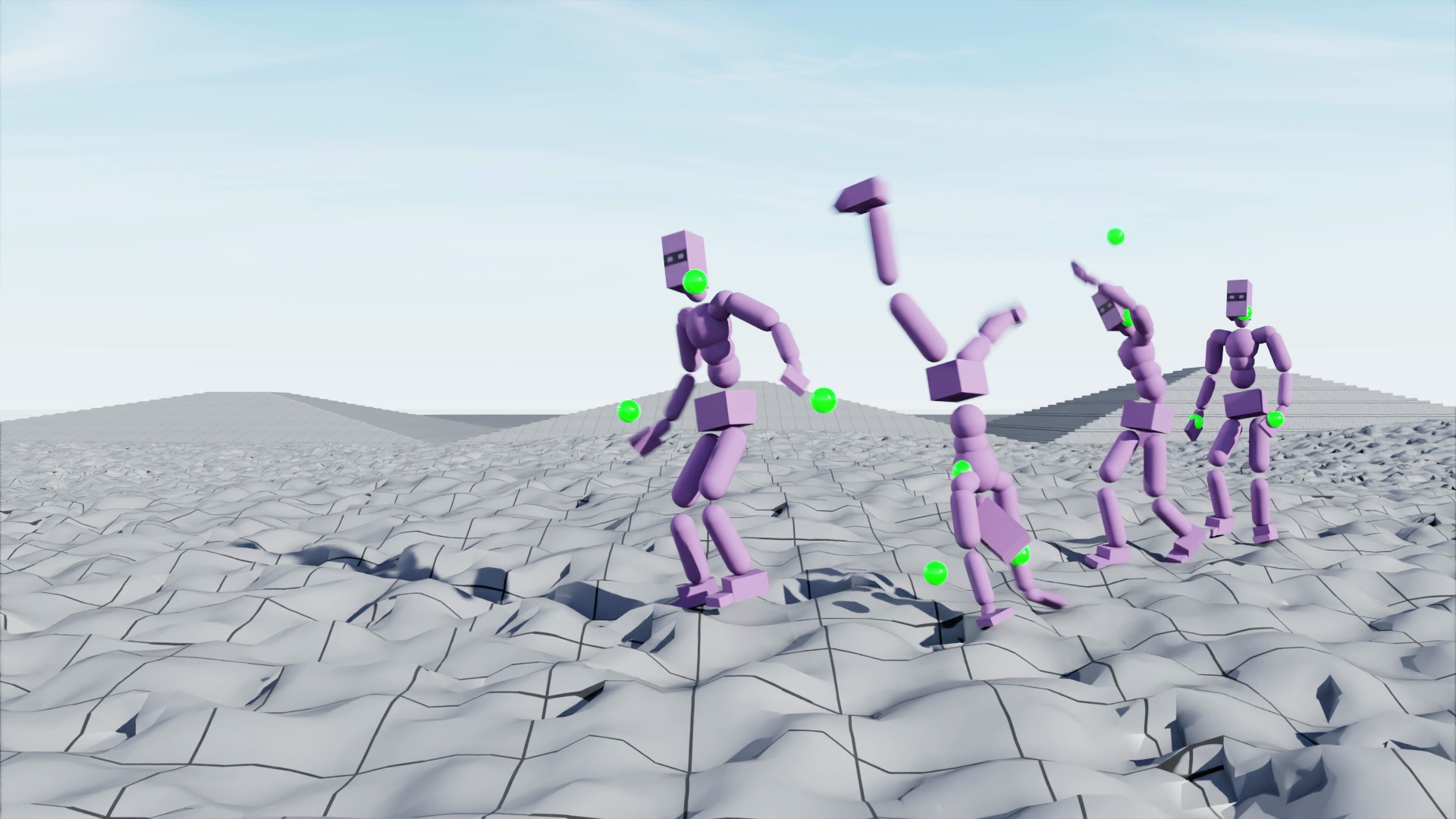}}
         \caption{\textbf{VR tracking:} cartwheel}
         \label{fig: vr cartwheel}
     \end{subfigure}
     \begin{subfigure}[b]{0.66\textwidth}
         \centering
         \includegraphics[width=0.198\textwidth]{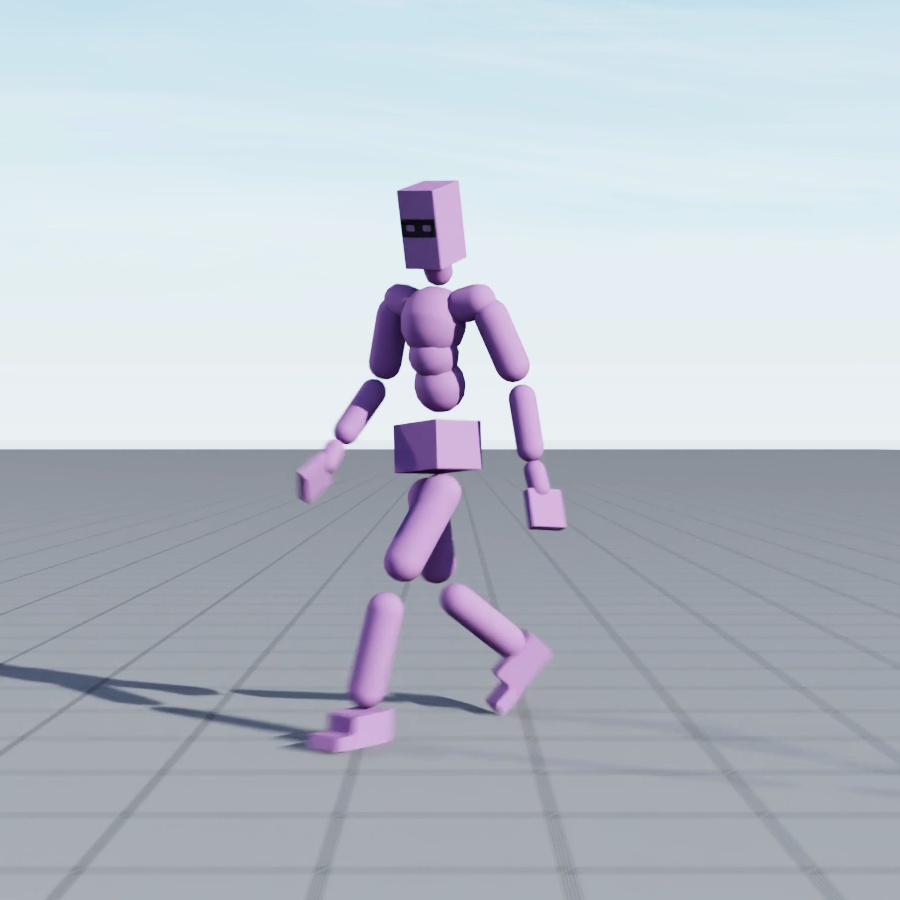}\hfill
         \includegraphics[width=0.198\textwidth]{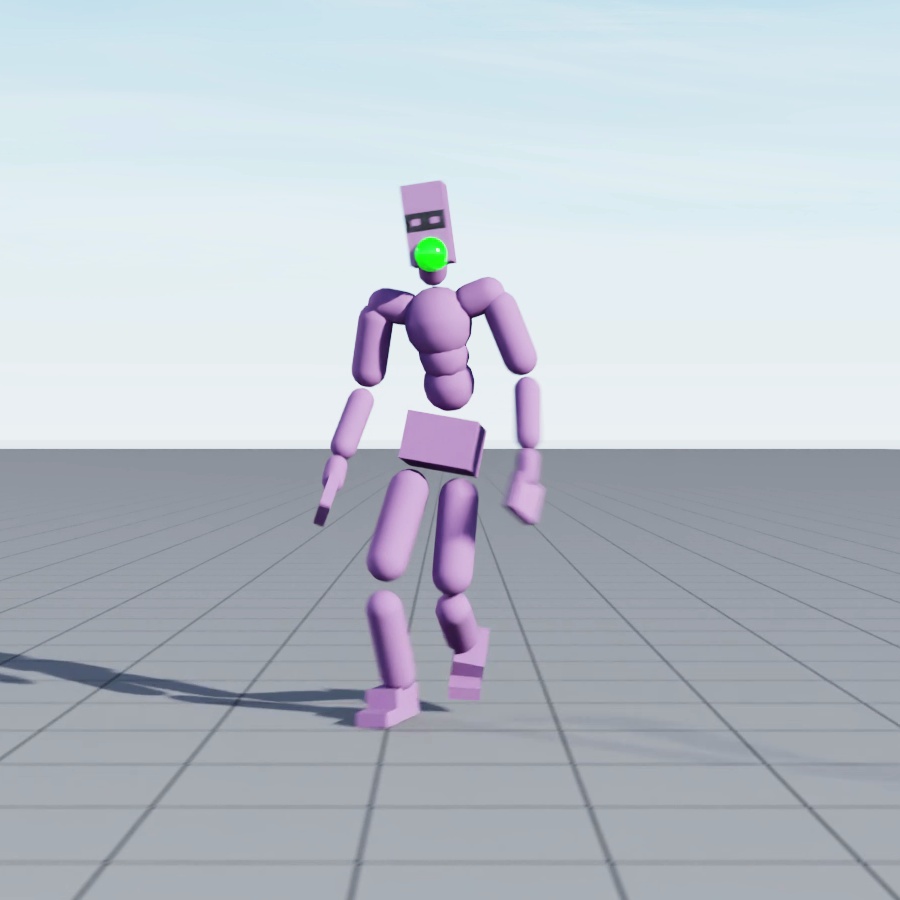}\hfill
         \includegraphics[width=0.198\textwidth]{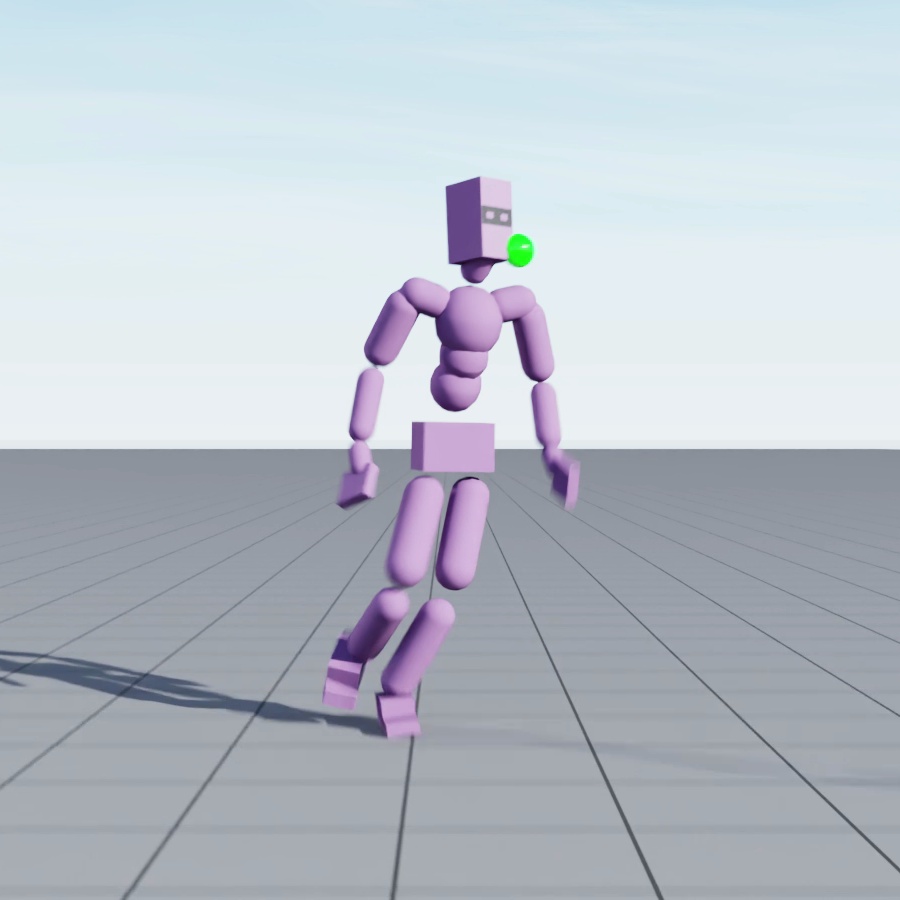}\hfill
         \includegraphics[width=0.198\textwidth]{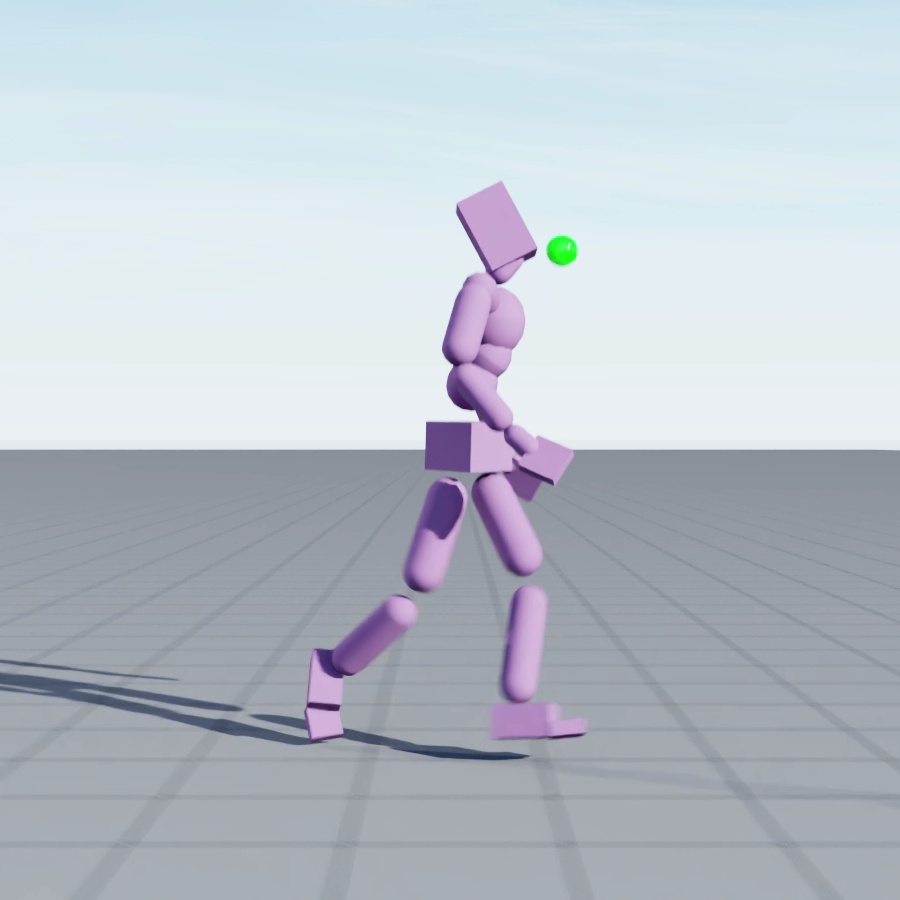}\hfill
         \includegraphics[width=0.198\textwidth]{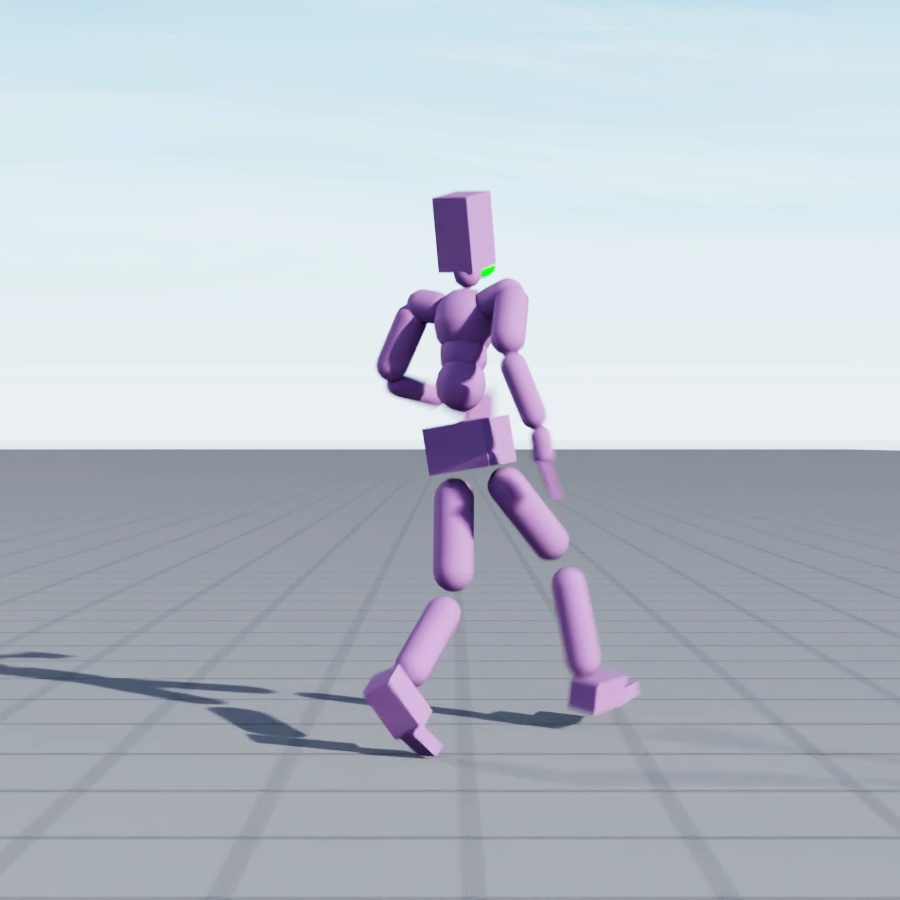}
         \caption{\textbf{Path following:} run}
         \label{fig: head run}
     \end{subfigure}
     
    \caption{\textbf{Motion tracking:} \alg~ generates full-body motion when tracking signals extracted from unseen kinematic motions. Precise fighting and dancing moves when tracking full-body information, a cartwheel from VR signals, and running by tracking the head (path following). The green spheres represent the target joint positions in each frame.}
    \label{fig: tracking}
\end{figure*}

\paragraph{Path-Following} The character is tasked with following a 3D path. This path specifies target positions for the head (including height) at each timestep. By varying the target locations, the character can be directed to perform different locomotion styles, such as walking, crouched-walking, and crawling. The paths are randomly generated with varying heights and speeds. Each episode has a length of 30 seconds.

\paragraph{Steering} The steering task is analogous to a joystick controller, where the character's movements are controlled by two target direction. One direction specifies the target heading direction and the other specifies the target direction (and speed). This enables separating the control between the direction the character should face and the direction it should move.

\paragraph{Reach} In addition to locomotion, we show that \alg~ can also be used to perform more fine-grained control over the movement of individual body parts. The goal of the reach task is for the right hand to reach a randomly changing target position. Once the target position changes, the character has 2 seconds to reach the position with its hand and stay at that location.

\paragraph{Object Interaction} Finally, we show that \alg~ can also be directed to generate natural interactions with objects by conditioning the model on features of a target object. In this task, we focus on sitting on a set of held-out objects. These objects were not used during the training phase. The character is first initialized at a random location between 2 and 10 meters away from the object. 

\section{Results}
The MaskedMimic framework enables versatile physics-based character control by formulating the problem as motion inpainting from partial constraints. In this section, we present key results demonstrating the effectiveness and flexibility of our approach.

\subsection{Motion Tracking}

\cref{tab: amass fullbody tracking,tab: amass reusable vr comparison,tab: amass tracking terrain} record the performance statistics for the motion tracking task, and \cref{fig: tracking} shows examples of the behaviors our model produces. \alg~ exhibits robust behaviors and improved generalization to new (test) motions compared to prior motion tracking models. When provided full-body targets, \alg~ tracks the entire motion of a karate fighter and a dancer, matching hand and foot positions of a fighting stance. When provided with partial target poses, for example recovering full motion from sparse VR sensors, \alg~ succeeds in reproducing a cartwheel across irregular terrain from only sparse VR constraints, as well as running while following a target trajectory for the head.

\begin{table}[]
    \centering
    \caption{\textbf{Full-body tracking, flat terrain:} Tracking full-body kinematic recordings from the AMASS dataset \cite{AMASS:ICCV:2019}. We highlight be best performing model on test motions.}
    \tabcolsep=0.14cm
    \begin{tabular}{l|cc|cc}
                                & \multicolumn{2}{c|}{\textbf{Train}} & \multicolumn{2}{c}{\textbf{Test}} \\
                                & Success & MPJPE & Success & MPJPE \\ \hline\hline

        {FC} (ours)  & 99.96\% & 30.4 & \cellcolor{green!25} 99.9\% & \cellcolor{green!25} 31.3 \\ \hline

         {PHC+}          & 100\% & 26.6 & 99.2\% & 36.1 \\ \hline\hline

         {\alg} (ours)          & 99.4\% & 32.9 & \cellcolor{green!25} 99.2\% & \cellcolor{green!25} 35.1 \\ \hline

         {PULSE}         & 99.8\% & 39.2 & 97.1\% & 54.1
    \end{tabular}
    \label{tab: amass fullbody tracking}
\end{table}

\begin{table}[]
    \centering
    \caption{\textbf{VR tracking, flat terrain:} Tracking VR-signals extracted from the AMASS dataset. In addition to the full-body tracking (MPJPE), we report that \alg~ received a \textbf{MPOJPE} (VR tracking error) of 39.5 (train) and 45.8 (test).}
    \tabcolsep=0.14cm
    \begin{tabular}{l|cc|cc}
                            & \multicolumn{2}{c|}{\textbf{Train}} & \multicolumn{2}{c}{\textbf{Test}} \\ 
                            & Success & MPJPE & Success & MPJPE \\ \hline\hline

        {\alg} (ours)       & 98.6\% & 50 & \cellcolor{green!25} 98.1\% & \cellcolor{green!25} 58.1 \\ \hline

        {PULSE}      & 99.5\% & 57.8 & 93.4\% & 88.6 \\ \hline

        {ASE}        & 79.8\% & 103 & 37.6\% & 120.5 \\ \hline

        {CALM}       & 16.6\% & 130.7 & 10.1\% & 122.4
    \end{tabular}
    \label{tab: amass reusable vr comparison}
\end{table}

\paragraph{Full-body tracking}
\cref{tab: amass fullbody tracking} shows the performance on the full-body motion tracking task. Performance is evaluated on both the AMASS train and test splits. We consider a trial "failed" if at any frame the average joint deviation is larger than 0.5m \cite{luo2021dynamics}. To complement the success metric, we report the MPJPE (Mean Per Joint Position Error, in millimeters). MPJPE measures how closely the character can track the target joint positions (in global coordinates).

Our fully-constrained tracker FC outperforms PHC+ \cite{luo2023perpetual,luo2023universal}, reducing the tracking failure rate on unseen motions by 62.5\%. In addition to a lower failure rate, our controller also supports a wider range of motions, irregular terrains, and object interactions. We attribute these performance improvements to our architecture and data augmentation techniques.

A key difference in our approach is the use of a single unified network, in contrast to the mixture-of-experts (MoE) model employed by PHC. While MoE approaches, such as training different controllers for each task \cite{peng2018deepmimic} or using a progressively growing mixture of multi-motion trackers \cite{luo2023perpetual}, have been applied to fully-constrained tracking problems, they present certain challenges. As motion variety increases, maintaining multiple experts becomes difficult in an online distillation (DAgger) regime. Moreover, the MoE architecture in PHC requires an additional gating network to select and blend between multiple networks (experts) depending on the target motion.

Our experiments demonstrate that a single monolithic network offers better generalization capabilities. This approach not only simplifies the architecture but also avoids the complexities associated with expert selection and blending. The superior performance of our model suggests that, in the context of full-body tracking, a well-designed unified network can effectively capture the diversity of motions without the need for specialized experts.

When comparing \alg~ with PULSE \cite{luo2023universal}, we observe that PULSE exhibits more pronounced overfitting to the training data, while \alg~ demonstrates superior generalization performance on the test set. This distinction can be attributed to two key factors. First, the expert used for training \alg~ possesses better generalization capabilities, a characteristic that may transfer to the student during the distillation process. Additionally, \alg~ is designed to tackle a wide range of tasks across diverse scenes, which likely contributes to the model's enhanced generalization capabilities. This multi-task, multi-environment training approach appears to foster a more robust and adaptable model, enabling it to perform well on unseen data and scenarios.

\paragraph{VR tracking}
\cref{tab: amass reusable vr comparison} compares the performance of various models on the VR tracking task on flat terrain. We report metrics on the 3 available sensors, measuring success rate for tracking the head and hands, in addition to the MPJPE measured on the unseen full-body motion, and the tracking error on the observed joints (MPOJPE, mean per observed joints positional error). We compare to PULSE \cite{luo2023universal}, ASE \cite{peng2022ase}, and CALM \cite{tessler2023calm}. These methods train a reusable low-level controller. Then, they train a new high-level controller to manage this specific task. In contrast, \alg~ is applied directly to this sparse tracking task without any additional training. Despite not being explicitly trained on this task, \alg~ outperforms other models by a significant margin when evaluated on tracking target trajectories extracted from the AMASS test set.

\paragraph{Joint sparsity}
The results presented in \cref{tab: amass reusable anyjoint comparison} illustrate \alg's capability to produce full-body motion while varying the conditioned joints. The analysis reveals a hierarchy of difficulty in joint tracking. Specifically, foot tracking presents a greater challenge compared to hand tracking. Additionally, pelvis tracking proves to be less demanding than tracking the head and hands. We find the last result particularly interesting. Reconstructing full-body motion from VR sensors is an important task for gaming and collaborative work. These results suggest that adding a pelvis sensor, or predicting the pelvis positioning using head-mounted sensors, could provide a noticeable boost in the ability to recover the user's motion.

\begin{table}[]
    \centering
    \caption{\textbf{MaskedMimic, joint sparsity, flat terrain:} Tracking partial-joint signals extracted from the AMASS dataset.}
    \tabcolsep=0.14cm
    \begin{tabular}{l|cc|cc}
                            & \multicolumn{2}{c|}{\textbf{Train}} & \multicolumn{2}{c}{\textbf{Test}} \\ 
                            & Success & MPOJPE & Success & MPOJPE \\ \hline\hline
        
        Full body   & 99.4\% & 32.9 & 99.1\% & 35.1 \\ \hline

        Pelvis      & 98.4\% & 31.4 & 98.4\% & 33.4 \\ \hline
         
        VR          & 98.6\% & 39.5 & 98.1\% & 45.8 \\ \hline

        Head        & 97.7\% & 42.6 & 97.9\% & 45.6 \\ \hline

        Hands       & 95.2\% & 60.2 & 93.4\% & 69.6 \\ \hline

        Feet        & 92.7\% & 88 & 91.8\% & 94.3
    \end{tabular}
    \label{tab: amass reusable anyjoint comparison}
\end{table}

\begin{table}[]
    \centering
    \caption{\textbf{\alg, irregular terrain:} We evaluate our models from both training stages on the task of tracking motions from the AMASS dataset across irregular terrains.}
    \tabcolsep=0.14cm
    \begin{tabular}{ll|cc|cc}
                                                &       & \multicolumn{2}{c|}{\textbf{Full-body}} & \multicolumn{2}{c}{\textbf{VR}} \\
                                                &       & Success & MPJPE & Success & MPOJPE \\ \hline\hline

        \multirow{2}{*}{{FC}} & Train & 98\% & 51.5 & \cellcolor{gray!25} & \cellcolor{gray!25} \\
                                                & Test & \cellcolor{green!25} 98.2\% & \cellcolor{green!25} 51 & \cellcolor{gray!25} & \cellcolor{gray!25} \\ \hline
                                                
        \multirow{2}{*}{{\alg}}          & Train & 94.7\% & 61.3 & 94.4\% & 62.7 \\
                                                & Test & 95.4\% & 62.9 & \cellcolor{green!25} 93.6\% & \cellcolor{green!25} 69.4
    \end{tabular}
    \label{tab: amass tracking terrain}
\end{table}

\paragraph{Irregular terrains}
Most prior models are not trained for interactions with objects and irregular terrains. Therefore, our evaluation on irregular terrain will compare the fully-constrained motion tracking controller with the more versatile \alg~ controller. As shown in \cref{tab: amass tracking terrain}, \alg~ exhibits similar success rates and tracking errors across both the train and test sets, when evaluated on randomly generated irregular terrain. These results further validate the robustness and generalization capabilities of our \alg~ model.

\subsection{Goal-Engineering}

\begin{figure*}[t]
     \centering
     \begin{subfigure}[b]{0.495\textwidth}
         \centering
         \includegraphics[width=0.2475\textwidth]{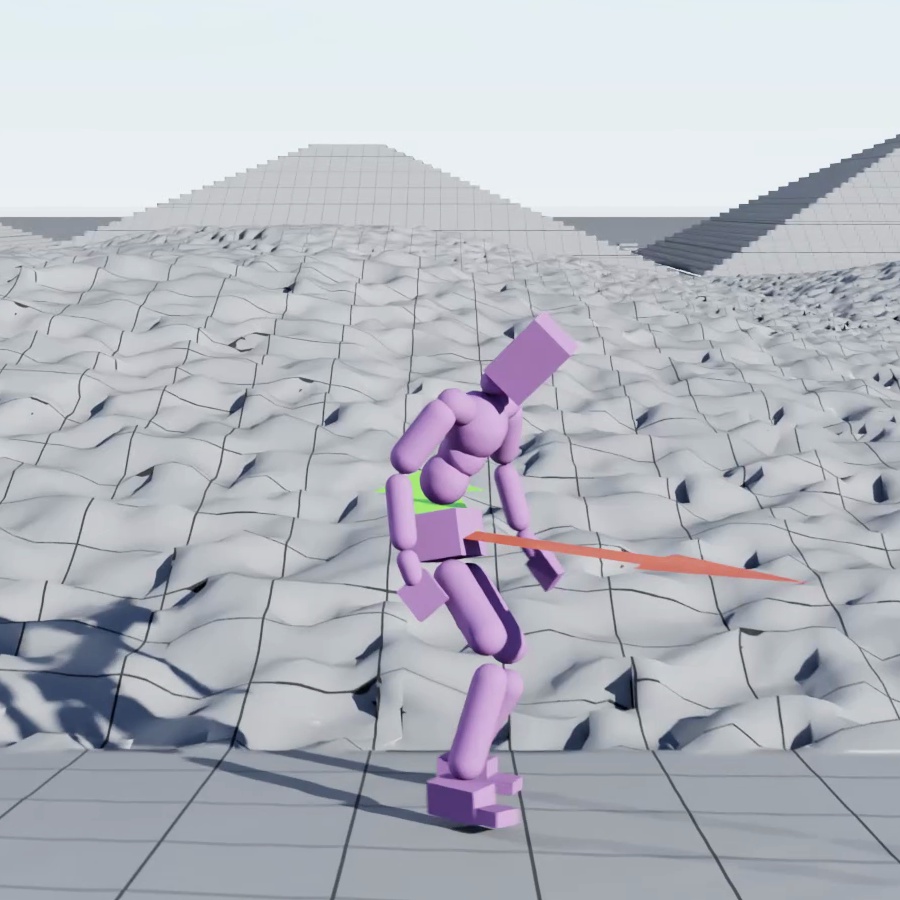}\hfill
         \includegraphics[width=0.2475\textwidth]{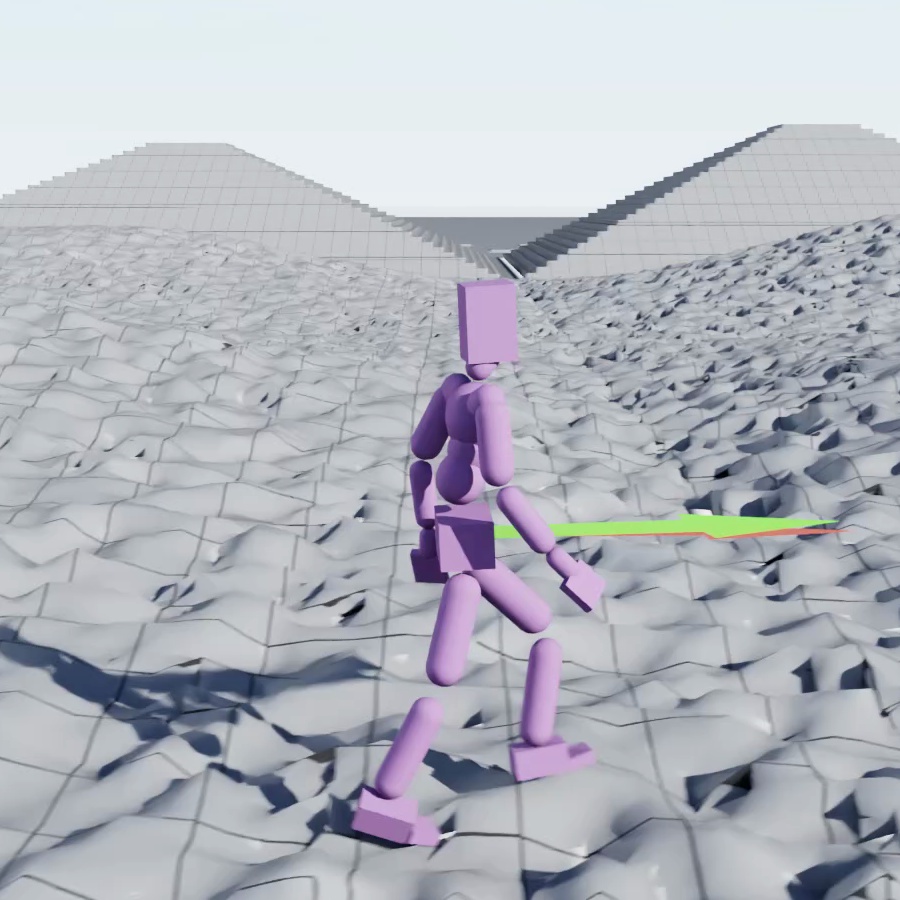}\hfill
         \includegraphics[width=0.2475\textwidth]{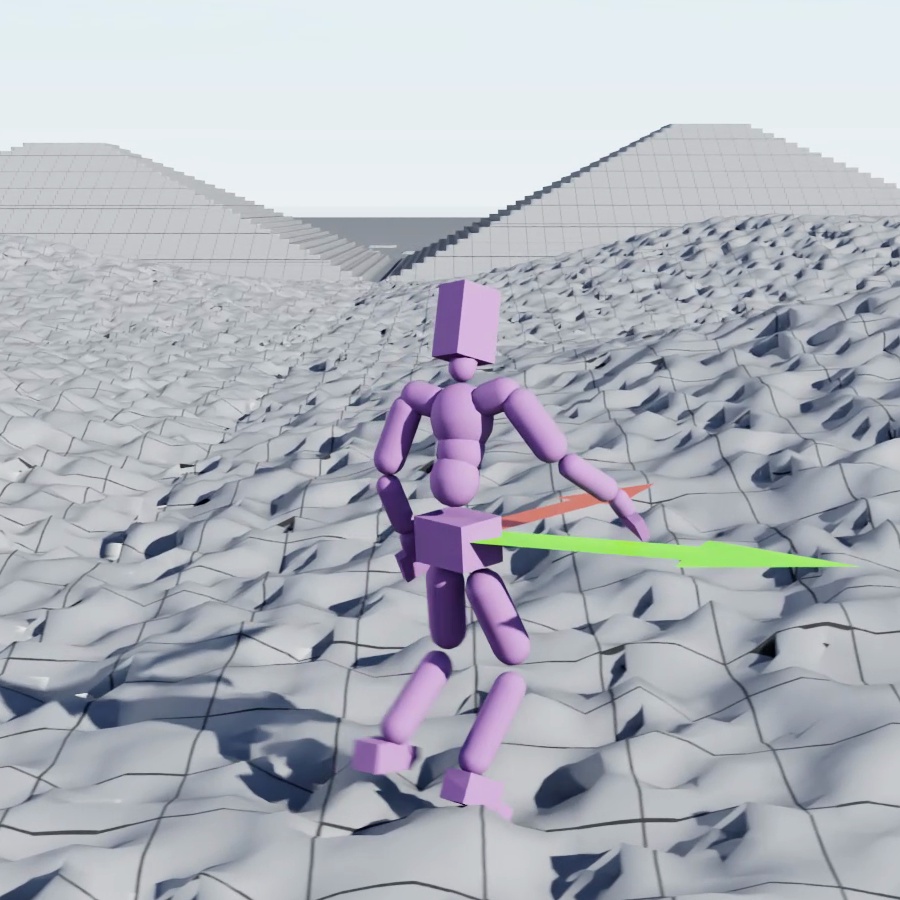}\hfill
         \includegraphics[width=0.2475\textwidth]{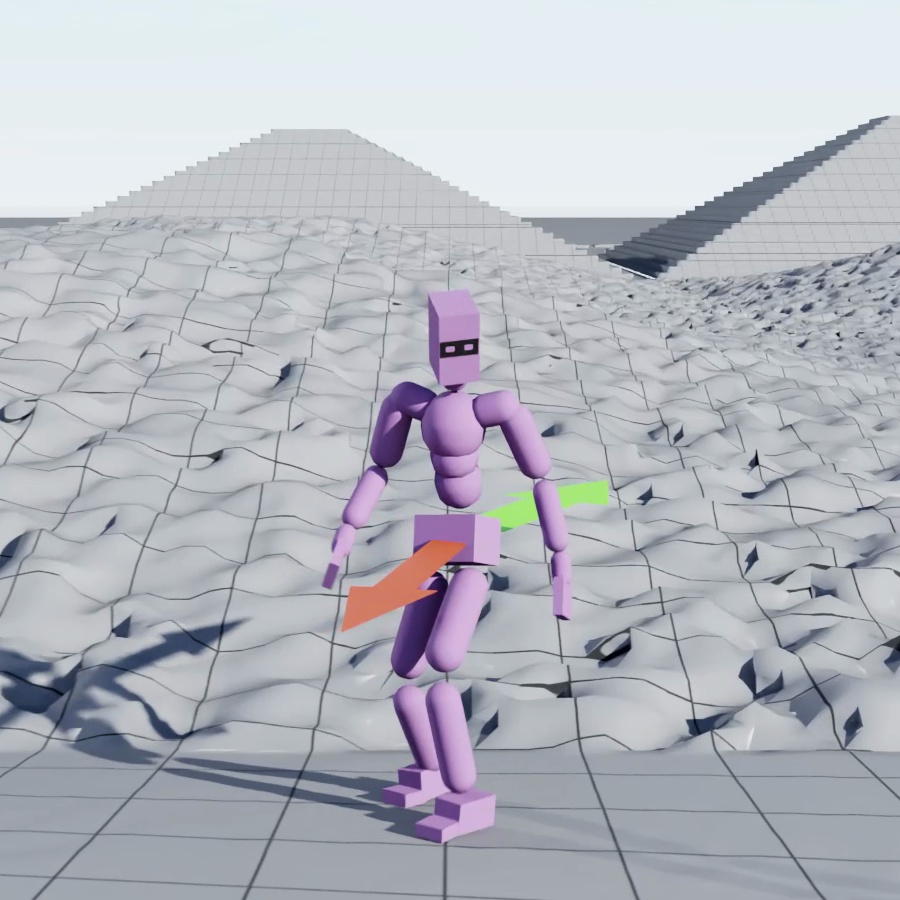}
         \caption{\textbf{Steering:} rough terrain}
         \label{fig: steering 7}
     \end{subfigure}
     \begin{subfigure}[b]{0.495\textwidth}
         \centering
         \includegraphics[width=0.2475\textwidth]{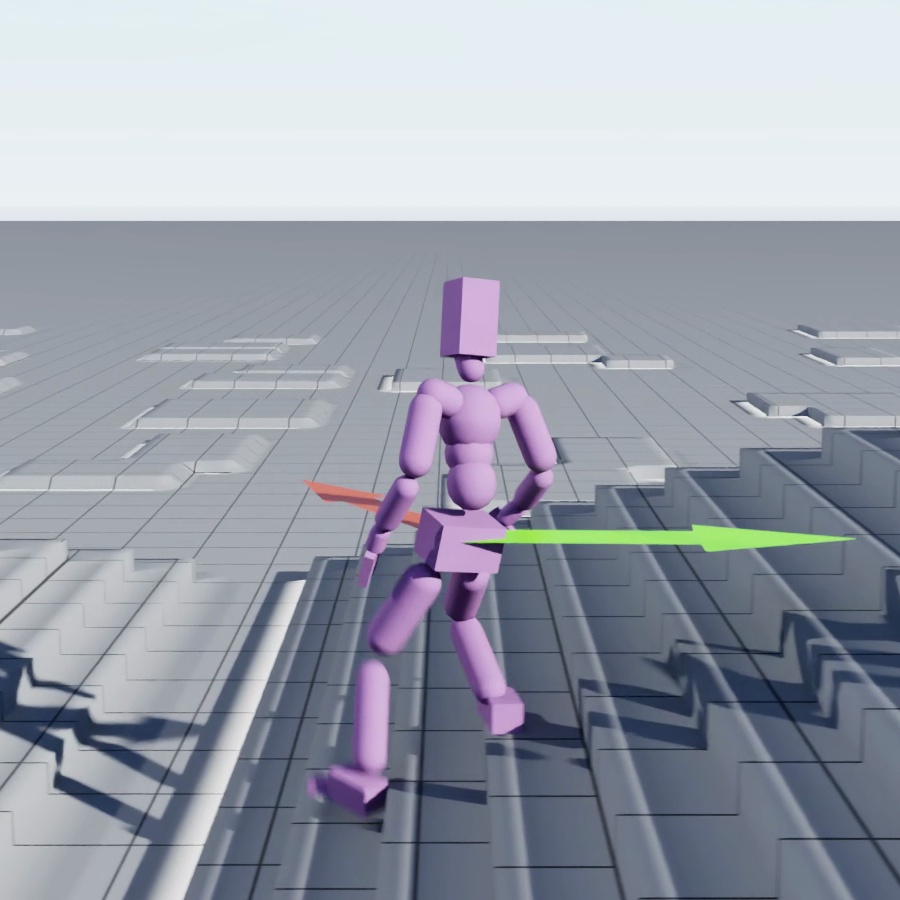}\hfill
         \includegraphics[width=0.2475\textwidth]{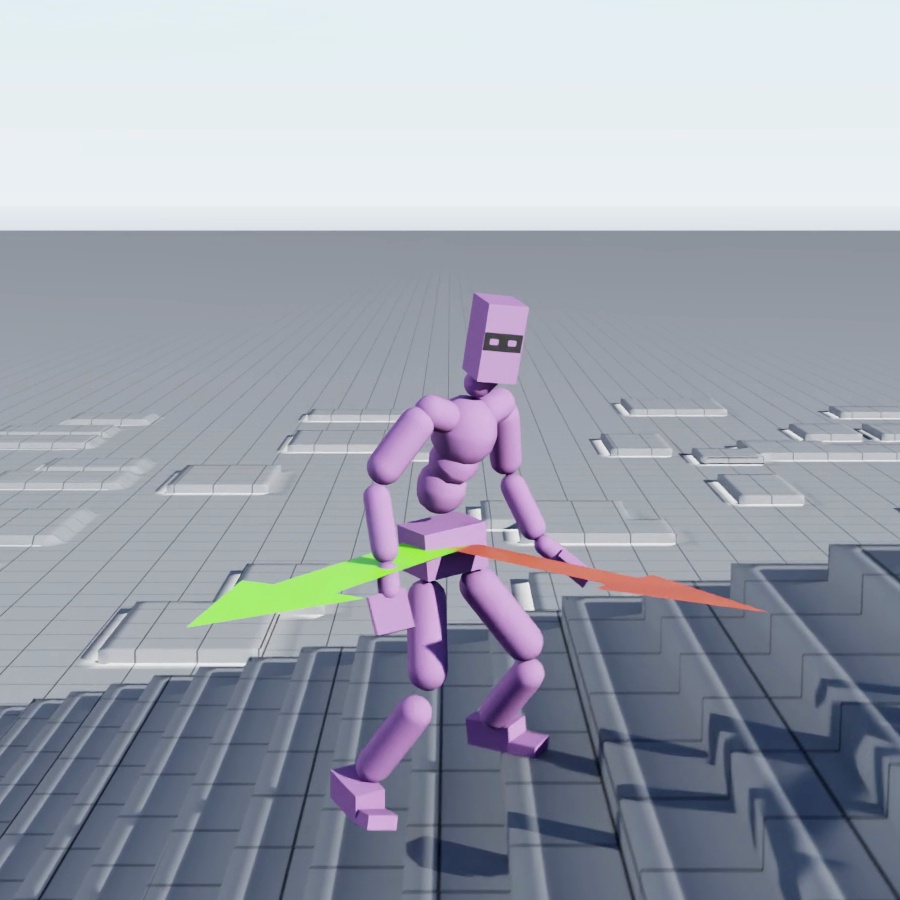}\hfill
         \includegraphics[width=0.2475\textwidth]{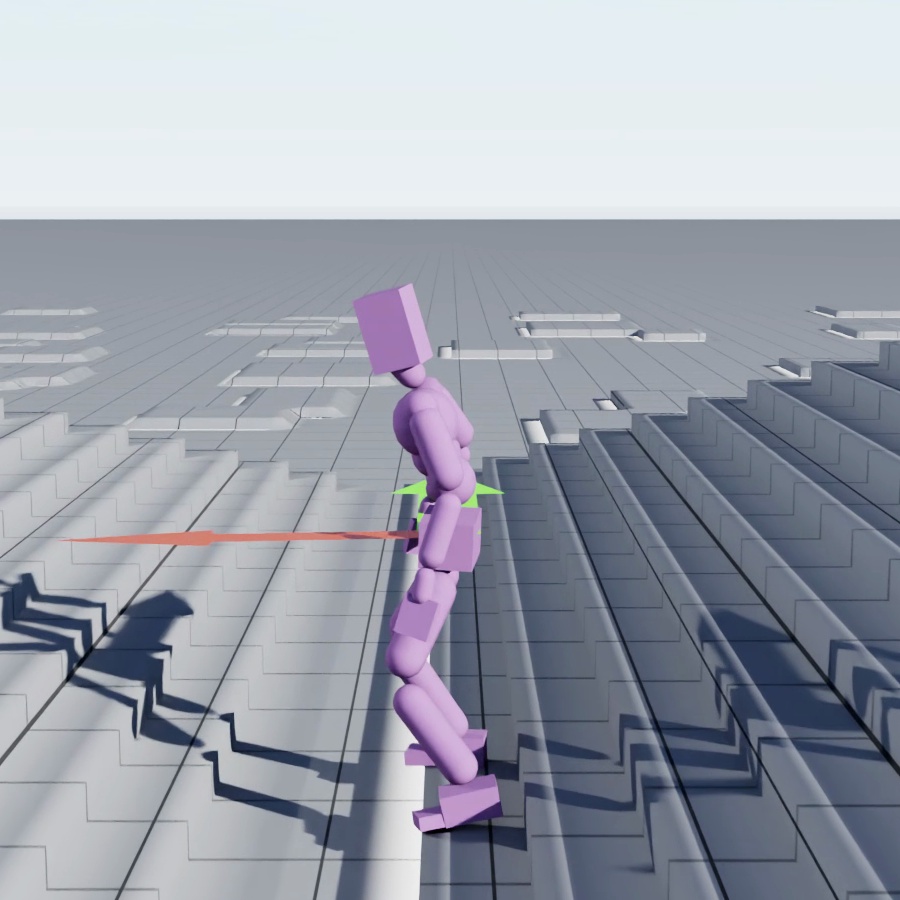}\hfill
         \includegraphics[width=0.2475\textwidth]{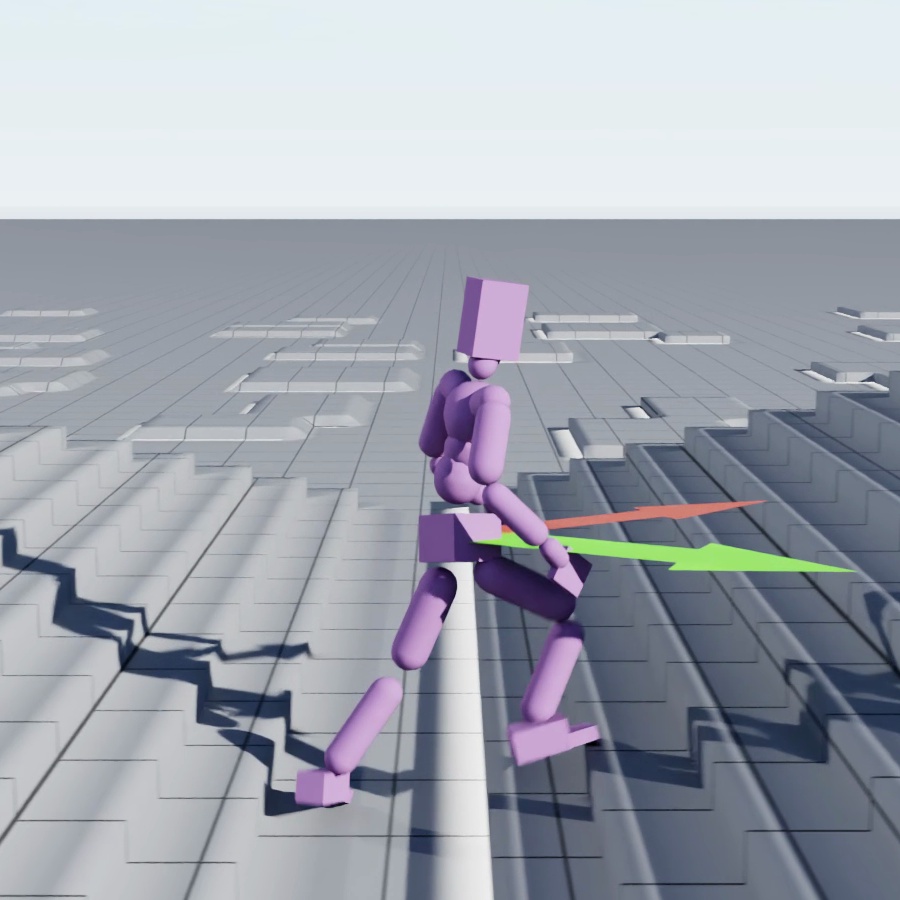}
         \caption{\textbf{Steering:} stairs}
         \label{fig: steering 13}
     \end{subfigure}\\
     
     \begin{subfigure}[b]{0.495\textwidth}
         \centering
         \includegraphics[trim={4cm 4cm 4cm 4cm},clip,width=0.2475\textwidth]{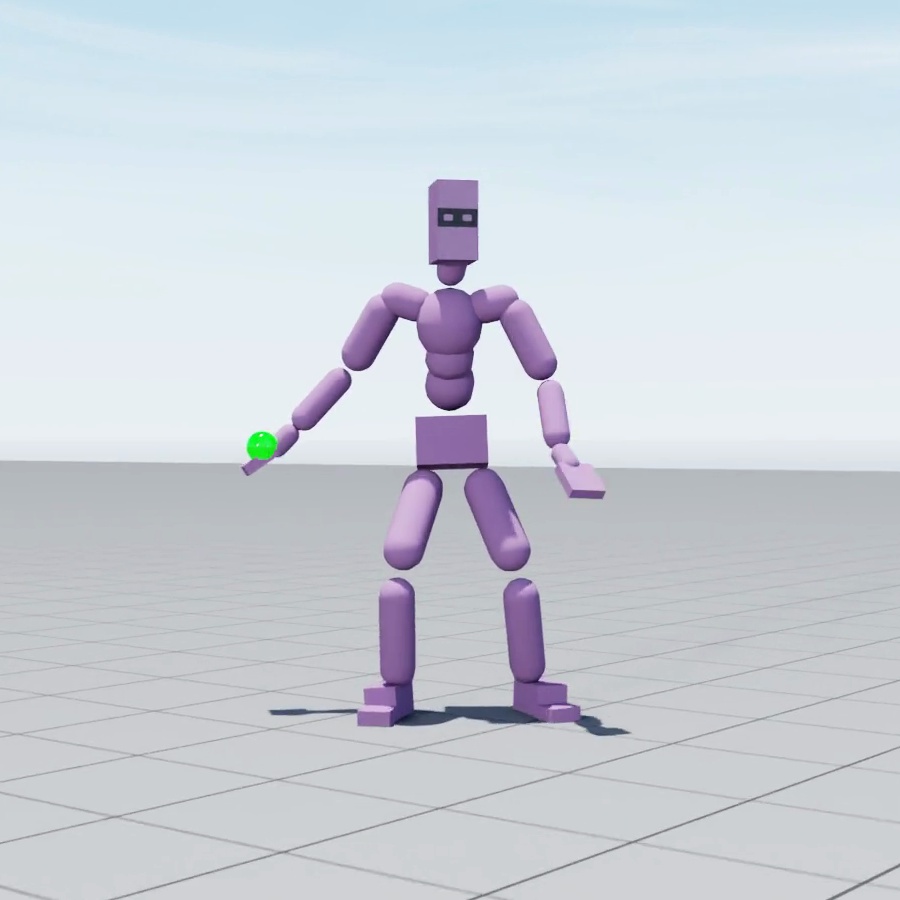}\hfill
         \includegraphics[trim={6cm 6cm 6cm 6cm},clip,width=0.2475\textwidth]{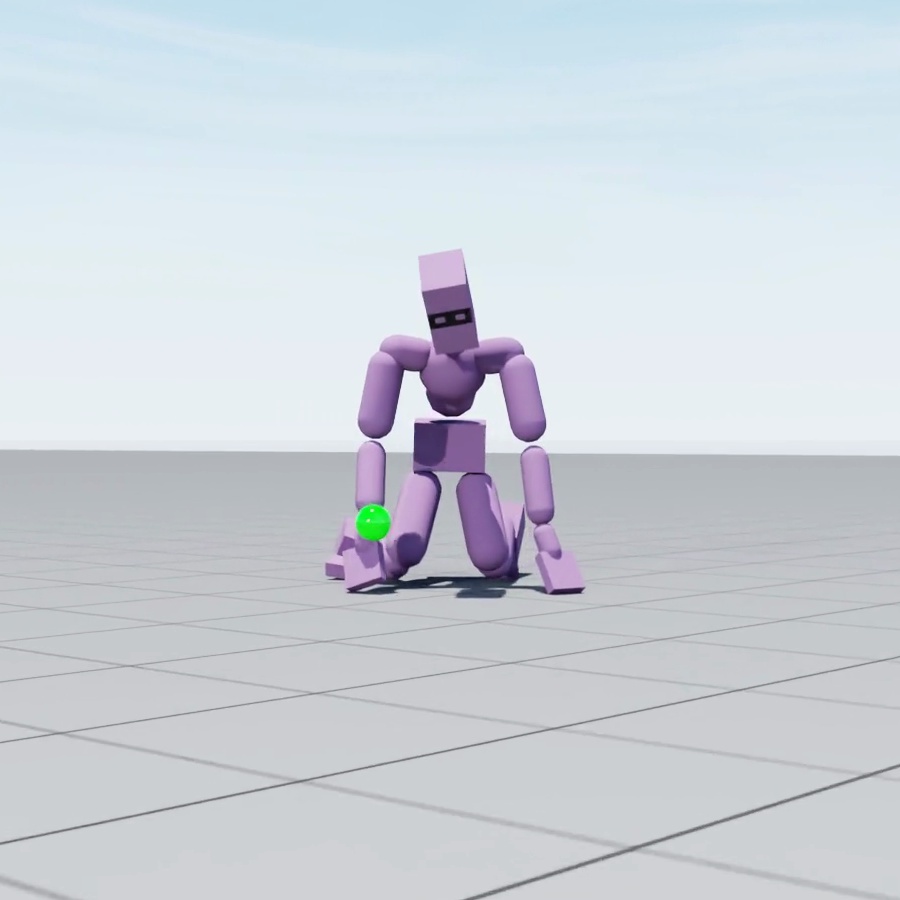}\hfill
         \includegraphics[trim={4cm 4cm 4cm 4cm},clip,width=0.2475\textwidth]{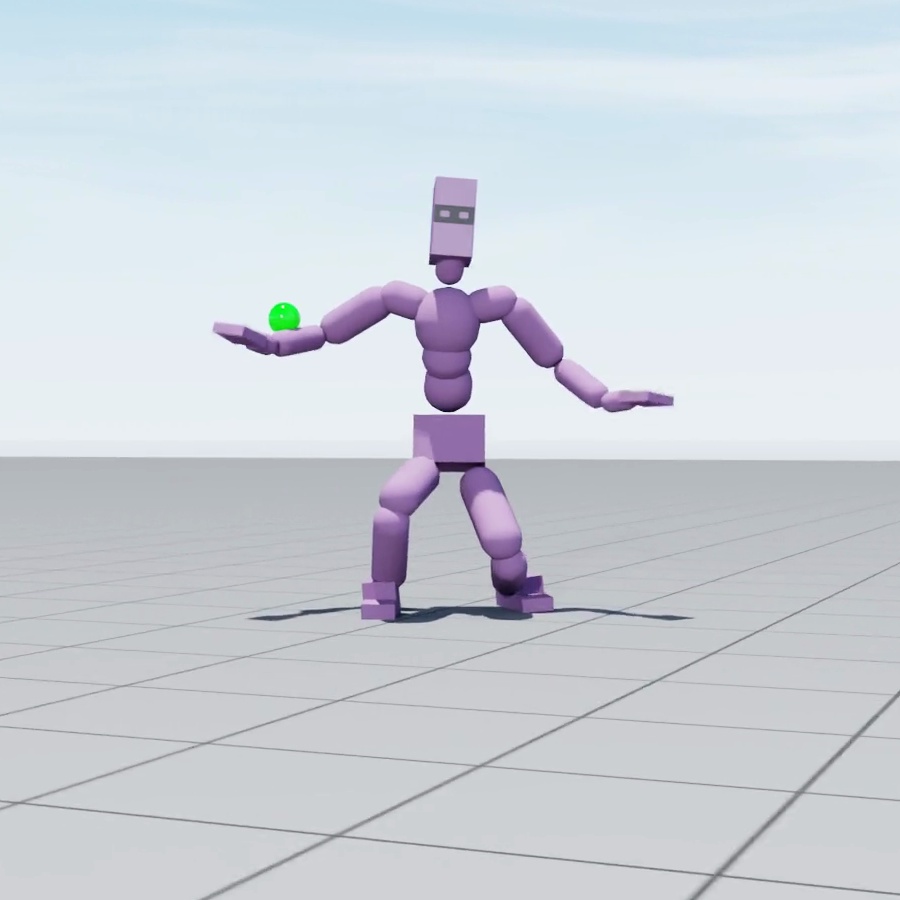}\hfill
         \includegraphics[trim={4cm 4cm 4cm 4cm},clip,width=0.2475\textwidth]{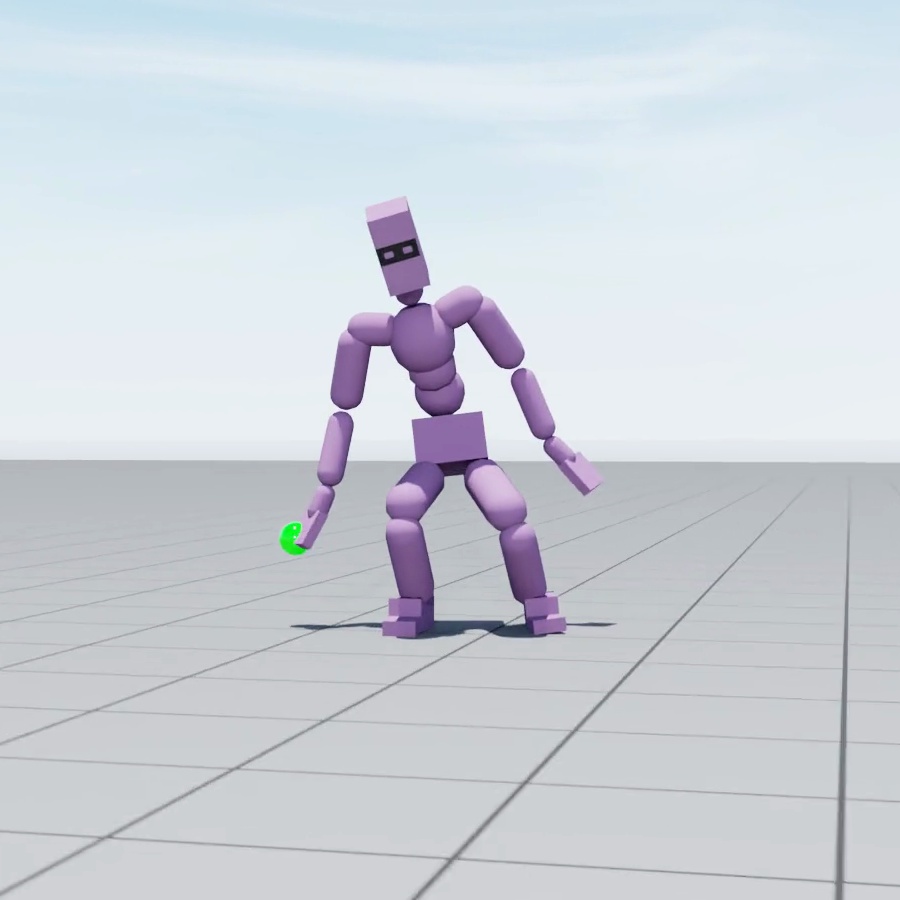}
         \caption{\textbf{Reach:} flat terrain}
         \label{fig: reach 1}
     \end{subfigure}
     \begin{subfigure}[b]{0.495\textwidth}
         \centering
         \includegraphics[width=0.2475\textwidth]{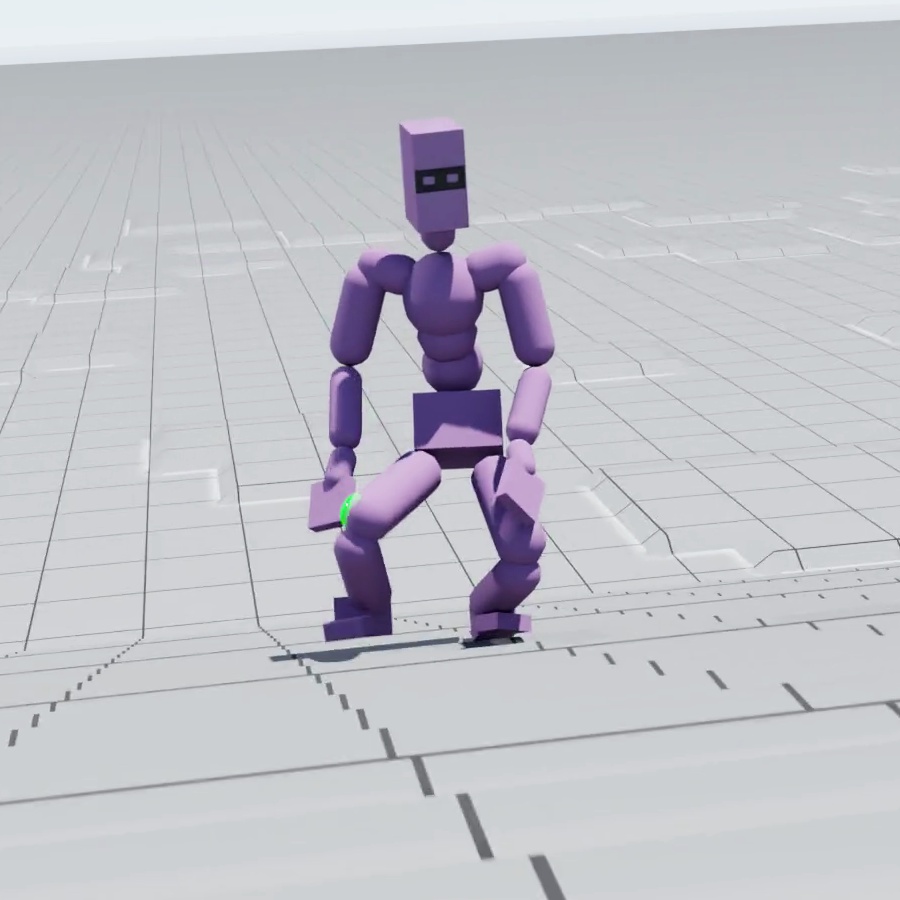}\hfill
         \includegraphics[width=0.2475\textwidth]{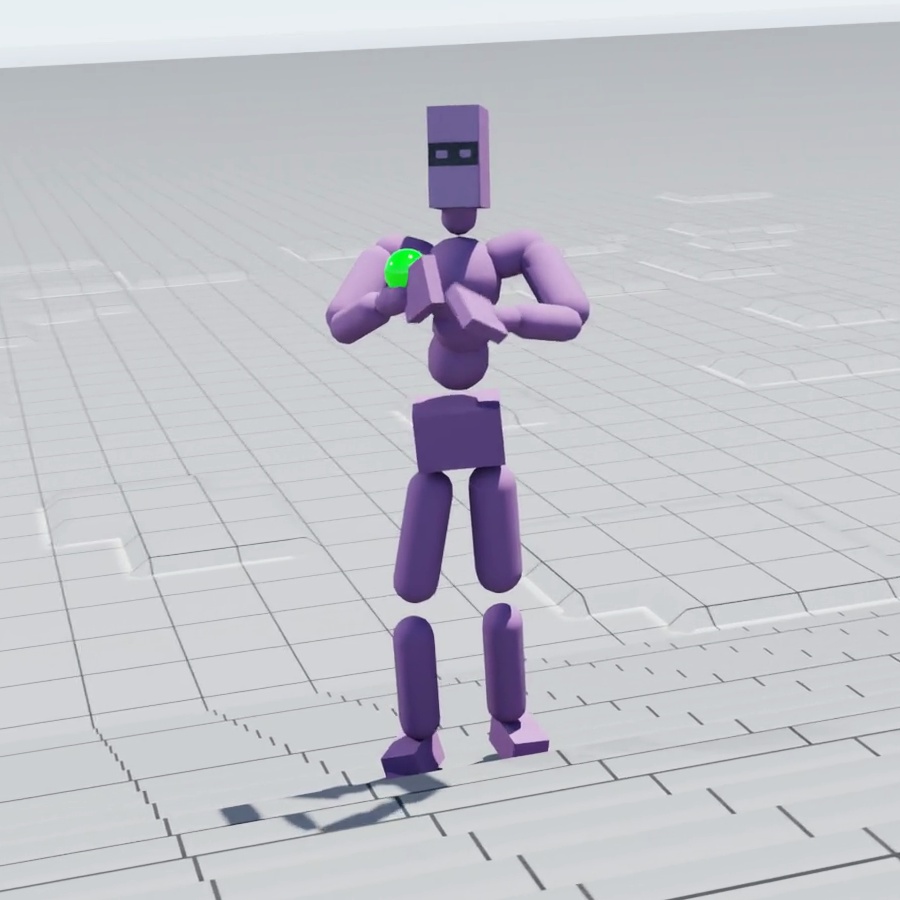}\hfill
         \includegraphics[width=0.2475\textwidth]{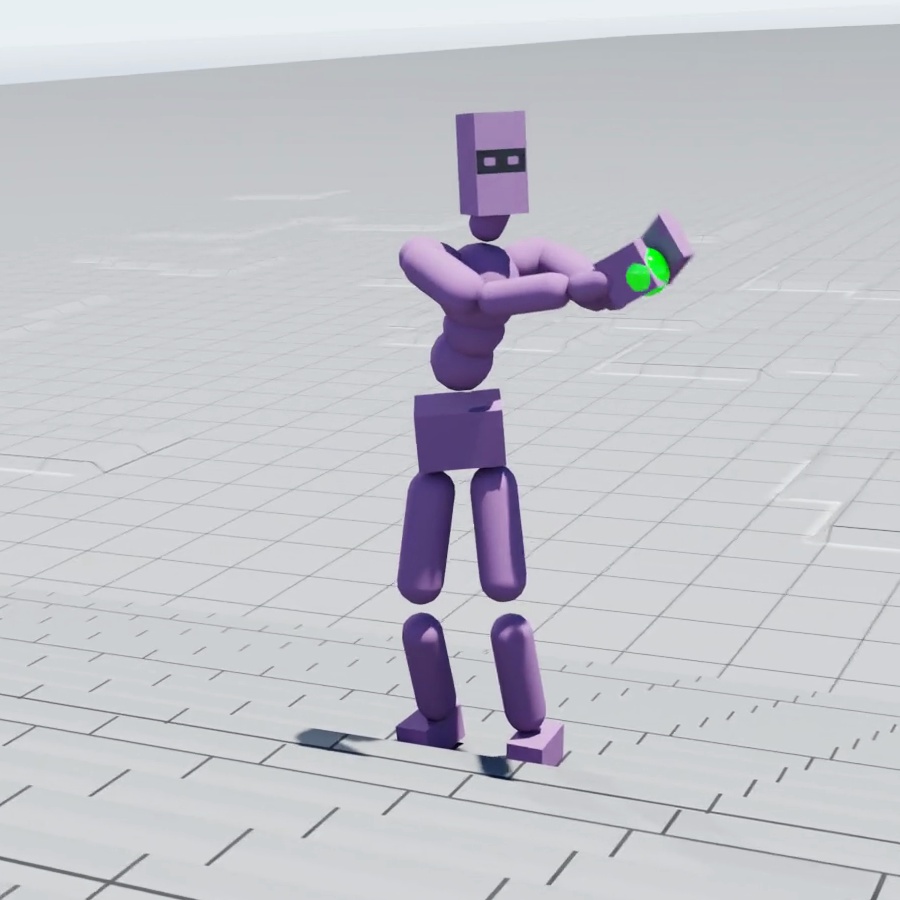}\hfill
         \includegraphics[width=0.2475\textwidth]{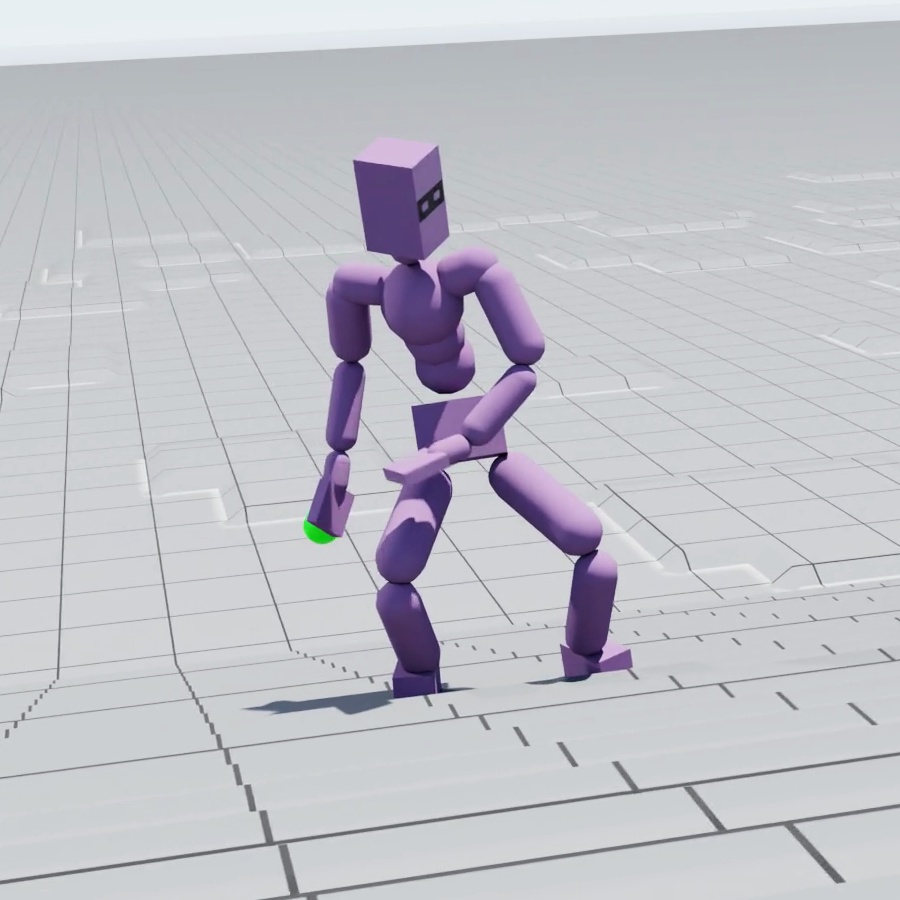}
         \caption{\textbf{Reach:} stairs}
         \label{fig: reach 20}
     \end{subfigure}\\
    
    \begin{subfigure}[b]{0.495\textwidth}
         \centering
         \includegraphics[trim={4cm 2cm 4cm 4cm},clip,width=0.2475\textwidth]{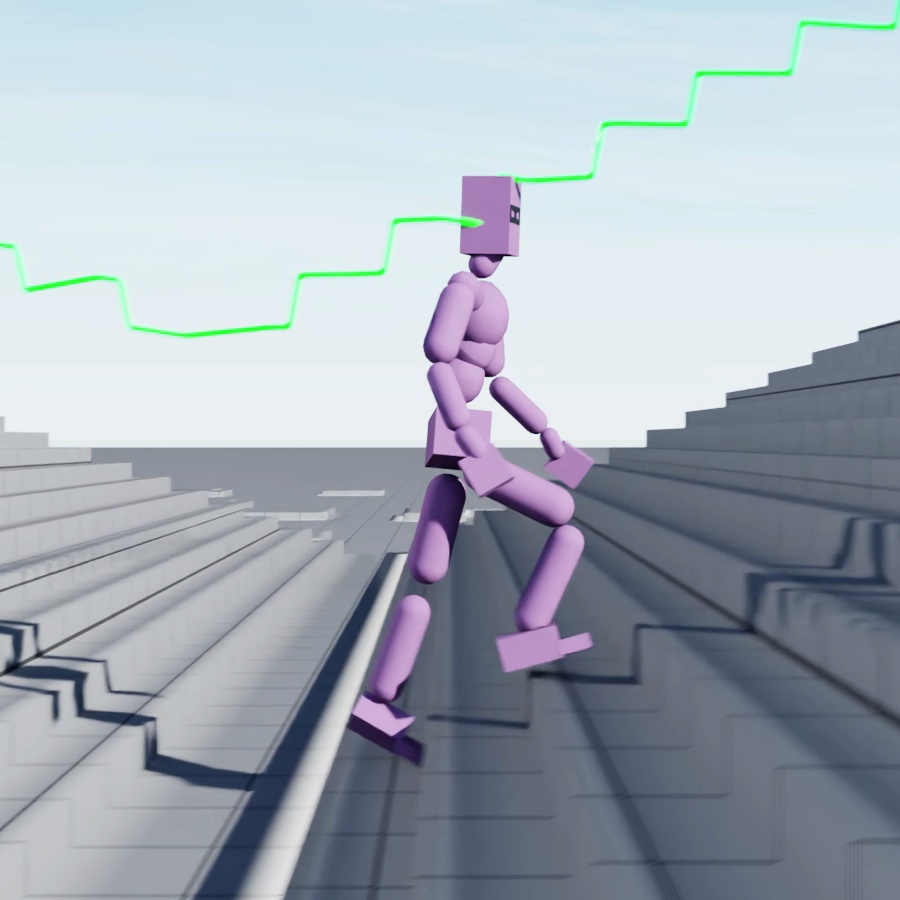}\hfill
         \includegraphics[trim={4cm 2cm 4cm 4cm},clip,width=0.2475\textwidth]{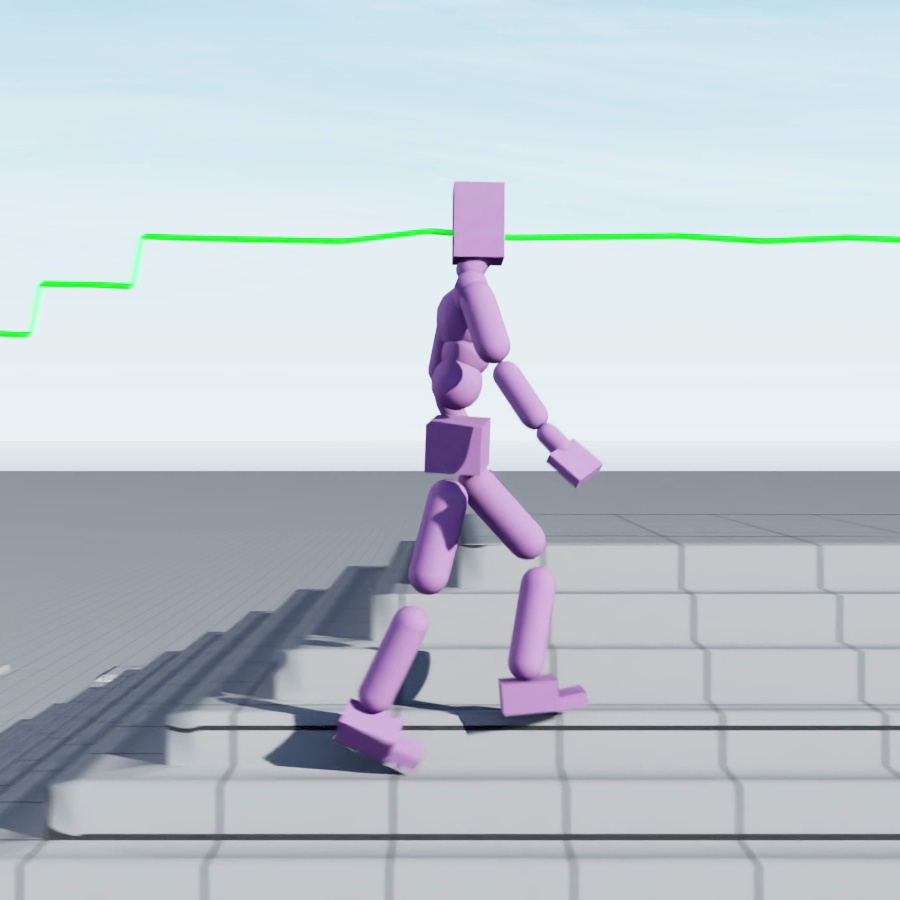}\hfill
         \includegraphics[trim={4cm 2cm 4cm 4cm},clip,width=0.2475\textwidth]{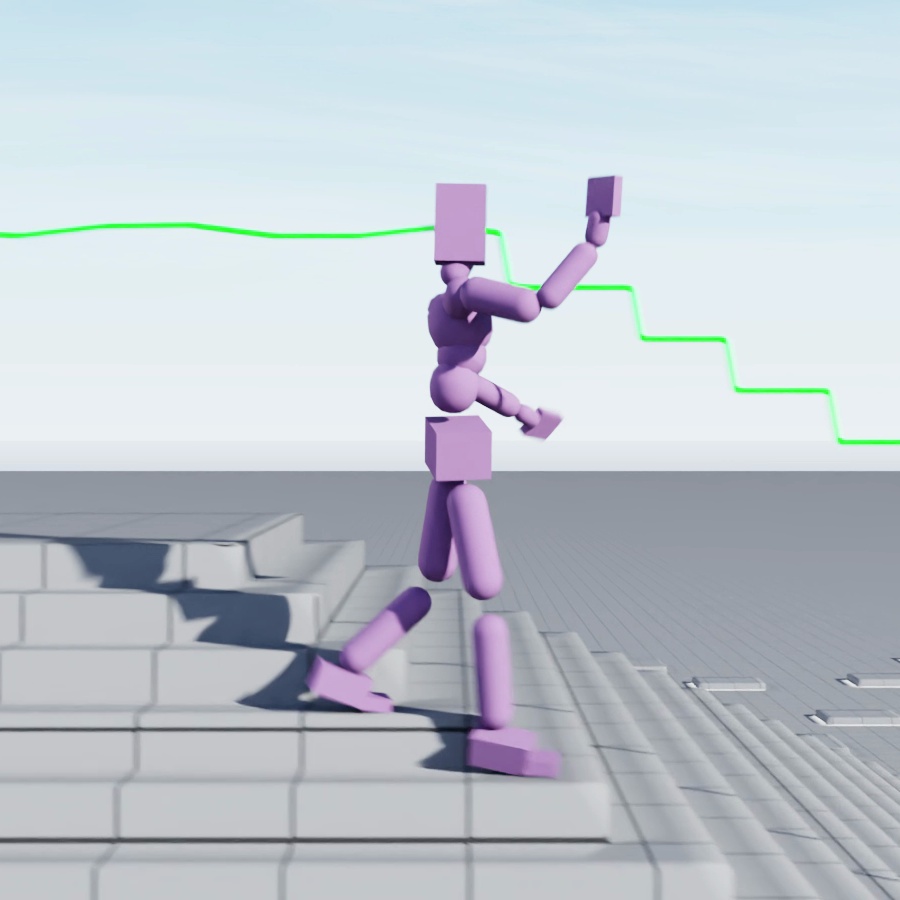}\hfill
         \includegraphics[trim={4cm 2cm 4cm 4cm},clip,width=0.2475\textwidth]{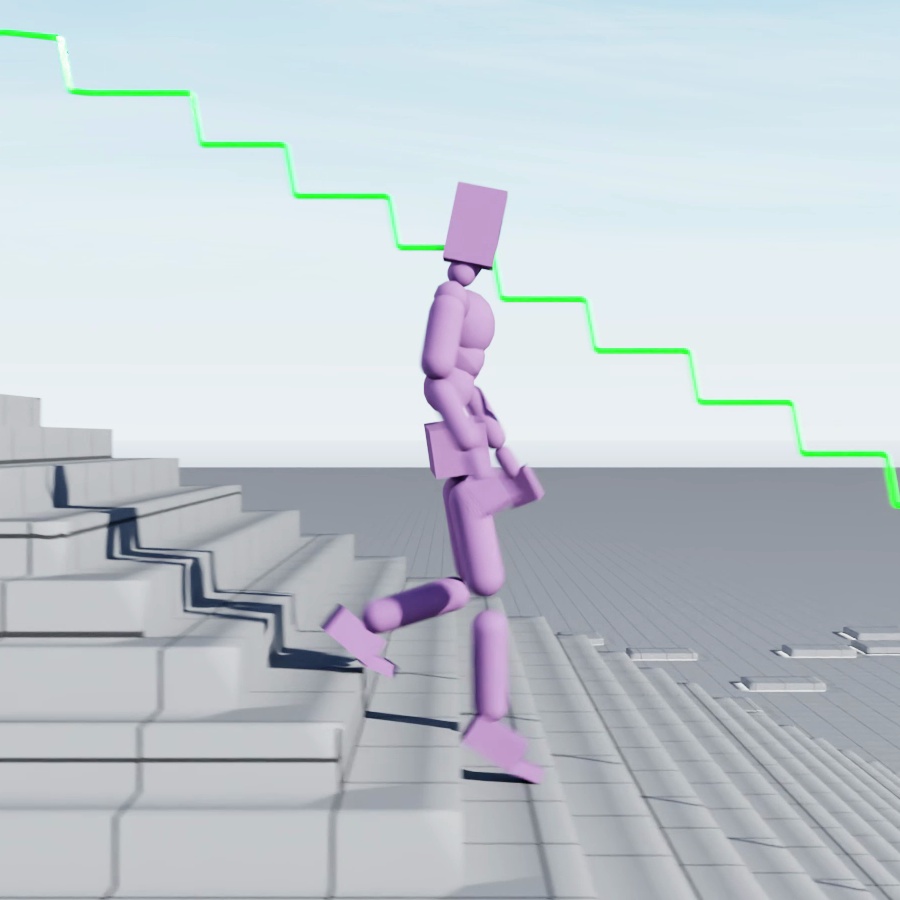}
         \caption{\textbf{Path-Following:} walk}
         \label{fig: locomotion walk}
     \end{subfigure}
     \begin{subfigure}[b]{0.495\textwidth}
         \centering
         \includegraphics[trim={4cm 2cm 4cm 4cm},clip,width=0.2475\textwidth]{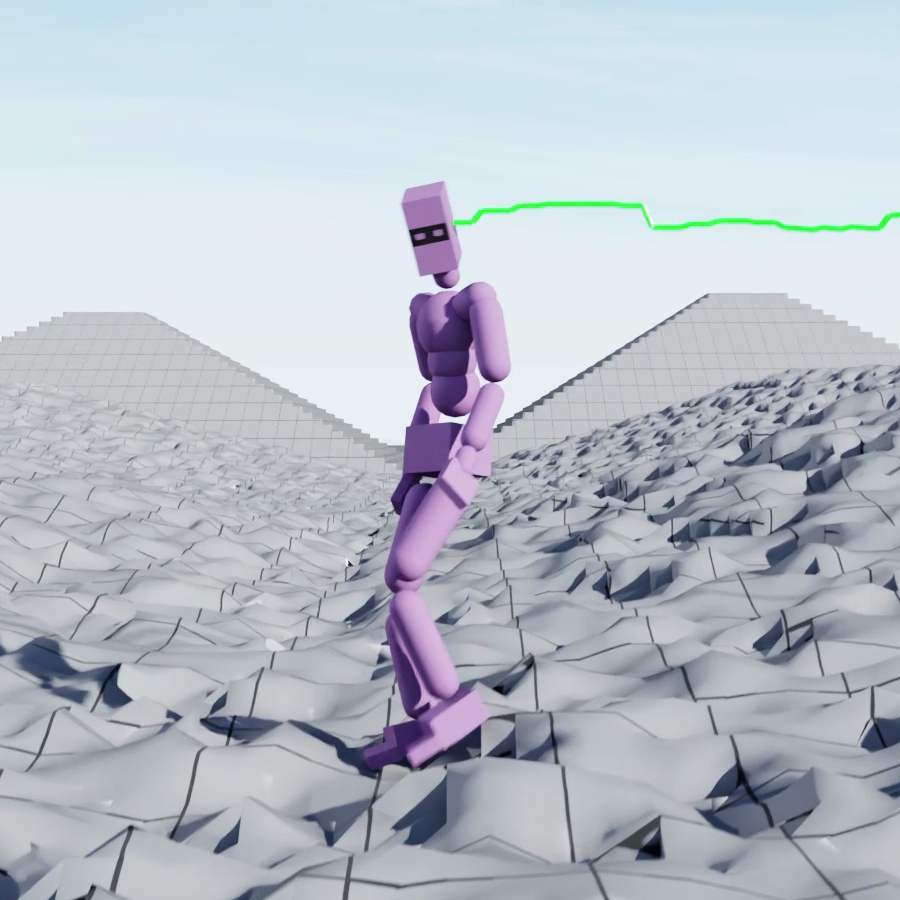}\hfill
         \includegraphics[trim={4cm 2cm 4cm 4cm},clip,width=0.2475\textwidth]{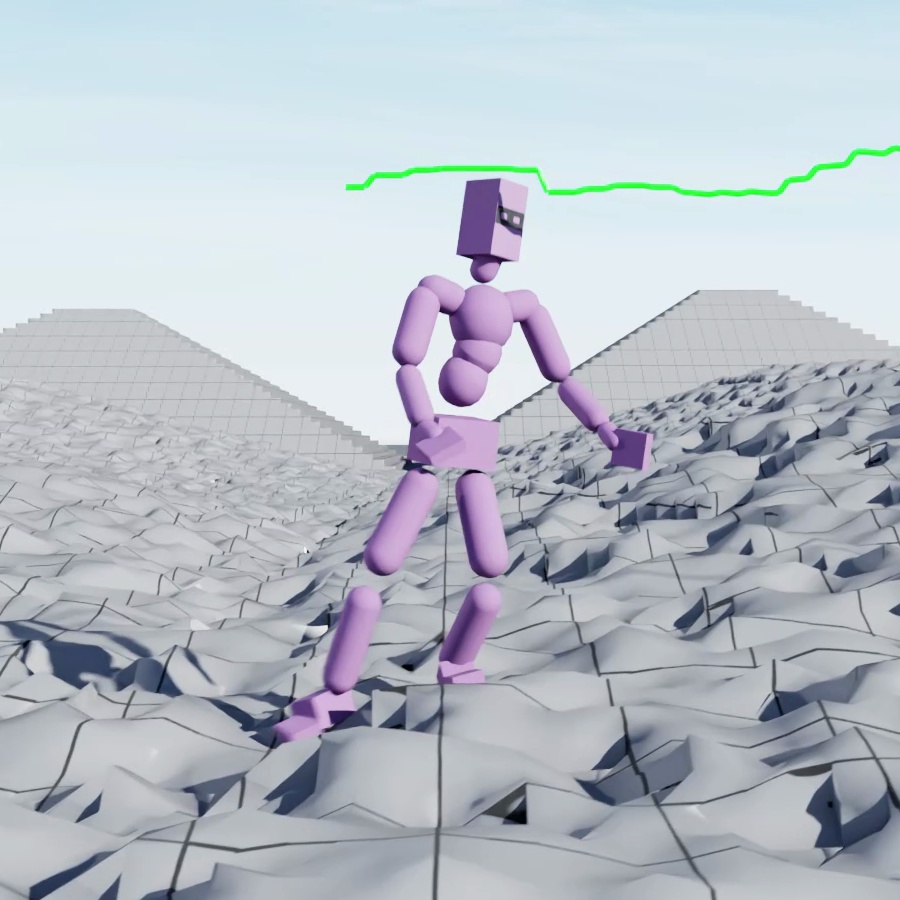}\hfill
         \includegraphics[trim={4cm 2cm 4cm 4cm},clip,width=0.2475\textwidth]{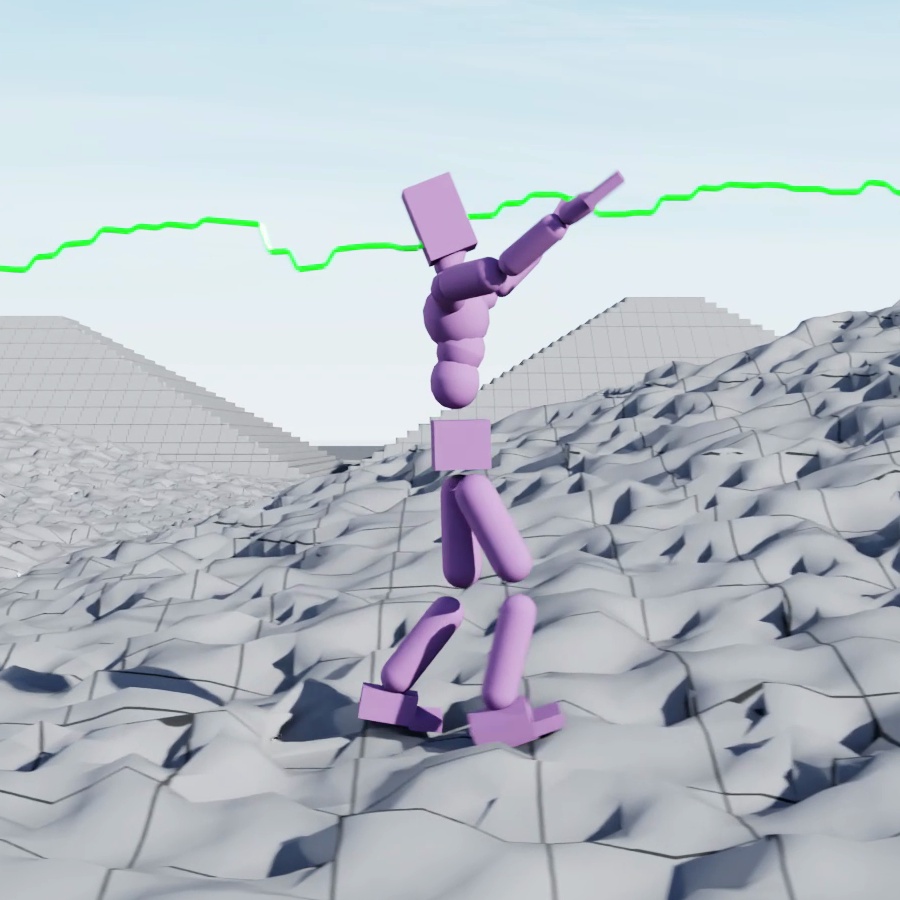}\hfill
         \includegraphics[trim={4cm 2cm 4cm 4cm},clip,width=0.2475\textwidth]{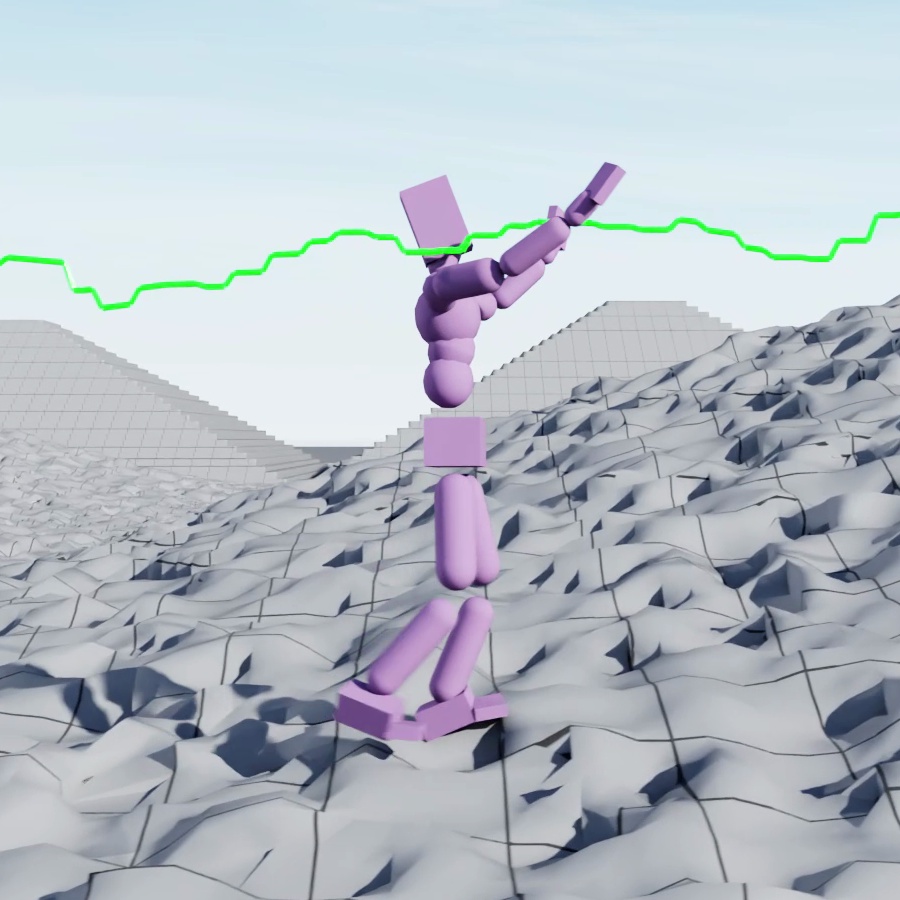}
         \caption{\textbf{Path-Following + Text:} ``a person raises both hands and walks forward''}
         \label{fig: locomotion walk text}
     \end{subfigure}
     \begin{subfigure}[b]{0.99\textwidth}
         \centering
         \scalebox{-1}[1]{\includegraphics[trim={0cm 14cm 0cm 15cm},clip,width=\textwidth]{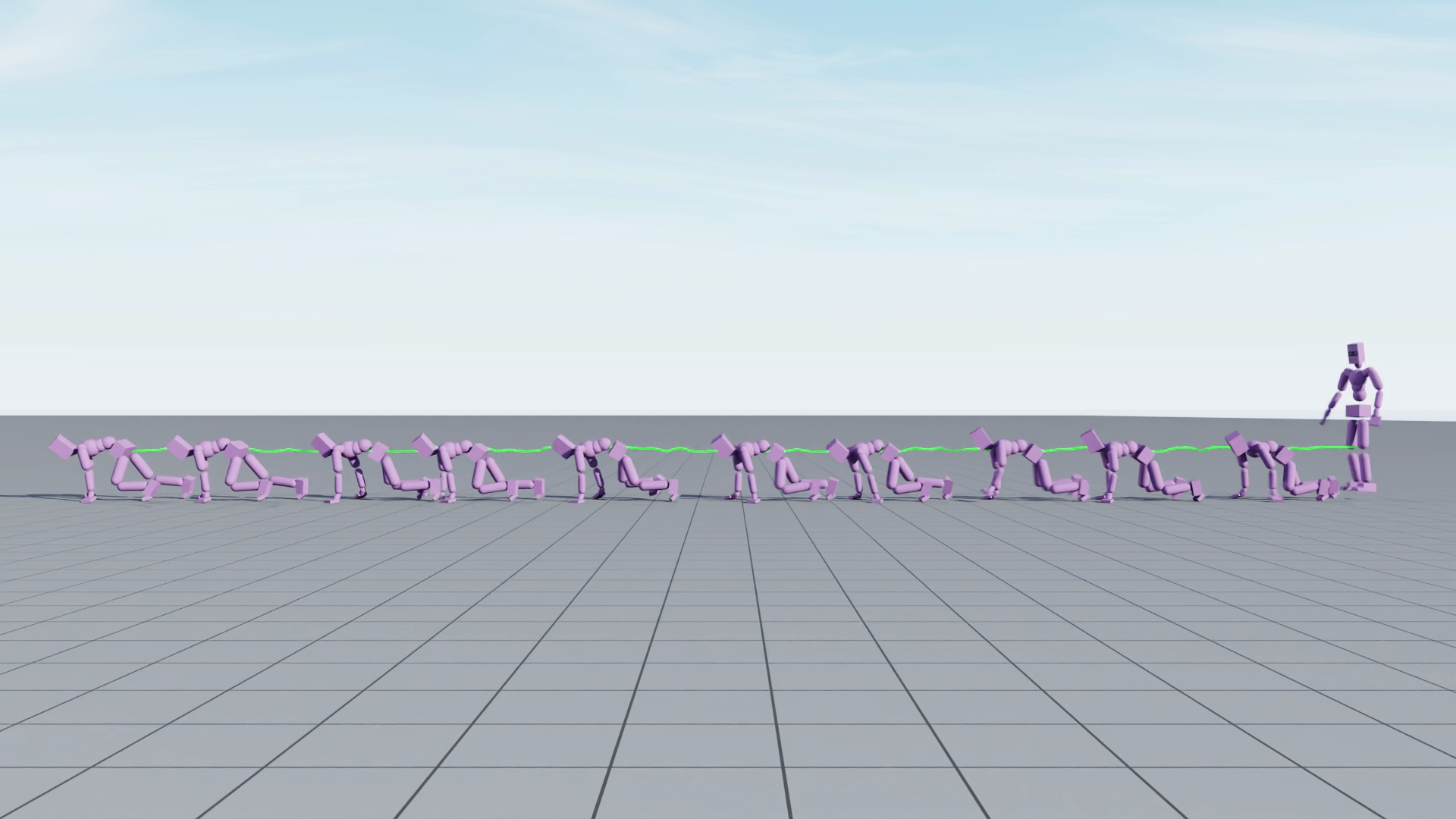}}
         \caption{\textbf{Path-Following:} crawl}
         \label{fig: locomotion crawl}
     \end{subfigure}
     
     \caption{\textbf{Tasks:} \alg~ can be used to solve new tasks across a wide range of terrains by conditioning the model on different user-specified goals.}
    \label{fig: tasks}
\end{figure*}

\begin{table}[]
    \centering
    \caption{\textbf{Tasks:} \alg~ is evaluated on a suite of tasks, where the model is directed to perform each task by conditioning on multi-modal goals.}
    \begin{tabular}{ll|cc}
                                                &           & Success & Error \\ \hline\hline
        \multirow{2}{*}{{Locomotion}}    & Flat      & 96.3\% & 11.2 [cm] \\
                                                & Terrain   & 96.3\% & 12.5 [cm] \\ \hline
        \multirow{2}{*}{{Steering}}       & Flat      & 97.8\% & 8.4 [cm/s] \\
                                                & Terrain   & 93.8\% & 8.4 [cm/s] \\ \hline
        \multirow{2}{*}{{Reach}}         & Flat      & 88.7\% & 20.3 [cm] \\
                                                & Terrain   & 87.3\% & 21.7 [cm]
    \end{tabular}
    \label{tab: tasks}
\end{table}

To solve tasks with \alg~, we utilize "goal-engineering", a process akin to "prompt-engineering" for language models \cite{diab2022stable}. For each task, we construct a simple finite-state-machine (FSM) that transitions between different goals provided to \alg~. At each timestep, the FSM evaluates the current system state and determines and updates the current control scheme. For example, sitting on a chair consists of (1) navigating towards the chair using inbetweening (any-joint-any-time constraints), then, once within range, (2) conditioning on the chair bounding-box representation. By conditioning \alg~ on different goals at each stage of the task, the controller can be directed to perform a wide range of tasks without any task-specific training. The performance of \alg~ on the various tasks is recorded in \cref{tab: tasks}, and performance on object interaction tasks is recorded in \cref{tab: ablation}. For each task, we report the average performance statistics recorded across 5000 random episodes. 

Behaviors produced by our model on various tasks when directed through goal-engineering are shown in \cref{fig: tasks}. \alg~ is able to generate naturalistic motions that follow user-specified goals across a variety of different irregular environments. For the steering task, we can construct a joystick-like controller by simply conditioning the model on target movement and heading directions, and the controller is then able to generate natural motions that follow the given commands. When provided target positions for the hand, our model produces diverse motions to reach different target locations, such as reaching and bending over. By conditioning the model on target head position and orientation ensures, \alg~ tracks a specific target path while also adhering to different height constraints. This then allows users to direct the character to walk up and down a flight of stairs and crawl across flat ground. Text can also be used to stylize the resulting motions. For example, \cref{fig: locomotion walk text} shows that a target path can be specified for the head of the character, with text specifying the style for the rest of the body.

\paragraph{Path-Following} The goal is to follow a given trajectory. A trial is marked as failed if, at any timestep, the character deviates by more than 2m from the goal position. We report the average displacement error for the 3D position in cm. To successfully follow a given target trajectory, consisting of a sequence of waypoint positions, we first compute the target rotation at each timestep using the direction between each two subsequent waypoints. At each timestep, \alg~ is provided the target positions and rotations at the next five timesteps, as well as a target 0.8 seconds into the future to provide the model with information for longer term planning. When the character is more than 0.4m from the target position, we only provide the distant target at 0.8s as input to the model, thereby providing the model with more flexibility in terms of how to move closer to the path. We observed a  tradeoff between user control and success rate. By providing the model with more flexibility in the goal inputs and not tightly constraining the near-term goals, the success rate increases and tracking error decreases, but at the cost of reduced user control.

\paragraph{Steering} This task provides two vectors, corresponding to the requested orientation and movement. The character needs to move in the direction and speed of the movement vector, while facing the orientation vector. We consider a trial as failed if the orientation deviates by more than 45 degrees, and we report the speed error in cm/s, measured along the target direction. Each time the vectors change, the character has 2 seconds before measurement begins. We solve this task using any-joint-any-time control by conditioning on the pelvis rotation and location. To ensure the character moves as requested, we first condition the target rotation alongside the time-left until measurement starts $(\hat{\theta}^\text{root}, \tau-t)$. Then, we condition the root rotation and offset 1 second into the future $(\hat{\theta}^\text{root}, \hat{p}^\text{root}, 1[sec])$, with $\hat{p}^\text{root} = [p\_x^\text{root} + v\_x, p\_y^\text{root} + v\_y, 0.9]$, to ensure it remains upright.

\paragraph{Reach} This task presents a problem of inbetweening. When the target position changes, the character has 2 seconds to reach the new position. The entire motion inbetween remains unspecified. A trial is considered as a fail if the hand deviates by more than 0.5m at the target timestep. We also report the average distance in cm, measured over the time the hand should remain in place. To solve this task, we only condition the model on the target position for the right hand, along with the remaining time to reach the target, clipped to a maximum of 1/6 seconds. We found that if the target is set at the immediate next frames, \alg~ is less successful at recovering when losing balance. However, by providing a goal further into the future, this provides \alg~ with more flexibility. The provided flexibility leads to more realistic and robust behaviors.

\begin{figure*}[t]
     \centering
     \begin{subfigure}[b]{0.245\textwidth}
         \centering
         \scalebox{-1}[1]{\includegraphics[trim={4cm 3cm 4cm 2cm},clip,width=0.33\textwidth]{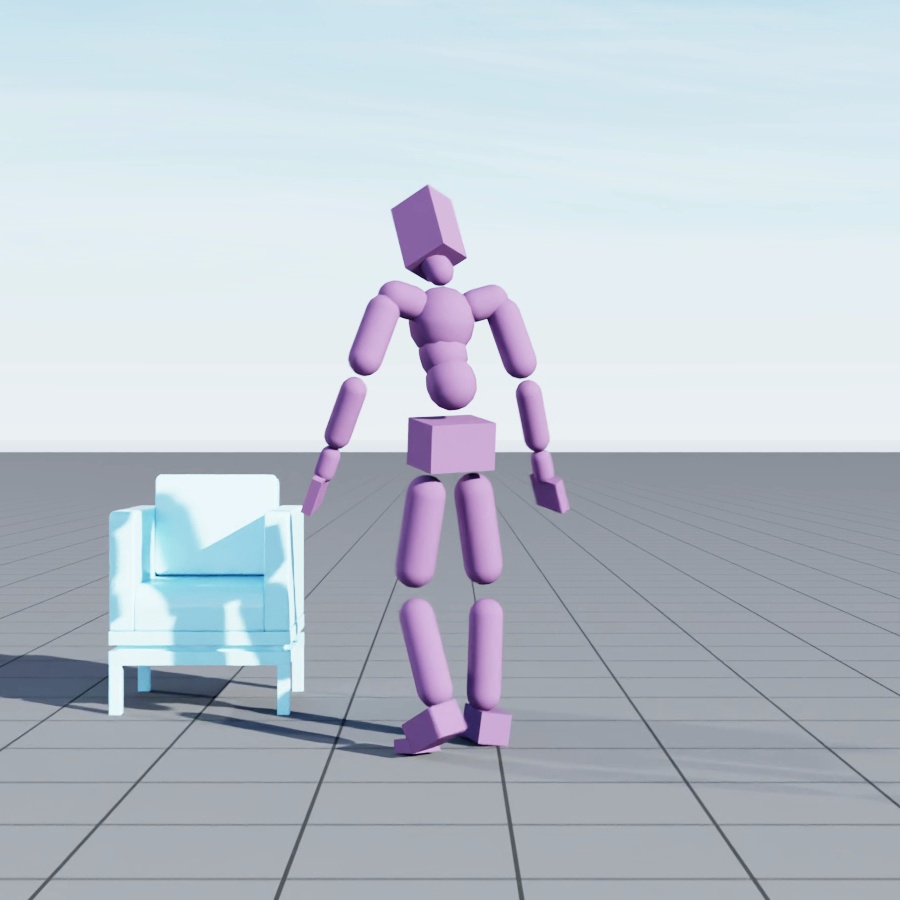}}\hfill
         \scalebox{-1}[1]{\includegraphics[trim={4cm 3cm 4cm 2cm},clip,width=0.33\textwidth]{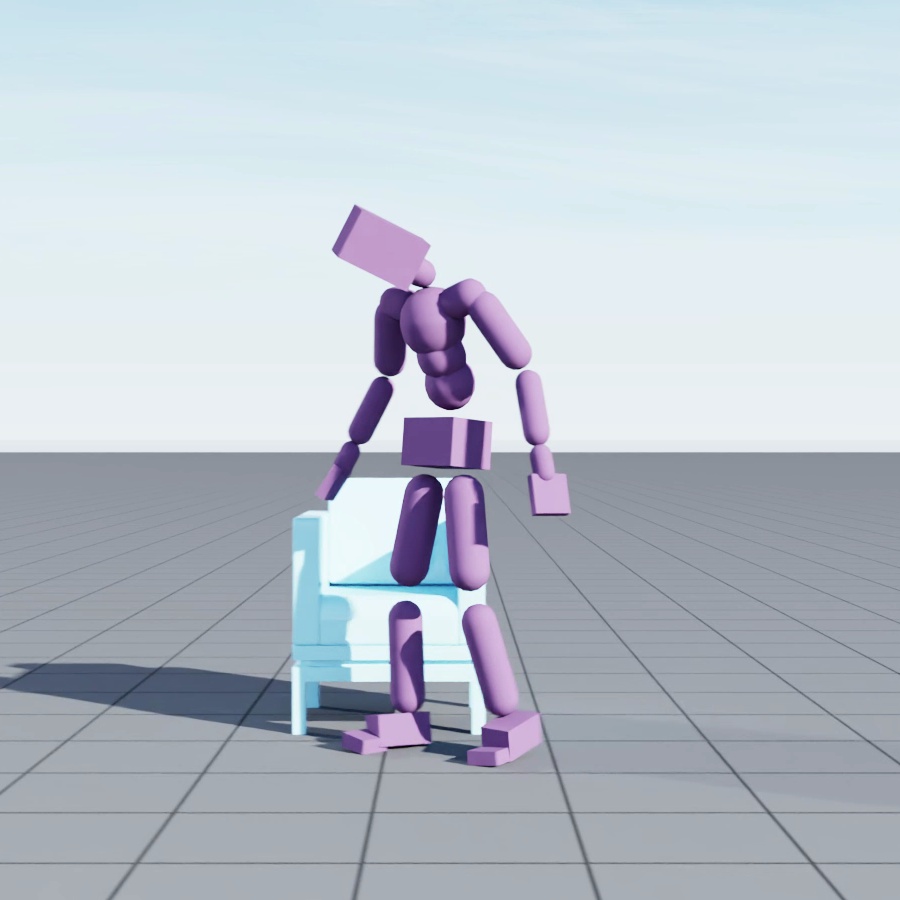}}\hfill
         \scalebox{-1}[1]{\includegraphics[trim={4cm 3cm 4cm 2cm},clip,width=0.33\textwidth]{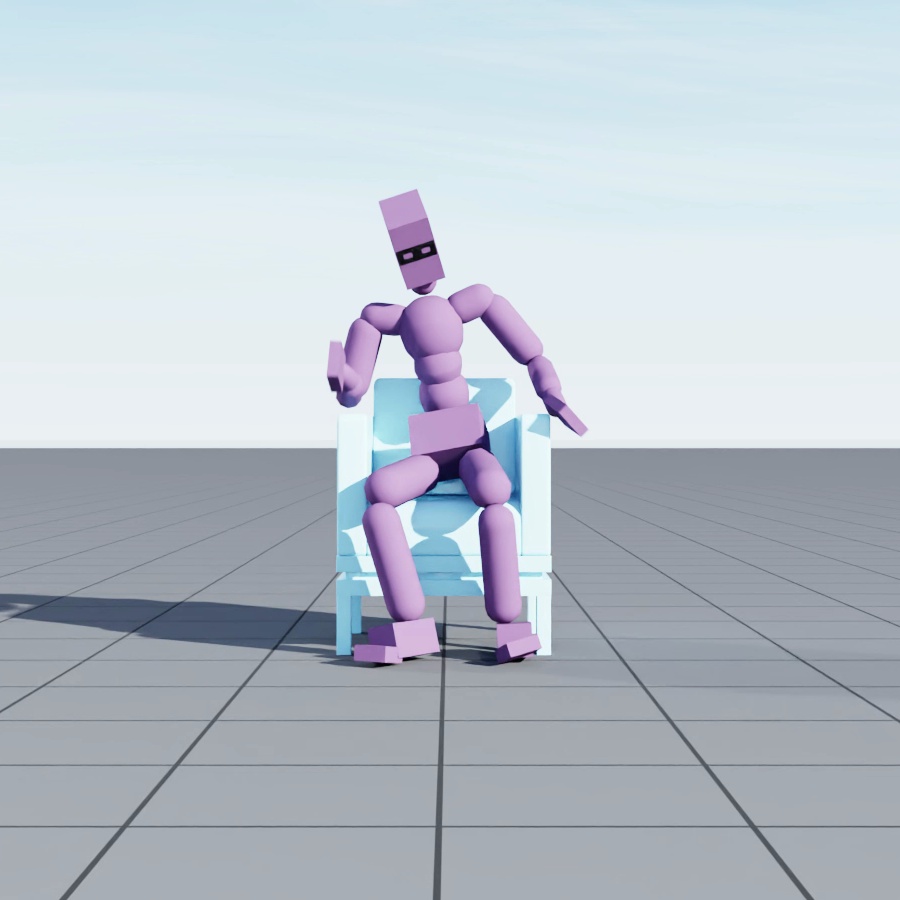}}
         \caption{Armchair}
         \label{fig: 1_couch}
     \end{subfigure}
     \begin{subfigure}[b]{0.245\textwidth}
         \centering
         \scalebox{-1}[1]{\includegraphics[trim={4cm 3cm 4cm 2cm},clip,width=0.33\textwidth]{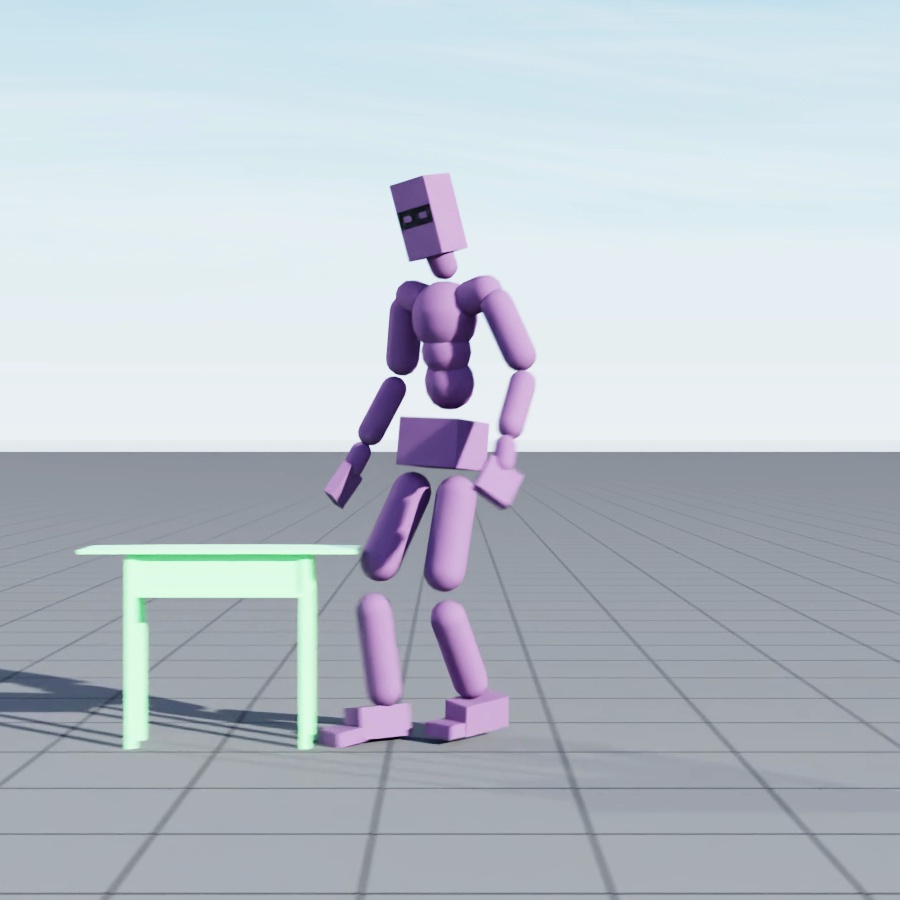}}\hfill
         \scalebox{-1}[1]{\includegraphics[trim={4cm 3cm 4cm 2cm},clip,width=0.33\textwidth]{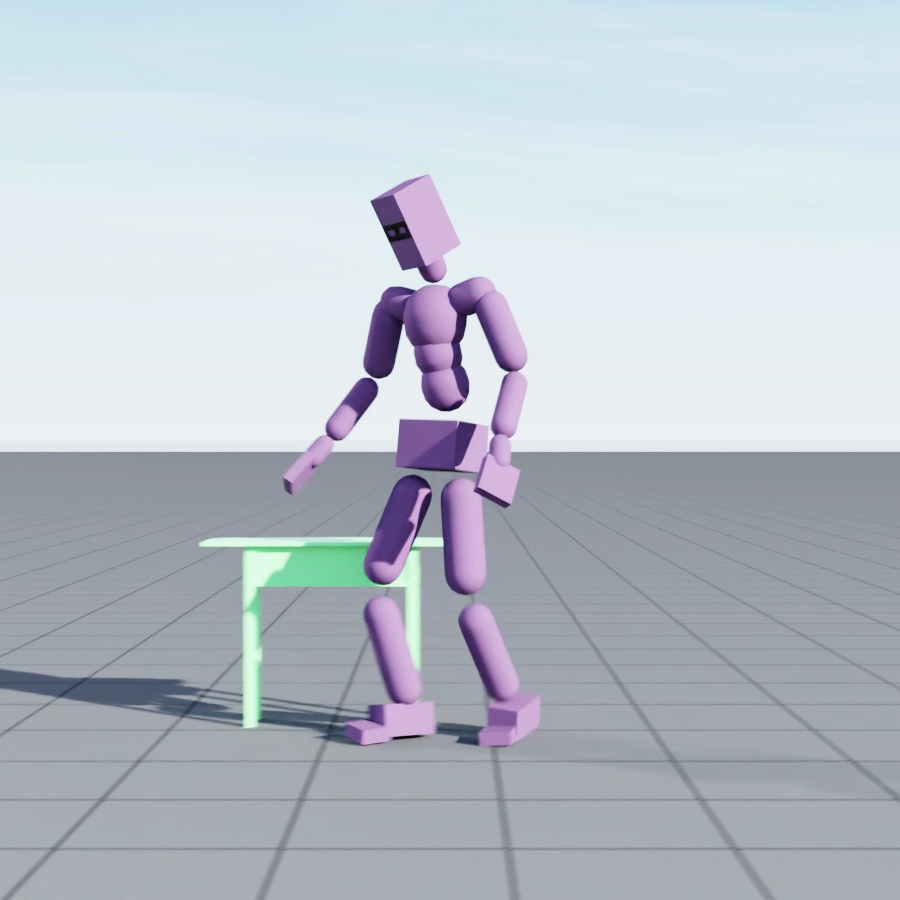}}\hfill
         \scalebox{-1}[1]{\includegraphics[trim={4cm 3cm 4cm 2cm},clip,width=0.33\textwidth]{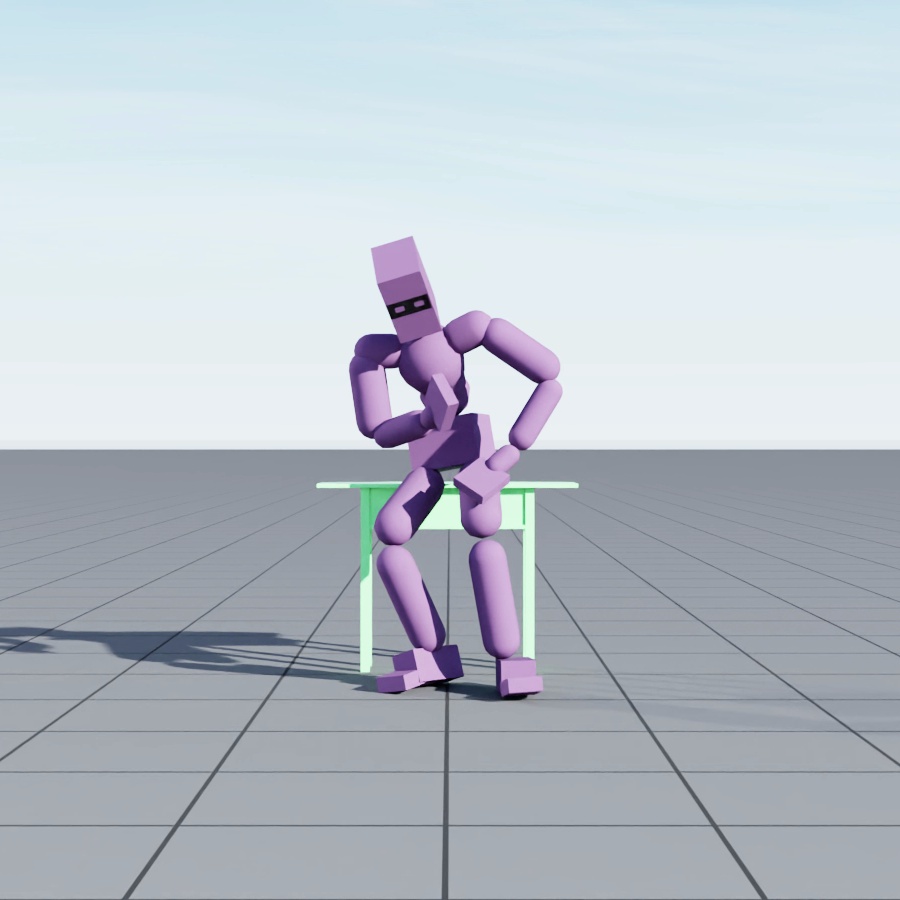}}
         \caption{Table}
         \label{fig: 10_table}
     \end{subfigure}
     \begin{subfigure}[b]{0.245\textwidth}
         \centering
         \includegraphics[trim={4cm 3cm 4cm 2cm},clip,width=0.33\textwidth]{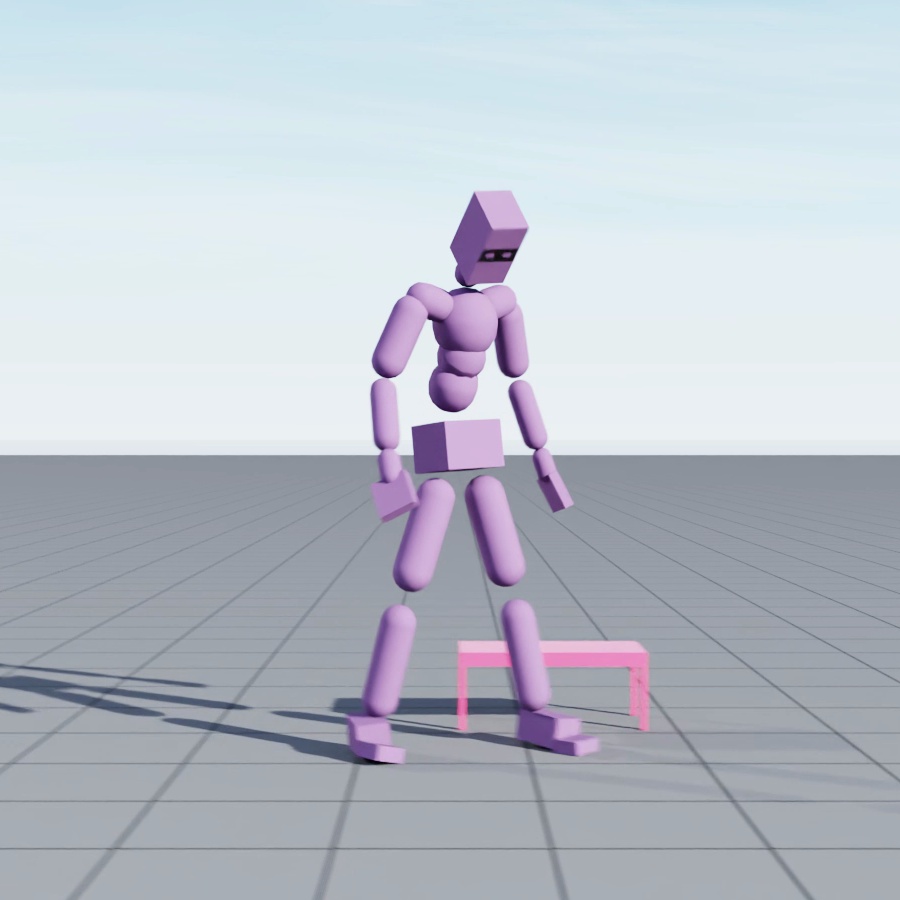}\hfill
         \includegraphics[trim={4cm 3cm 4cm 2cm},clip,width=0.33\textwidth]{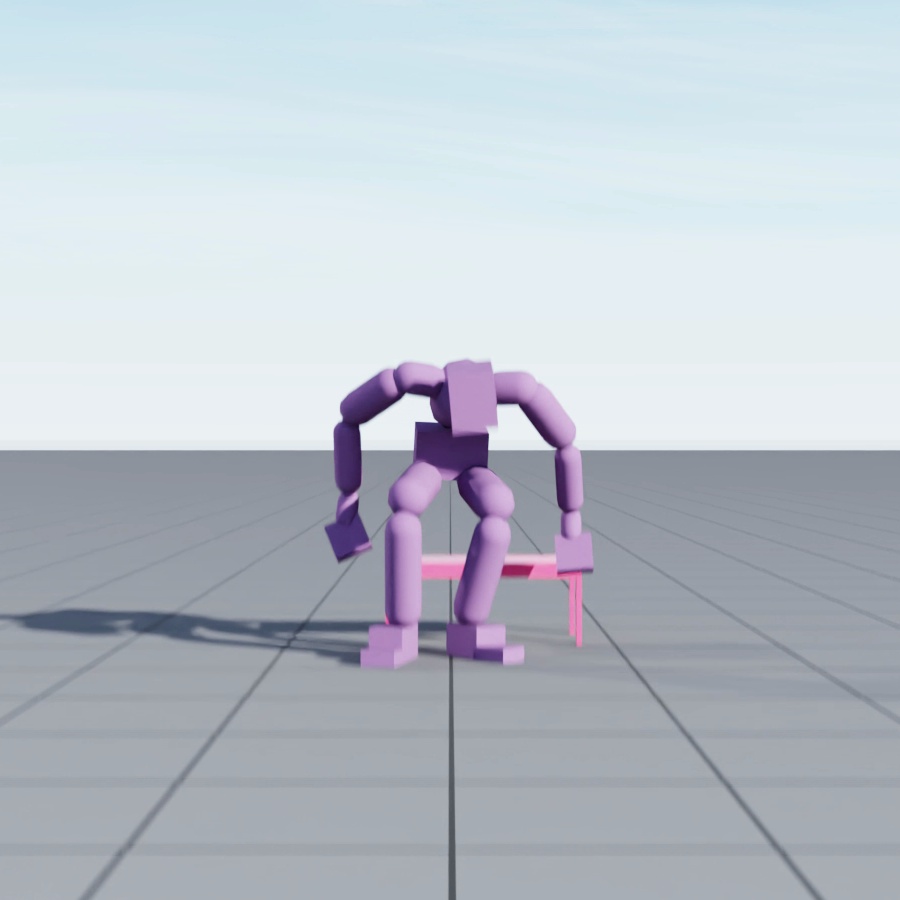}\hfill
         \includegraphics[trim={4cm 3cm 4cm 2cm},clip,width=0.33\textwidth]{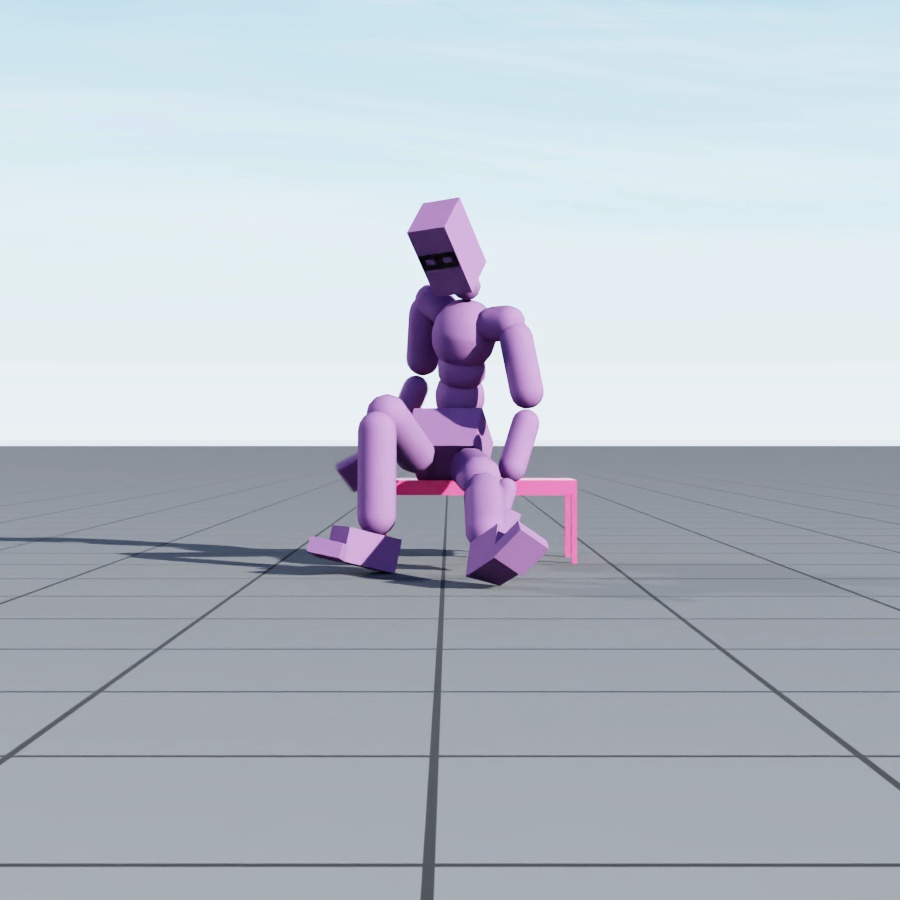}
         \caption{Stool}
         \label{fig: 15_lowstool}
     \end{subfigure}
     \begin{subfigure}[b]{0.245\textwidth}
         \centering
         \scalebox{-1}[1]{\includegraphics[trim={4cm 3cm 4cm 2cm},clip,width=0.33\textwidth]{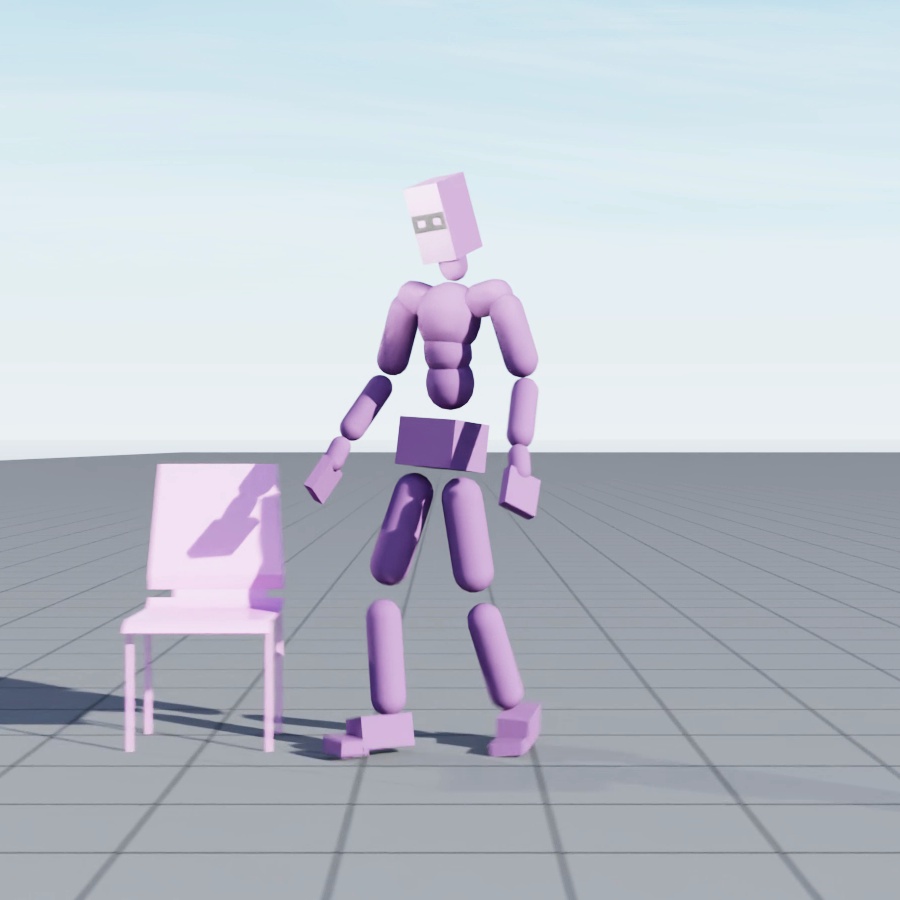}}\hfill
         \scalebox{-1}[1]{\includegraphics[trim={4cm 3cm 4cm 2cm},clip,width=0.33\textwidth]{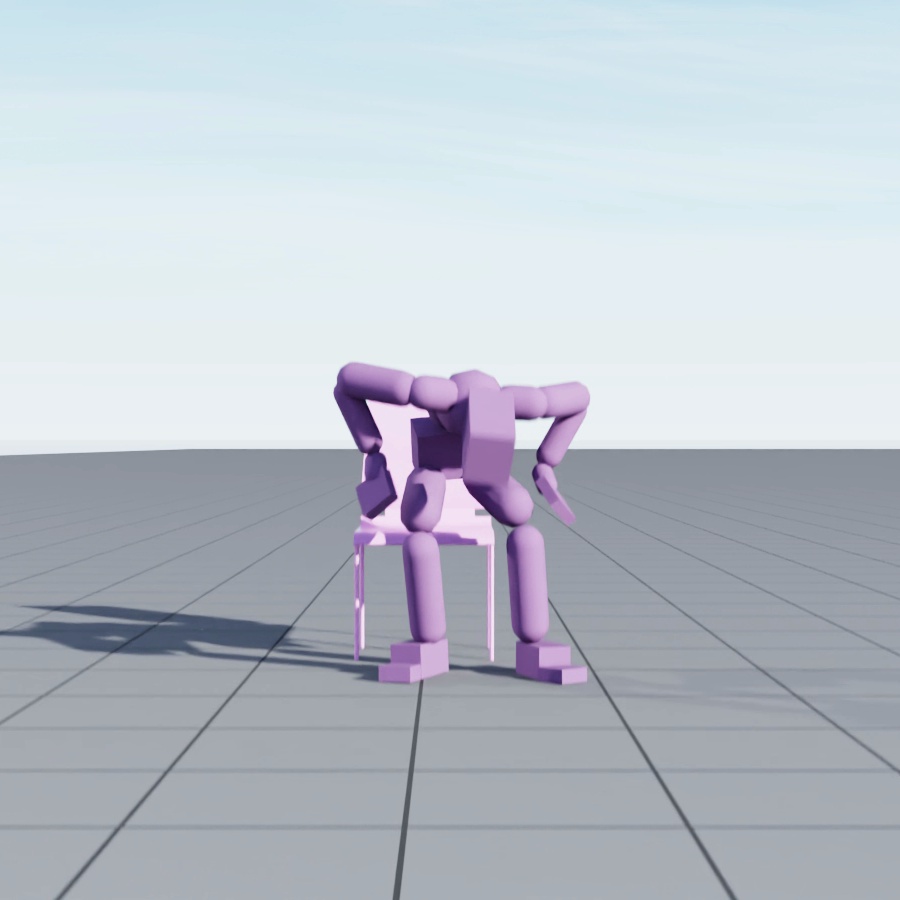}}\hfill
         \scalebox{-1}[1]{\includegraphics[trim={4cm 3cm 4cm 2cm},clip,width=0.33\textwidth]{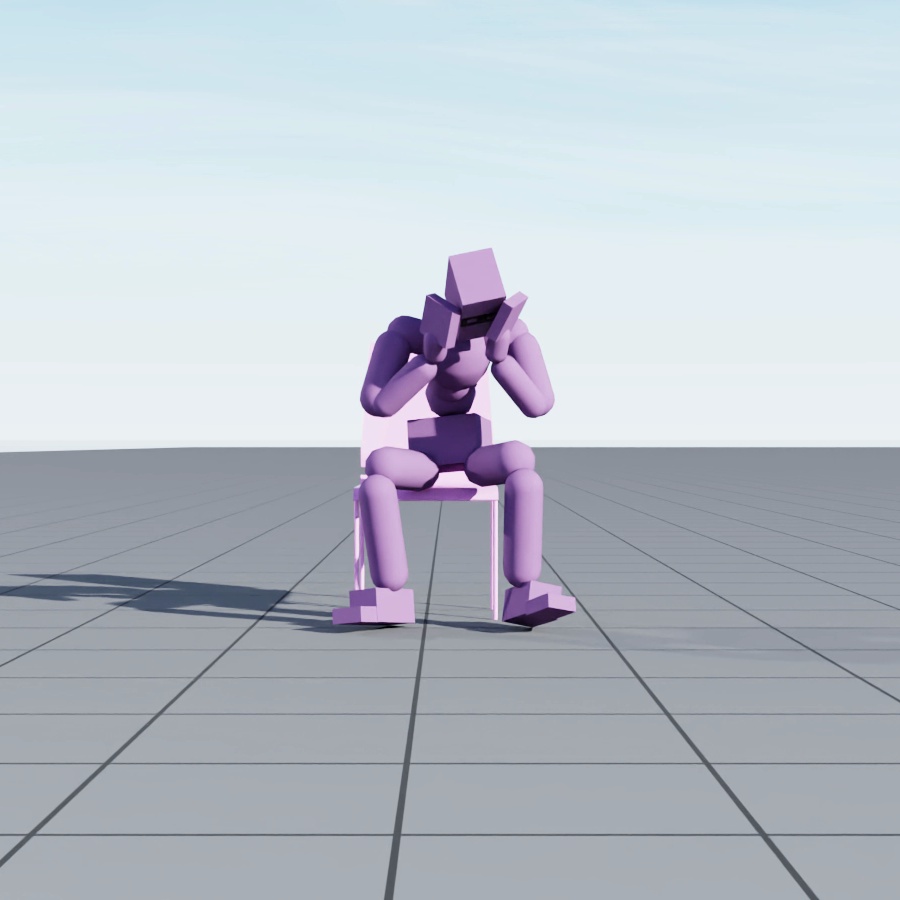}}
         \caption{Chair}
         \label{fig: 11_chair}
     \end{subfigure}\\

    \begin{subfigure}[b]{0.99\textwidth}
         \centering
         \includegraphics[width=0.2475\textwidth]{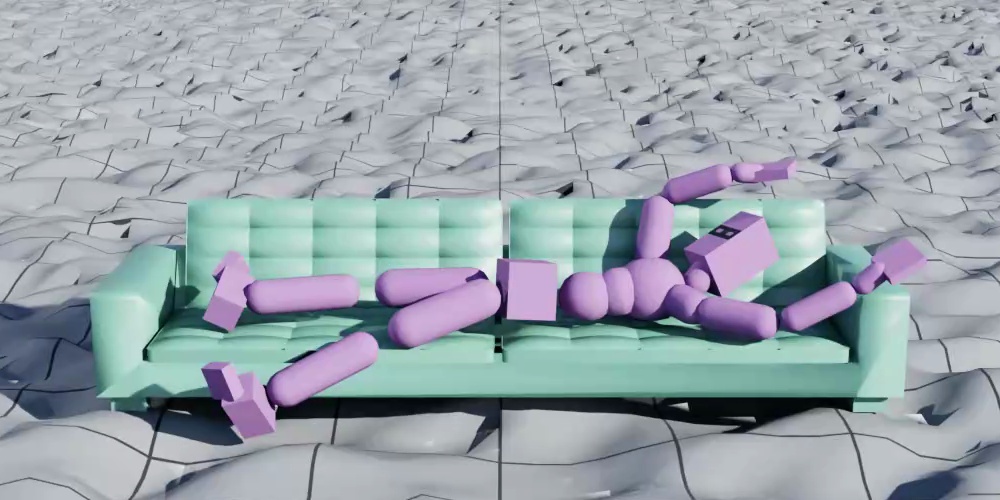}\hfill
         \includegraphics[width=0.2475\textwidth]{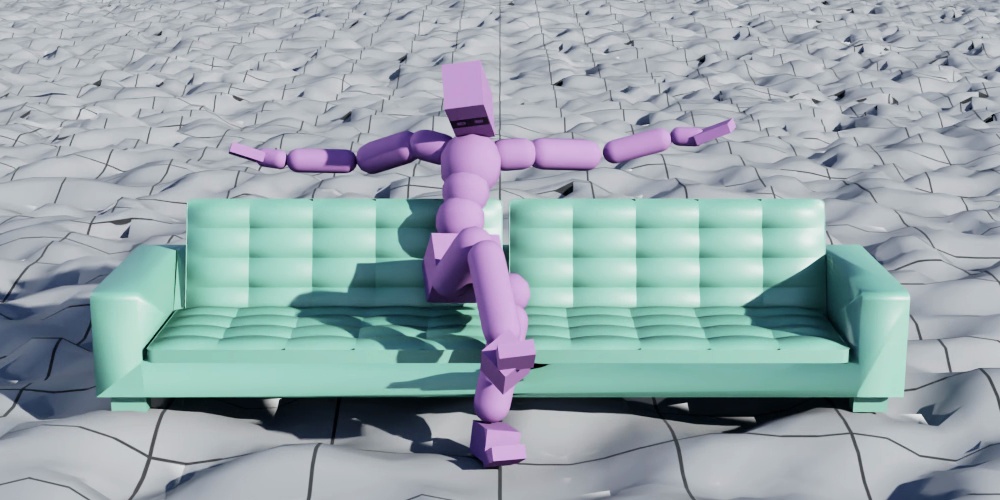}\hfill
         \includegraphics[width=0.2475\textwidth]{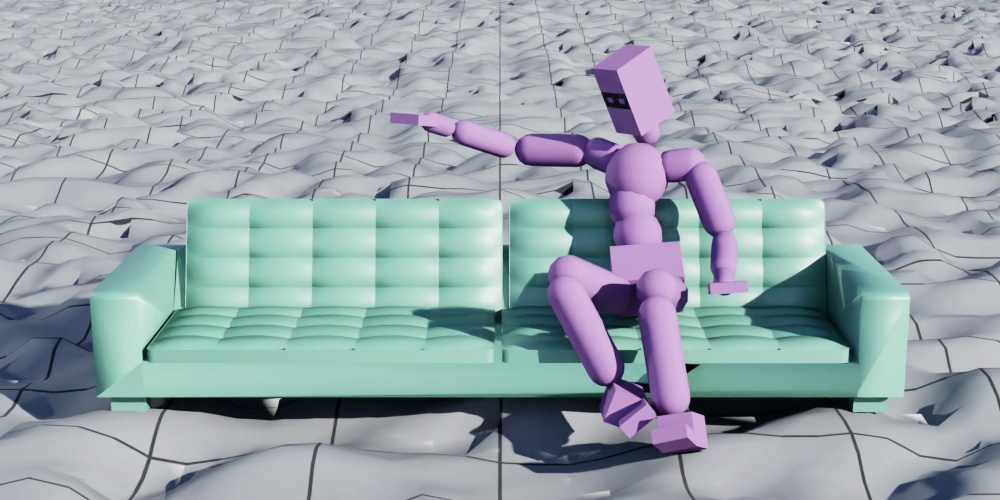}\hfill
         \includegraphics[width=0.2475\textwidth]{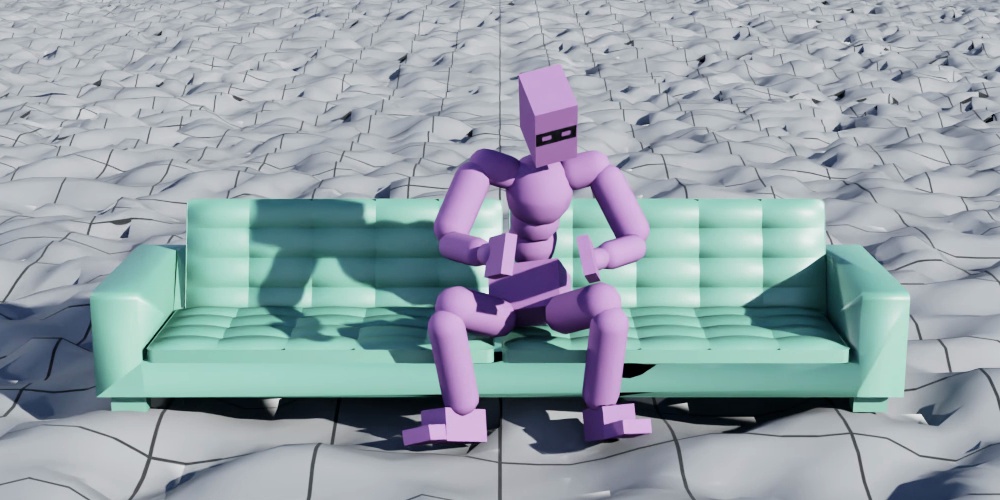}
         \caption{Sofa}
         \label{fig: sofa terrain}
     \end{subfigure}
     
     \caption{\textbf{Objects:} By conditioning \alg~ on the bounding-box of an object, our model is able to produce diverse motions for approaching and interacting with a wide variety of test objects. In \cref{fig: sofa terrain}, the character successfully generalizes to interacting with a sofa placed on irregular terrain, a scenario that was not observed during training.}
    \label{fig: objects}
\end{figure*}

\subsection{Object Interaction and Ablation} 
The previous tasks were solved by leveraging any-joint-any-time constraints. In this task, the character is spawned at a random location, far from an object. The goal is to reach the object and sit on it. To tackle this type of task, we combine three control modalities: any-joint-any-time, text, and object conditioning. First, when the character is further than 2 meters from the target object, we utilize any-joint-any-time control. The character is conditioned on a goal direction for moving towards the object at a speed of 1[m/s]. While moving towards the object, it is also provided a text command of ``the person walks normally". These two commands allow the character to walk in a stable and natural manner towards the provided object. Then, once within 2 meters of the object, the target direction and text command are removed. From this point, the controller is conditioned on the object's bounding box. Conditioning on the object leads the model to generate a motion for interacting with the object (e.g. sitting). \alg~ is able to seamless transition between different controls and generate natural interactions with a target object. The object interaction motions in the SAMP dataset \cite{hassan2021stochastic} consist of a person walking towards an object and sitting down on it. As such, we find that by simply providing the character with the object representation was sufficient for it to generate a natural object interaction motion.

\cref{fig: objects} shows examples of motions generated by \alg \ when interacting with chairs and sofas. These objects are taken from a test set, not observed during training. In \cref{fig: 1_couch,fig: 10_table,fig: 15_lowstool,fig: 11_chair}, we observe diverse motions for approaching  objects of varying shapes and sizes. Then, in \cref{fig: sofa terrain}, we place a sofa on rough terrain. Although this combination of an object with irregular terrain was not observed during training, \alg~ succeeds in moving to the object and sitting/lying down. Notably, \alg~ does not produce a single solution. The VAE architecture is designed to model multiple possible solutions for a given set of constraints, enabling the model to generate diverse interactions for a given object. As seen in \cref{fig: sofa terrain}, \alg~ sits on diverse locations and poses.

\begin{table}[]
    \centering
    \caption{\textbf{Objects + ablation:} We evaluate \alg~ and conduct an ablation on various design decisions. Experiments are conducted on the sitting task with a set of test objects. We evaluate versions of the model with key components removed (\cref{sec: sparsemimic}), and measure the impact on the average success rate and error (i.e. average minimal distance from a valid sitting position on the object).}
    \tabcolsep=0.14cm
    \begin{tabular}{l|cc}
                        & Success & Error [cm] \\ \hline\hline

        {\alg} (ours)   & 96.9\% & 10.5 \\ \hline
        {No history} & 94.9\% & 12.7 \\ \hline
        {No VAE} & 93.2\% & 12.2 \\ \hline
        {No residual prior} & 21.1\% & 57.4 \\ \hline
        {No structured masking} & 0\% & 274.4
    \end{tabular}
    \label{tab: ablation}
\end{table}

We report the performance statistics in \cref{tab: ablation}. As solving object interactions combines multiple control modalities, and requires precise scene awareness, this task also serves as an ablation study. In our ablation, we compare several core design decisions. A trial is considered successful if the character successfully sits on the object during the episode, e.g., pelvis reaches within 20cm of a valid seating position. In addition, we report the average minimal distance between the object and the character's pelvis. When evaluating the impact of various design decisions, we find that: 
(1) Typically, motions start standing still. By providing the \textit{historical poses}, \alg~ is less likely to get stuck in place. (2) The \textit{VAE} structure ensures a more robust controller by providing a better way of encoding the diversity of solutions. (3) The inductive bias present in the residual architecture \cite{yao2022controlvae} ensures the prior properly controls the latent space encoding. A non-residual prior proves incapable of controlling the motions. Finally, (4) having a structured masking mechanism is crucial. This mechanism ensures the model observes repeating joints and sometimes only high-level commands.

\begin{figure*}[t]
     \centering
     \begin{subfigure}[b]{0.247\textwidth}
         \centering
         \scalebox{-1}[1]{\includegraphics[width=0.4995\textwidth]{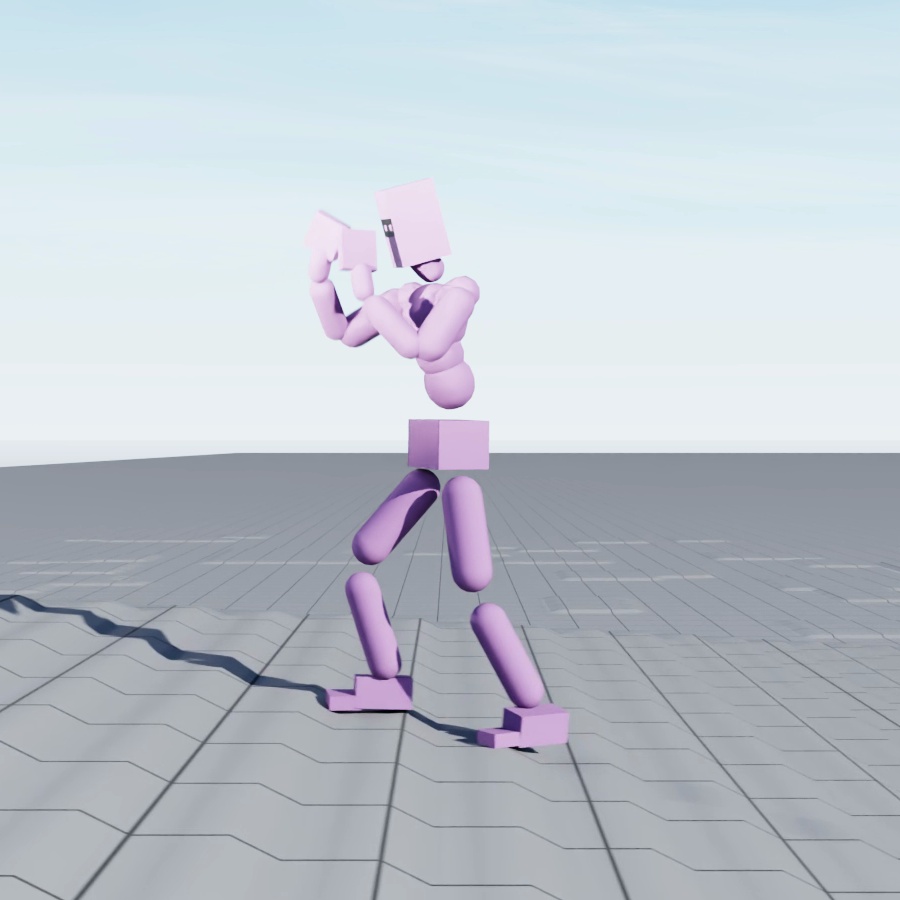}}\hfill
         \scalebox{-1}[1]{\includegraphics[width=0.4995\textwidth]{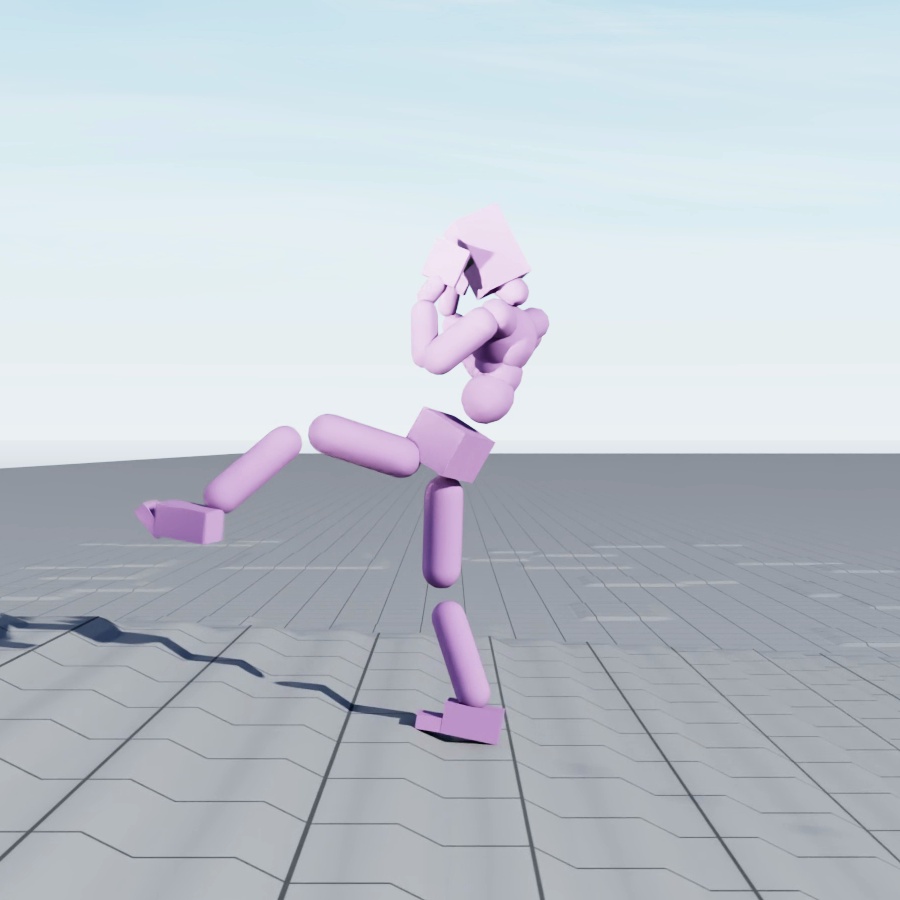}}
         \caption{"a person kicking forward"\\~}
         \label{fig: kick}
     \end{subfigure}
     \begin{subfigure}[b]{0.247\textwidth}
         \centering
         \includegraphics[width=0.4995\textwidth]{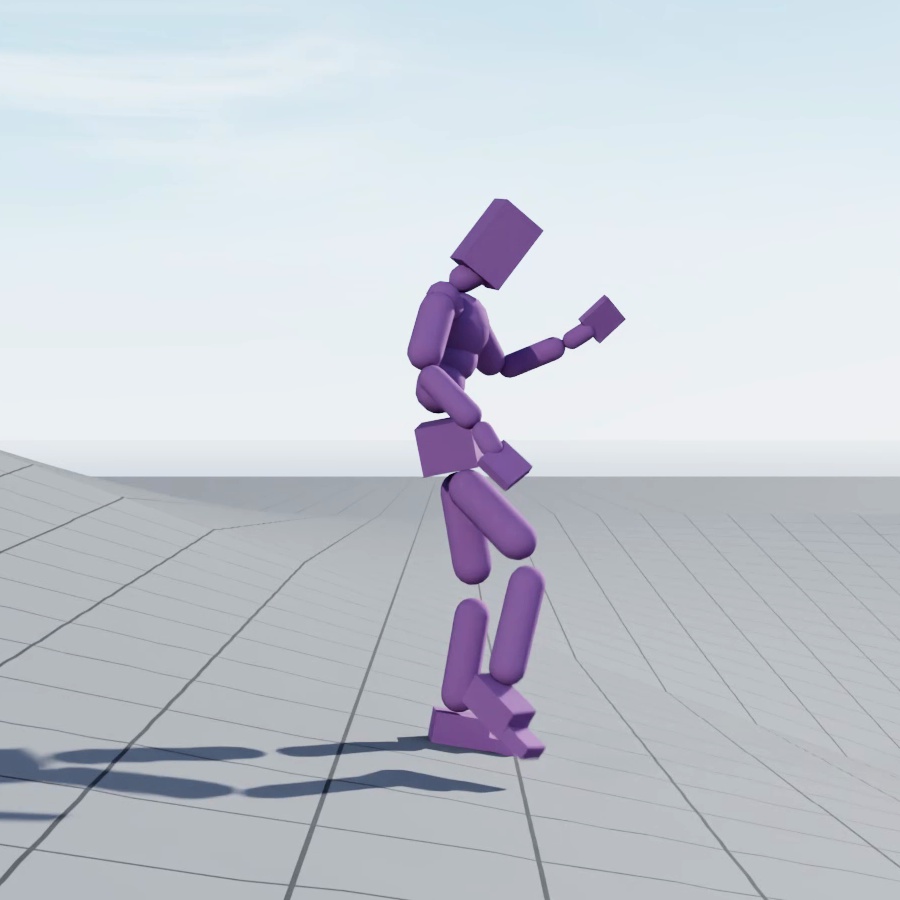}\hfill
         \includegraphics[width=0.4995\textwidth]{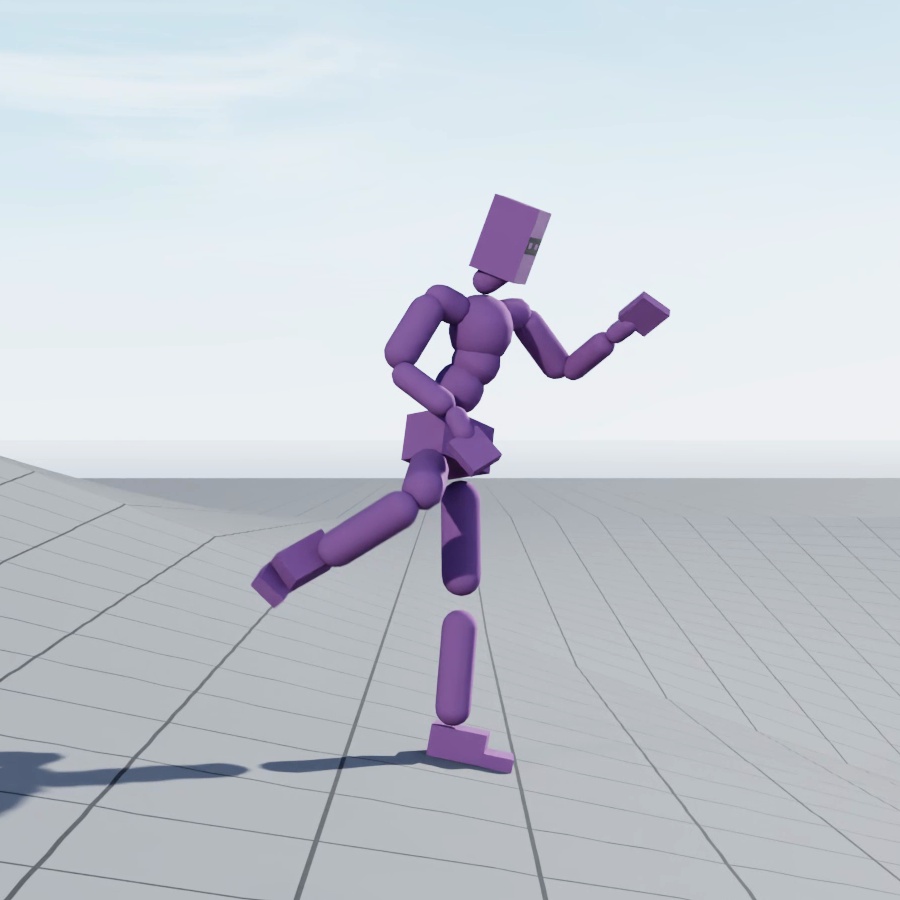}
         \caption{"a person standing on one leg trying to balance himself"}
         \label{fig: oneleg}
     \end{subfigure}
     \begin{subfigure}[b]{0.247\textwidth}
         \centering
         \includegraphics[width=0.4995\textwidth]{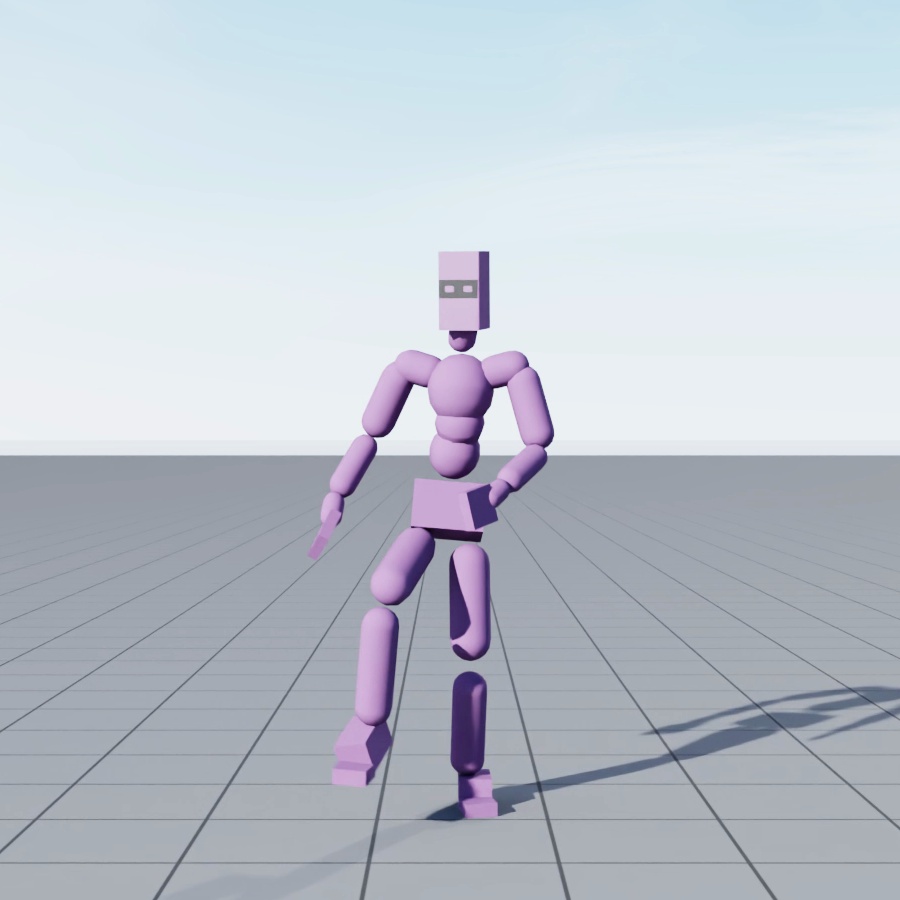}\hfill
         \includegraphics[width=0.4995\textwidth]{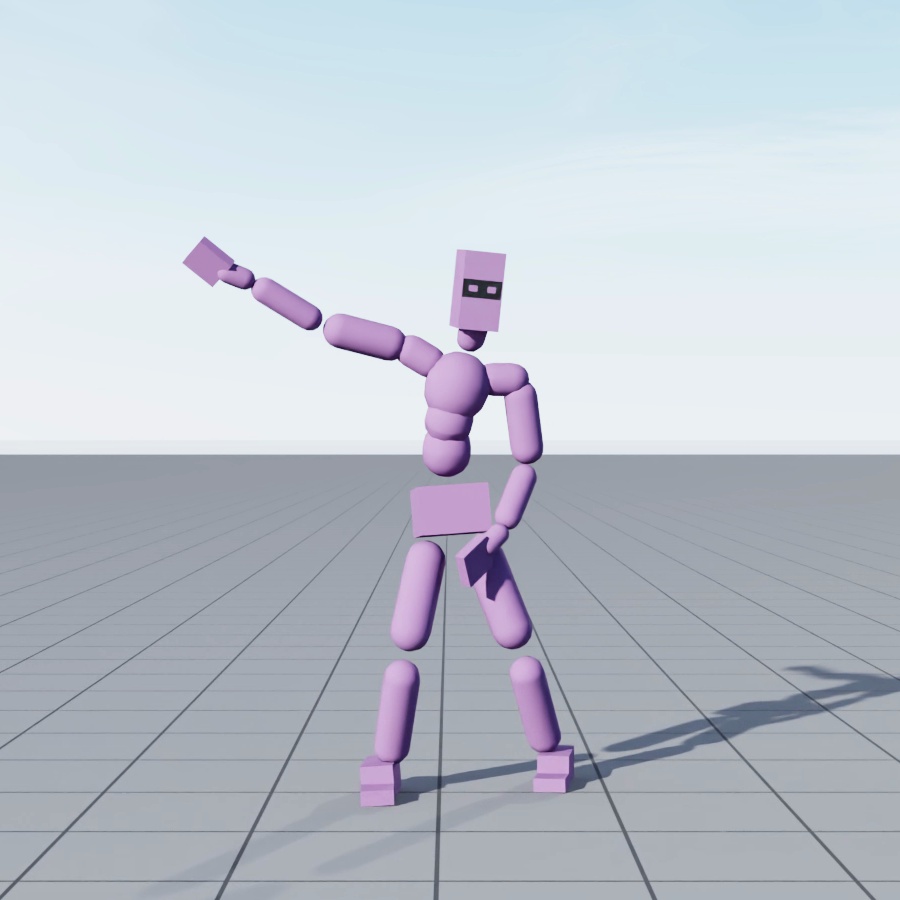}
         \caption{"a person dances like michael \\jackson"}
         \label{fig: michaeljackson}
     \end{subfigure}
     \begin{subfigure}[b]{0.247\textwidth}
         \centering
         \scalebox{-1}[1]{\includegraphics[width=0.4995\textwidth]{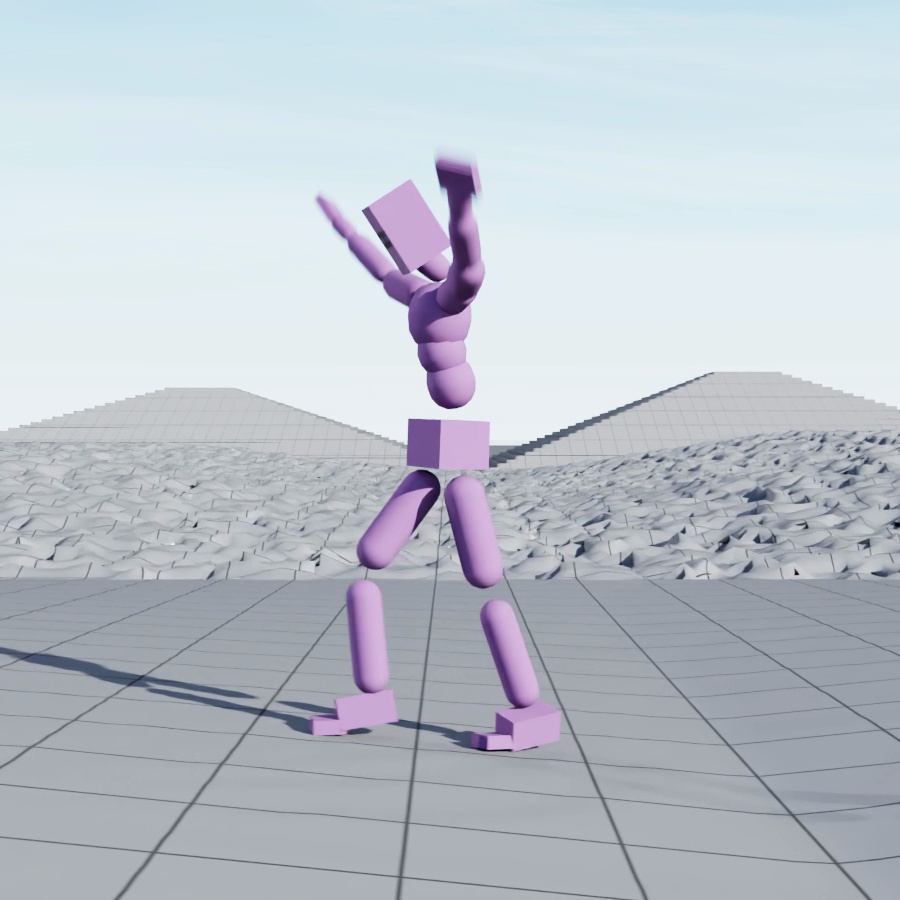}}\hfill
         \scalebox{-1}[1]{\includegraphics[width=0.4995\textwidth]{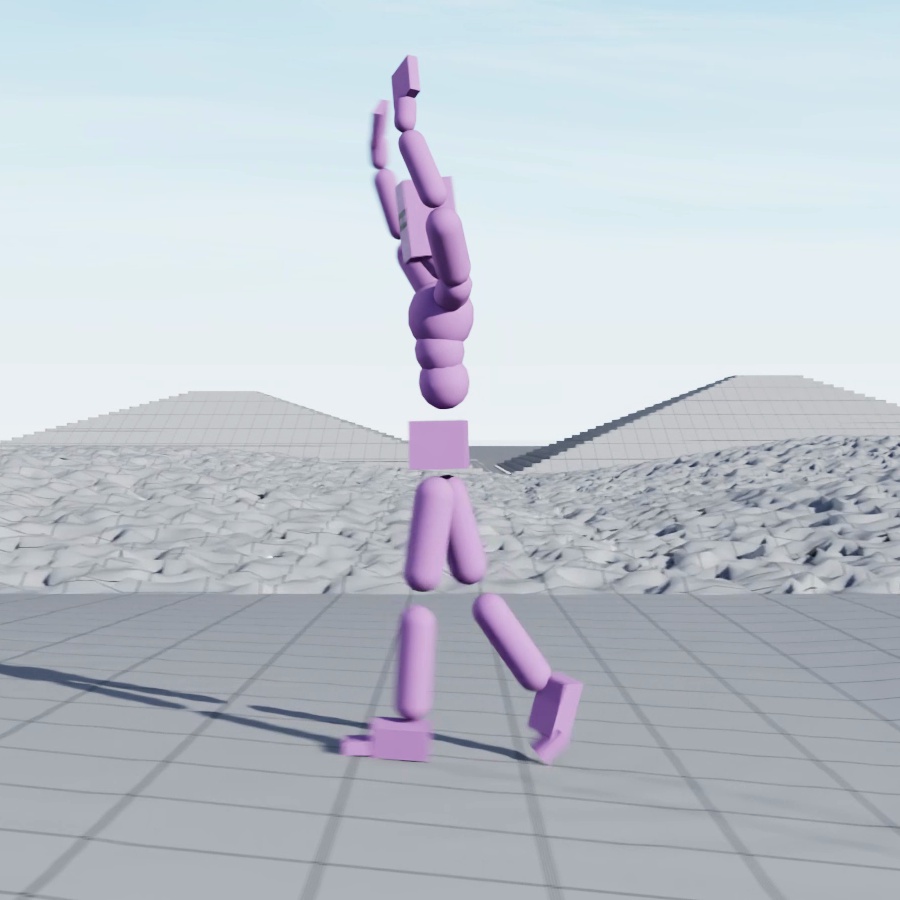}}
         \caption{"a person raises both hands in the air and walks forward"}
         \label{fig: raisehands}
     \end{subfigure}\\
     \begin{subfigure}[b]{0.247\textwidth}
         \centering
         \includegraphics[width=0.4995\textwidth]{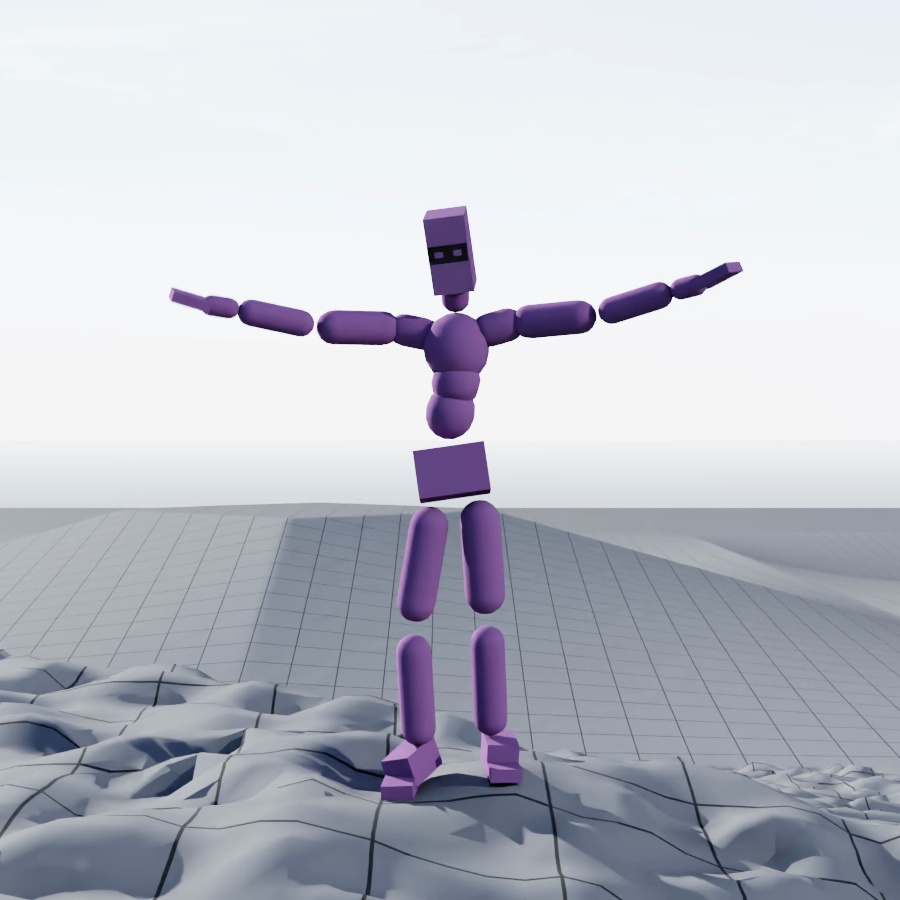}\hfill
         \includegraphics[width=0.4995\textwidth]{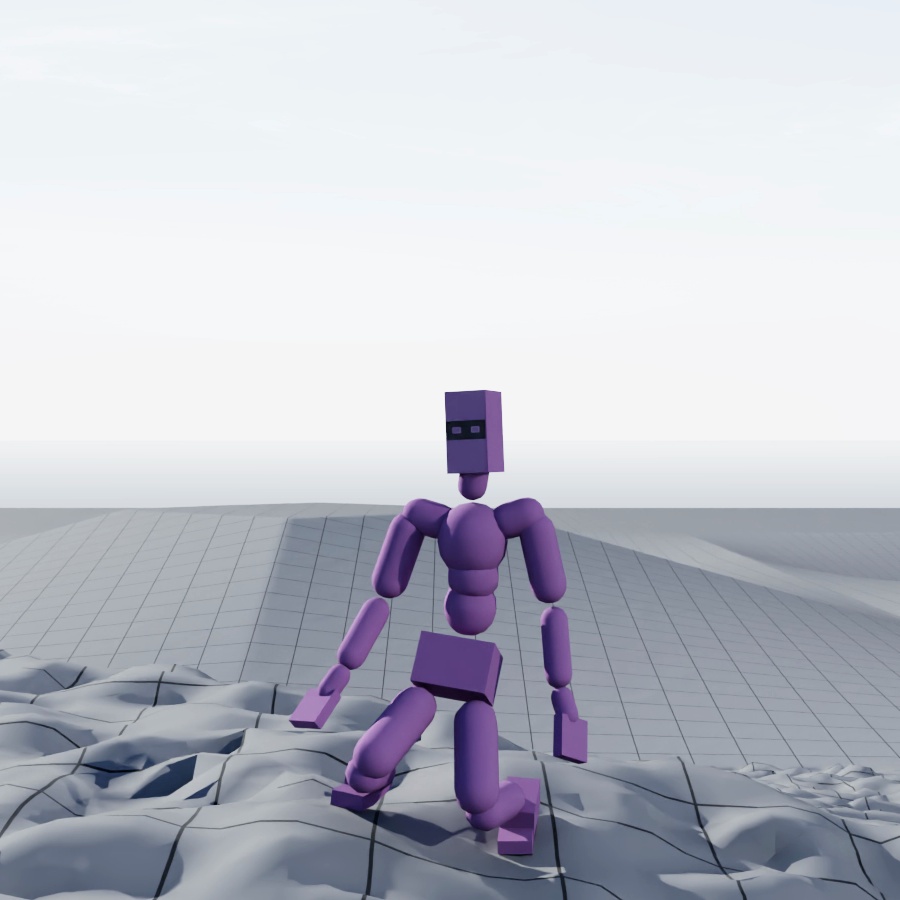}
         \caption{"a person lifts both arms to the sides then steps forward and kneels down"}
         \label{fig: kneel}
     \end{subfigure}
     \begin{subfigure}[b]{0.247\textwidth}
         \centering
         \includegraphics[width=0.4995\textwidth]{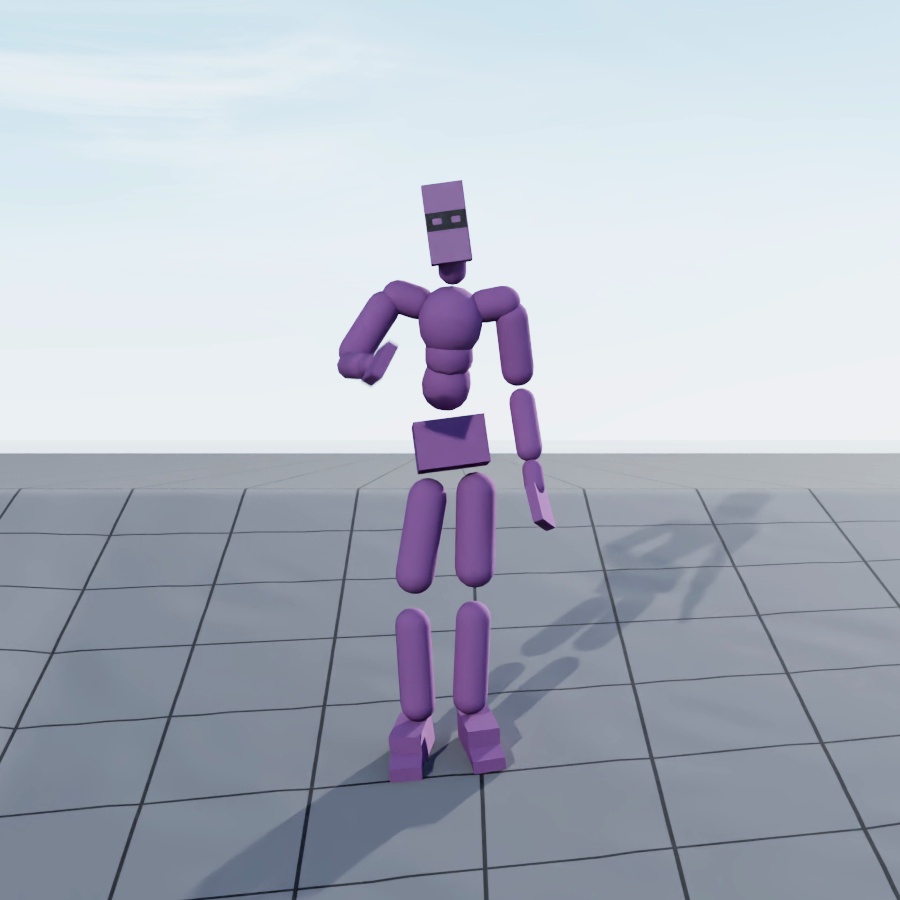}\hfill
         \includegraphics[width=0.4995\textwidth]{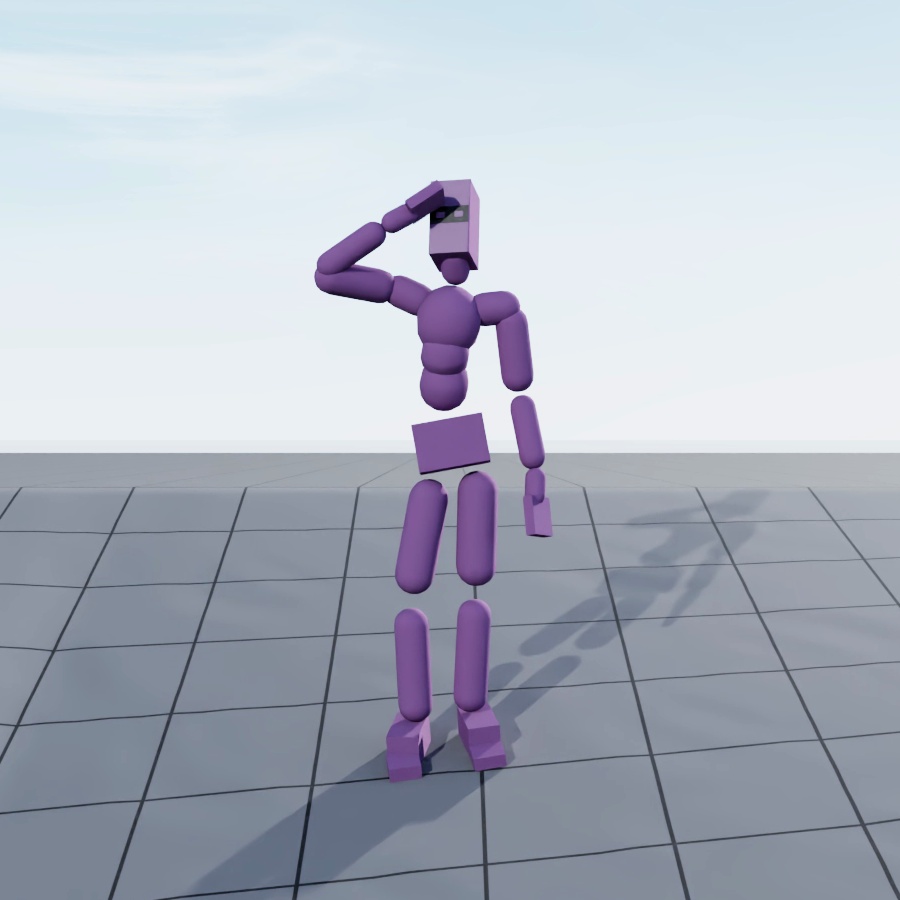}
         \caption{"a person raises their right arm in a salute"}
         \label{fig: salut}
     \end{subfigure}
     \begin{subfigure}[b]{0.247\textwidth}
         \centering
         \includegraphics[width=0.4995\textwidth]{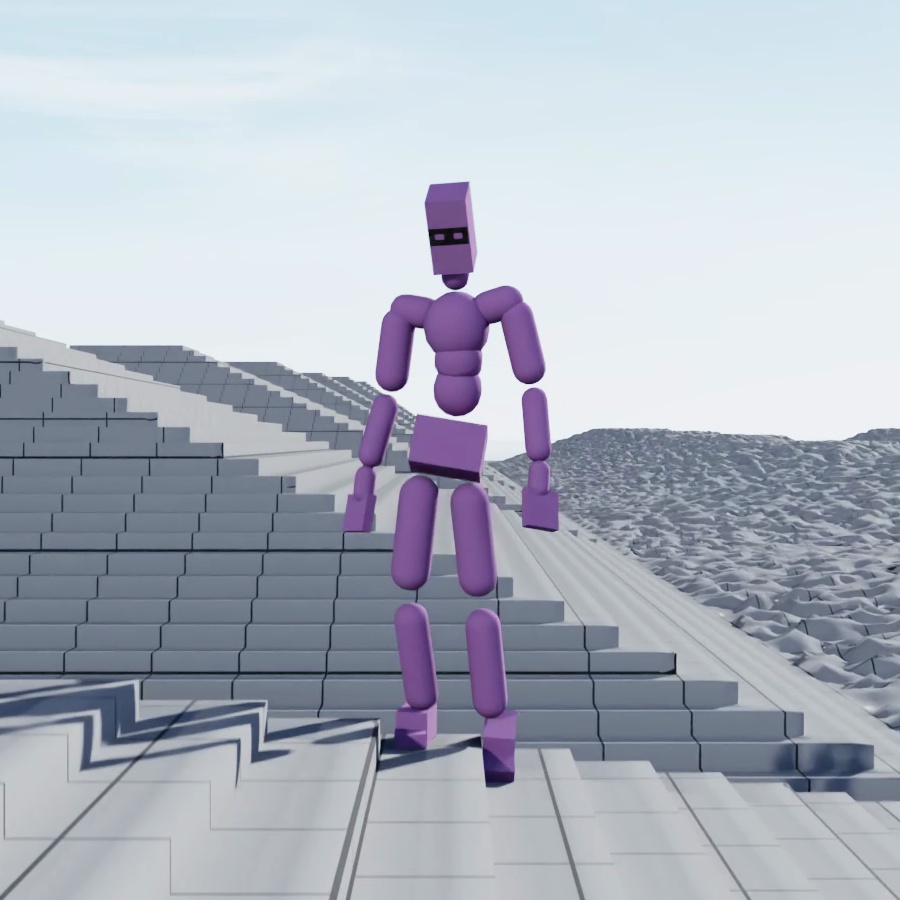}\hfill
         \includegraphics[width=0.4995\textwidth]{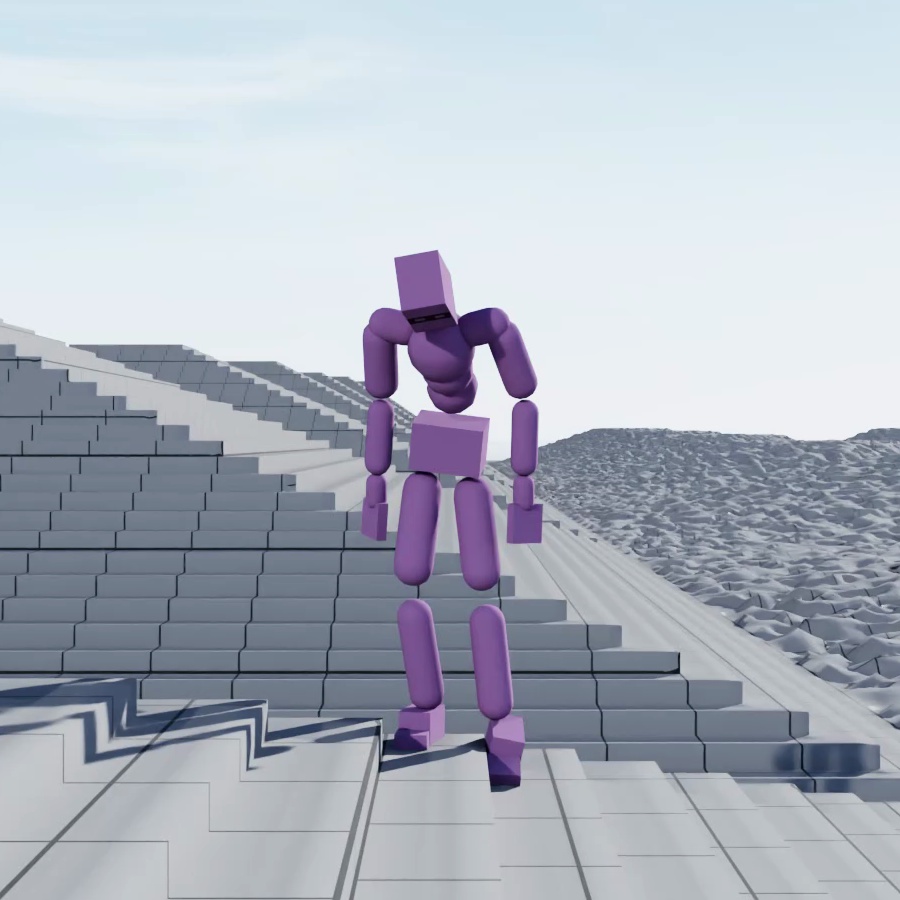}
         \caption{"a man bows very slowly"\\~}
         \label{fig: bow}
     \end{subfigure}
     \begin{subfigure}[b]{0.247\textwidth}
         \centering
         \scalebox{-1}[1]{\includegraphics[width=0.4995\textwidth]{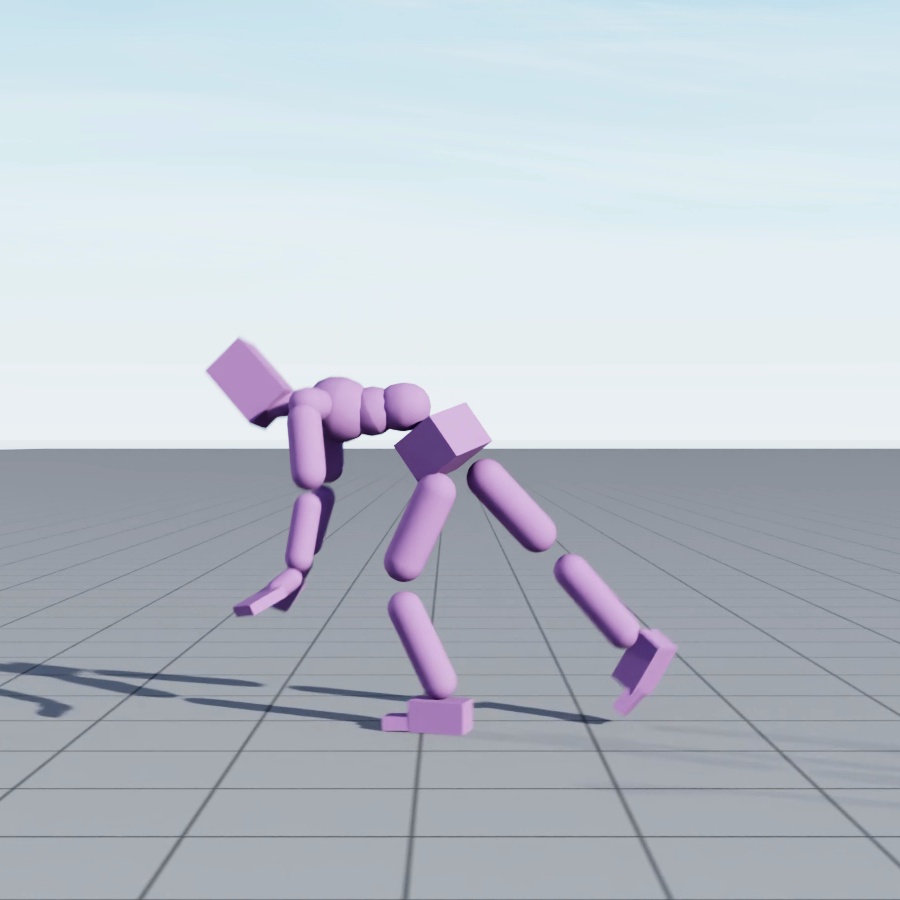}}\hfill
         \scalebox{-1}[1]{\includegraphics[width=0.4995\textwidth]{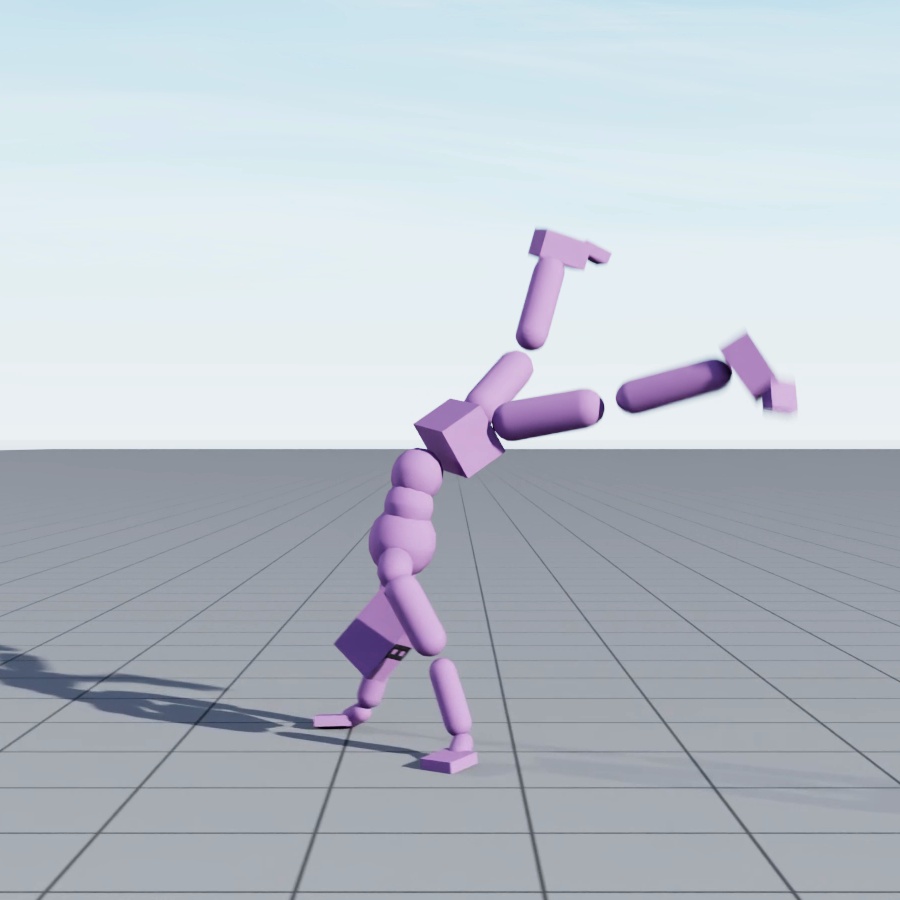}}
         \caption{"a man steps forward and does a handstand"}
         \label{fig: handstand}
     \end{subfigure}
     
     \caption{\textbf{Text control:} \alg~ generates full-body motion from text-only control. These examples were all generated from a neutral pose, not indicative of the requested motion.}
    \label{fig: text}
\end{figure*}

\subsection{Text Control}

In \cref{fig: text}, we present examples of text-to-motion control with \alg. All motions are initialized in a neutral standing state. From this initial state, the model is provided different text commands, and \alg~ is then able to generate the corresponding behaviors. 
However, while \alg~ shows promising behavior when directed to perform relatively simple atomic behaviors (e.g., `salut', `kick', etc...), it struggles when provided with commands that require long-term reasoning, such as ``a person takes 4 steps and then raises their hands". We speculate that this is the result of the relatively short history in the model's observations, which can hamper long-term reasoning capabilities.

\section{Limitations and Future Work}\label{sec: limitations}

Although \alg~ presents a unified model for controlling physically simulated humanoids, there remains a number of limitations with our model. We identify three main avenues for improvement: motion quality, automatic goal-engineering, and learning new capabilities.

\paragraph{Motion Quality} While \alg~ demonstrates high success rates in generating diverse motions, there are three notable areas for improvement in terms of motion quality. First, some generated motions exhibit unnatural jittering behaviors. This could be mitigated by finetuning \alg~ using a discriminative reward \cite{peng2021amp} to discourage unrealistic behaviors. Second, the controller does not perfectly replicate the entire training dataset, with some challenging motions such as backflips and breakdancing remaining difficult to reproduce.

Furthermore, when navigating irregular terrain, the character tends to mimic standard walking motions rather than planning more suitable foot placements to avoid rugged regions. Although the controller is robust, it does not fully capture the longer-horizon planning behaviors observed in humans. We hypothesize that this limitation stems from the naive mapping of motions from flat to irregular terrains based on the root-to-floor distance normalization. Future work could address this issue by collecting motions with accompanying scene information or developing improved motion retargeting schemes.

\paragraph{Goal-Engineering} Through goal-engineering, \alg~ can be directed to perform a wide range of tasks, which prior methods required training task-specific models. However, designing goals for controlling a large group of characters to produce natural behaviors in more complex scenes (e.g., crowds) could prove challenging and labour-intensive. We are interested in exploring methods to automate the goal-engineering process, for example by leveraging large-language-models \cite{wang2023voyager,ma2023eureka}.

\paragraph{New Capabilities} Going beyond interactions with static scenes, we are interested in expanding the \alg's capabilities to interact with dynamic scenes. For example, a character should be capable of manipulating objects in the scene, moving objects around and using tools. Furthermore, we are interested in applying \alg~ to synthesize more complex multi-agent interactions.

\section{Discussion}\label{sec: discussion}

In this paper, we introduced \alg, a unified model for physics-based character control through motion inpainting. Our model supports diverse control modalities and demonstrates robust performance across various tasks and environments. We presented a novel approach that formulates the problem of generating motions for physically simulated characters as an inpainting process from partial information (masked). \alg~ is trained on an extensive corpus of randomly masked motion clips. These motion clips contain multi-modal inputs, such as target joint positions/rotations, text descriptions, and objects. By strategically masking out portions of the input motion data, the model is forced to learn to fill in the missing information in a coherent and diverse manner. This allows the system to generate a wide variety of plausible motions that satisfy a flexible variety of different constraints, without requiring exhaustive manual specification of the entire target motion sequence.

We demonstrate that many tasks in character animation can be intuitively described through the lens of partial constraints. These partial constraints specify the goal-directed aspects of the desired motion. Among the examples we presented were joystick steering, reaching, VR-tracking, full-body tracking, path-following, object interaction, and even text-to-motion synthesis. Remarkably, a single unified control architecture can be used to perform all of these diverse tasks without any task-specific training or reward engineering. The key insight is that by learning to generate motions from partial specifications, the model acquires a general understanding of how to produce realistic and physically-plausible character movements that can be directed towards various goals. This approach greatly simplifies the animation process and enables more intuitive and flexible control over the behaviors of physically simulated characters.

\section{Acknowledgements}
We would like to express our sincere gratitude to Haggai Maron for his invaluable assistance in the writing process of this paper. His insights and feedback significantly improved the quality and clarity of our work. We also thank Liang Pan for scene interaction examples, which helped illustrate key concepts in our research. Additionally, we are grateful to the anonymous reviewers for their constructive comments that enhanced the final version of this paper.

\bibliographystyle{ACM-Reference-Format}
\bibliography{bibliography}

\newpage

\onecolumn
\twocolumn

\clearpage

\onecolumn

\appendix

\section{State and action space}

In this work, we consider a 3D physically-simulated humanoid character based on the neutral SMPL body shape, with 69 degrees of freedom. A similar character was used, amongst others, in UHC \cite{luo2022universal}, PACER \cite{rempe2023trace}, and PHC \cite{luo2023perpetual}. To encode the state, we follow the same representation technique from \citet{peng2021amp}. The agent observes:
\begin{itemize}
    \item Root (character's pelvis) height.
    \item Root rotation with respect to the character's local coordinate frame.
    \item Local rotation of each joint.
    \item Local velocity of each joint.
    \item Positions of hands and feet, in the character's local coordinate frame.
\end{itemize}

The character's local coordinate frame is defined with the origin located at the root, the x-axis oriented along the root link's facing direction, and the y-axis aligned with the global up vector. The 3D rotation of each joint is encoded using two 3D vectors, corresponding to the tangent \textbf{u} and the normal \textbf{v} of the link's local coordinate frame expressed in the link parent's coordinate frame \cite{peng2021amp}. In total, this results in a 358D state space.

To control the character, the agent provides an action $a$ that represents the target rotations for PD controllers, which are positioned at each of the character's joints. Similar to \citet{peng2021amp,peng2022ase,juravsky2022padl,tessler2023calm}, the target rotation for 3D spherical joints is encoded using a 3D exponential map \cite{grassia1998practical}, resulting in a 69D action space.

\section{\alg -- Prior Construction}

\textbf{Any-joint-any-time constraints:} A body part can be constrained on both position and rotation. Any joint can be constrained at any timestep during the trajectory. Each such constraint is represented using a 12D vector, representing the constraint (position or rotation) with respect to the humanoid's current local coordinate frame (6D) and with respect to the corresponding body part. E.g., the left-hand spatial constraint will be $[p_\tau^\text{left hand} - \hat{p}_t^\text{left hand}, p_\tau^\text{left hand} - \hat{p}_t^\text{root}]$.

The translation constraints are zero-padded from 3D to 6D whereas the rotation constraints are represented naturally in 6D.

When a target-pose is not provided (no constraints at all), we mask it from entering the transformer.

\textbf{Target objects:} An object is represented by its bounding box (8 coordinates, 3D each), its direction (6D), and its category type (1D). Each coordinate, and the direction, are represented w.r.t. the character's local coordinate frame.

\textbf{Text:} To represent the textual labels we opt for using XCLIP \cite{ni2022expanding}, a CLIP-like model trained on video-language pairs. XCLIP embeds each sentence into a 512-dim vector.

\textbf{Historical poses:} We store the previous 40 poses, with 1 out of each 8 provided at each step, similar to SuperPADL \cite{superpadl2024}. Each pose is represented w.r.t. the current character's local coordinate frame. A time entry is appended to each pose vector, representing how far back that pose is.

When historical data does not exist (start of episode), we mask out the non-existent historical poses.
When a motion is initialized mid-sequence, we initialize the historical data using the historic kinematic poses.

\textbf{Heightmap:} The heightmap surrounding the character is represented by 16x16 samples on a square grid, rotated with the character. The points are spaced with 10cm gaps.

The heightmap is flattened and encoded using a fully connected encoder. Similar to the current character state, the heightmap is always provided.

\begin{figure}
    \centering
    \includegraphics[width=0.5\linewidth]{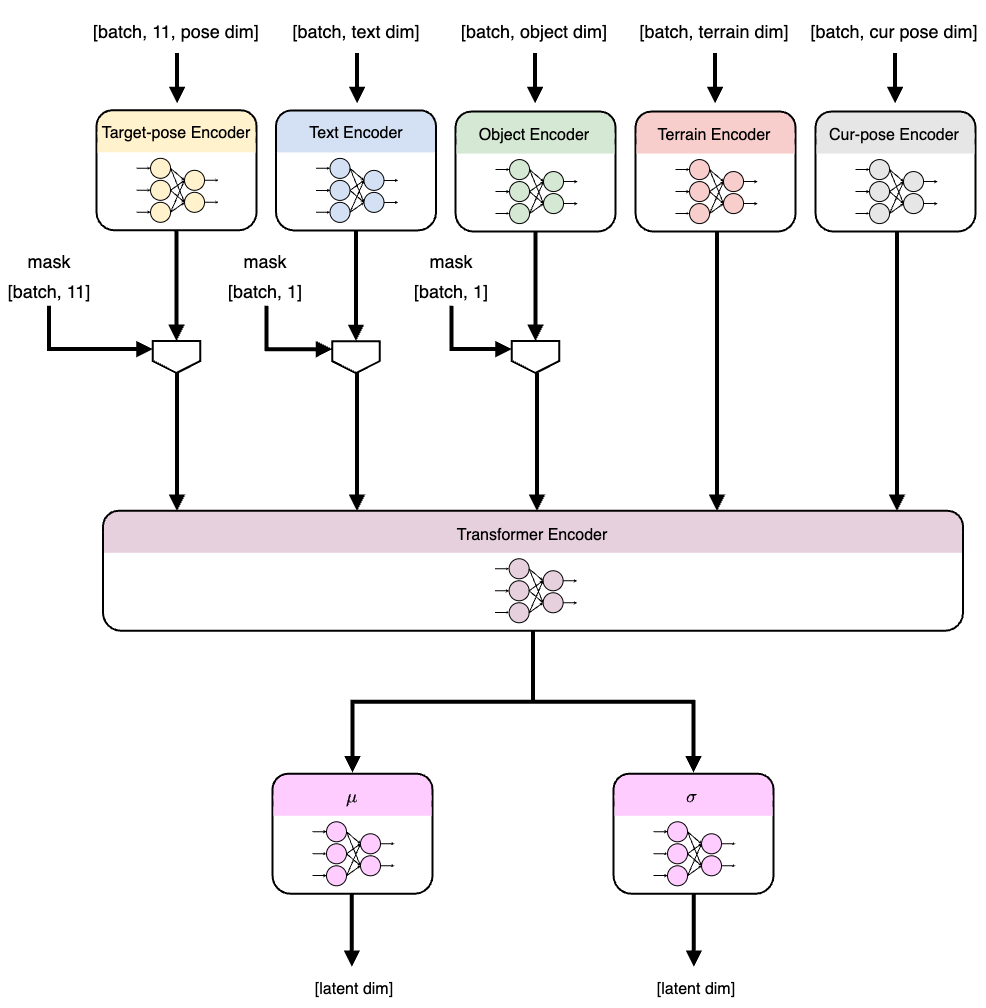}
    \caption{Illustration of the prior architecture. Each modality shares an encoder across all inputs of that same modality. Goal-modalities can be masked out. Each entry is a token, provided to a transformer-encoder model. The output from the transformer is provided to two fully-connected heads, producing the mean and log-std of the latent distribution.}
    \label{fig: prior transformer}
\end{figure}

\textbf{Architecture:} An illustration of the prior architecture is provided in \cref{fig: prior transformer}. Each of the observations above is encoded using a shared encoder for each entry type. For example, all historical poses are encoded using the same encoder, resulting in 5 tokens. Whereas the text embedding is encoded using a different and unique encoder.

Before encoding, each of these inputs is normalized using a running-mean std normalizer, a standard practice in reinforcement learning literature.

Each of the encoders above is constructed as a fully connected network with hidden size [256, 256].

The transformer utilizes a latent dimension of 512 whereas the internal feed-forward size is 1024. We use 4 layers and 4 self-attention heads. We do not use dropout.

All the obtained tokens above are fed into the transformer, alongside the current-pose encoding.

We take the first output from the transformer and pass that through two MLP-heads ($\text{latent dim} \rightarrow 256 \rightarrow 128 \rightarrow 64$) to obtain the distribution $N(\mu, \sigma)$.

For consistency, during training and inference, we maintain a fixed noise throughout an episode. Using the reparametrization trick:
\begin{equation}
    z_t = \mu_t + \sigma_t * \epsilon \,,
\end{equation}
we re-sample $\epsilon$ when an episode terminates.

\textbf{Encoder:} The encoder is modeled as an MLP, with hidden layers [1024, 1024, 1024]. The encoder is provided the current pose, heightmap, unmasked future poses and the masks for the future poses.
The encoder main network output size is 1024 which is then mapped to the posterior's mean and variance using an MLP with hidden-size 512.

\textbf{Decoder:} The decoder observes the current character's pose, the terrain and the sampled latent z. It is modeled as a fully connected network [1024, 1024, 1024]. It's output is the action, the PD targets.

The full-architecture is illustrated in \cref{fig: vae}.

\subsection{Training hyperparameters}

Each episode starts from a randomly sampled state from the distribution dictated by the prioritized sampling mechanism. For each episode we sample a unique noise, which is then kept fixed until the episode is reset. This noise is used for the latent sampling using the reparametrization trick. During the episode, the agent interacts with the environment for rollouts of $32$ steps. \alg~ is then trained with respect to the collected rollout using A2C \cite{mnih2016asynchronous} with $\gamma=0.99$. We utilize GAE \cite{schulman2015high} with $\tau=0.95$. The actor learning rate is fixed at $2e-5$, the critic at $1e-4$. We use the Adam optimizer \cite{kingma2014adam} with betas $[0.9, 0.999]$. The activations are selected as ReLU. In addition, the observations are normalized using the standard running mean-std normalization scheme.

For the reward parameters we set:
\begin{equation}
    w^\text{gr} = 0.3, w^\text{gt} = 0.5, w^\text{jv} = 0.1, w^\text{jav} = 0.1, w^\text{rh} = 0.2, w^\text{enrg}=0.0005 \,,
\end{equation}
and
\begin{equation}
    c^\text{gr} = 2, c^\text{gt} = 100, c^\text{jv} = 0.5, c^\text{jav} = 0.1, c^\text{rh} = 100 \,.
\end{equation}

For \alg \ the KL coeff is scaled from 0.0001 to 0.01 across 6000 epochs, starting from epoch 3000.

We parallelize training over 16,384 Isaac Gym \cite{makoviychuk2021isaac} environments across 4 A100 GPUs, for a total of 14 days. The policy takes decisions at a rate of 30 Hz. The latent space $\mathcal{Z}$ is defined using 64D.

When training the fully constrained controller, we use a mini-batch size of 16,384 on each GPU. We found that A2C provides more stability and thus improved end-performance. To ensure on-policy learning, we perform gradient accumulation over the entire epoch. For \alg \ we opt for 8,192 and a single epoch using a supervised behavior cloning objective. As \alg~ is trained with a supervised learning objective, do not perform gradient accumulation and allow it to perform more gradient steps.

\subsection{Masking}

We describe the masking logic below, with a snippet of the code in \cref{code: masking function}.

Our model is trained with $K=11$ future poses. The first 10 are reserved for near-term constraints (the 10 immediate frames), whereas the 11th entry is reserved for a random long-term pose. 

For any of the first 10 future-poses $q_t$, the sparsity pattern depends on the pattern preceding it. With probability 98\% we apply the same pattern as before, whereas with probability 2\% we sample a random subset of observable joints and their constraint type (position and rotation).

In addition, with probability 1\% we sample a time gap. The size of the gap is randomly sampled between 1 and 9 (including). A time gap denotes a sequence of poses that are completely masked out. This way the model encounters the challenge of inbetweening but is ensured to have at least 1 observable future pose.

When there are "long-term" constraints, such as a long-term target pose, text, or a target-object, we multiply the sampled time-gap by 4. This allows periods in which the model generates motion without any near-term constraints.

In each episode, if there is an object, there is a 20\% chance it will be masked out (hidden). For text, there is 80\% chance it will be masked out, and a long-term target-pose will be provided with 20\% chance.

\begin{lstlisting}[language=Python, caption=Masking code snippet, label=code: masking function]
# conditionable_bodies_ids: Tensor for supported body indices.
# time_gap_mask_steps: Counter for each env for how remaining time-gap.
# motion_text_embeddings_mask, object_bounding_box_obs_mask, target_pose_obs_mask (long-distance target): Mask for observation type, per env.

# Returns: mask of size [num_envs, len(conditionable_bodies_ids) * 2], whether the joint is constrained or not, for both position and rotational constraints.

def sample_body_masks(self, num_envs: int, env_ids: Tensor, new_episode: bool):
    new_mask = torch.zeros(num_envs, self.conditionable_bodies_ids.shape[0], 2, dtype=torch.bool, device=self.device)
    no_time_gap_or_repeated = torch.ones(num_envs, dtype=torch.bool, device=self.device)

    if not new_episode:
        # Reset the time gap mask if the time has expired
        remaining_time = self.time_gap_mask_steps.cur_max_steps[env_ids] - self.time_gap_mask_steps.steps[env_ids]

        # Mark envs with active time-gap
        active_time_gap = remaining_time > 0
        no_time_gap_or_repeated[active_time_gap] = False

        # For those without, check if it should start one
        restart_timegap = (remaining_time <= 0) & (torch.rand(num_envs, device=self.device) < self.config.time_gap_probability)
        self.time_gap_mask_steps.reset_steps(env_ids[restart_timegap])

        # If we have text or object conditioning or target pose, then allow longer time gaps
        text_mask = restart_timegap & self.motion_text_embeddings_mask[env_ids].view(-1)
        object_mask = restart_timegap & self.object_bounding_box_obs_mask[env_ids].view(-1)
        target_pose_obs_mask = restart_timegap & self.target_pose_obs_mask[env_ids].view(-1)
        allow_longer_time_gap = text_mask | object_mask | target_pose_obs_mask
        self.time_gap_mask_steps.cur_max_steps[env_ids[allow_longer_time_gap]] *= 4

        # Where there's no time-gap, we can repeat the last mask
        repeat_mask = (remaining_time < 0) & (torch.rand(num_envs, device=self.device) < self.config.repeat_mask_probability)
        single_step_mask_size = self.conditionable_bodies_ids.shape[0] * 2
        new_mask[repeat_mask] = self.target_bodies_masks[env_ids[repeat_mask], -single_step_mask_size:].view(-1, self.conditionable_bodies_ids.shape[0], 2)

        no_time_gap_or_repeated[repeat_mask] = False

    # Compute number of active bodies for each env
    max_bodies_mask = torch.rand(num_envs, device=self.device) >= 0.1
    max_bodies = torch.where(max_bodies_mask, self.conditionable_bodies_ids.shape[0], 3)
    random_floats = torch.rand(num_envs, device=self.device)
    scaled_floats = random_floats * (max_bodies - 1) + 1
    num_active_bodies = torch.round(scaled_floats).long()

    # Create tensor of [num_envs, len(self.config.conditionable_bodies)] with the max_bodies dim being arange
    active_body_ids = torch.zeros(num_envs, self.conditionable_bodies_ids.shape[0], device=self.device, dtype=torch.bool)
    constraint_states = torch.randint(0, 3, (num_envs,  self.conditionable_bodies_ids.shape[0]), device=self.device)

    for idx in range(num_envs):
        # Sample the active body ids for each env
        active_body_ids[idx, np.random.choice(self.conditionable_bodies_ids.shape[0], size=num_active_bodies[idx].item(), replace=False)] = True
    
    translation_mask = (constraint_states <= 1) & active_body_ids
    rotation_mask = (constraint_states >= 1) & active_body_ids

    new_mask[no_time_gap_or_repeated, :, 0] = translation_mask[no_time_gap_or_repeated]
    new_mask[no_time_gap_or_repeated, :, 1] = rotation_mask[no_time_gap_or_repeated]

    return new_mask.view(num_envs, -1)
\end{lstlisting}

\end{document}